\documentclass[final]{cvpr}
\usepackage{times}
\usepackage{epsfig}
\usepackage{graphicx}
\usepackage{amsmath}
\usepackage{amssymb}
\usepackage{color}
\usepackage{subfigure}
\usepackage{float}
\usepackage{makecell}
\usepackage[numbers,sort,compress]{natbib}
\usepackage{caption}
\usepackage{nicefrac}

\usepackage{tabularx}
\usepackage{array}

\makeatletter
\@namedef{ver@everyshi.sty}{}
\makeatother

\usepackage{tikz}
\usetikzlibrary{tikzmark, calc, fit}

\newcolumntype{Y}{>{\centering\arraybackslash}X}

\usepackage{multirow}
\usepackage{booktabs}
\usepackage{inconsolata}
\usepackage{xcolor}
\usepackage{comment}

\usepackage[pagebackref=true,breaklinks=true,colorlinks,bookmarks=false]{hyperref}
\usepackage{cleveref}

\DeclareMathSymbol{@}{\mathord}{letters}{"3B}

\newcommand{\best}[1]{\textbf{#1}}

\newcommand{\DATASET}{Mirror3D\xspace}

\newcommand{\mnet}{Mirror3DNet\xspace}

\newcommand{\nyuraw}{NYUv2-raw\xspace}
\newcommand{\nyuref}{NYUv2-ref\xspace}

\newcommand{\mpmesh}{MP3D-mesh\xspace}
\newcommand{\mpmeshref}{MP3D-mesh-ref\xspace}

\newcommand{\denselist}{\itemsep 0pt\parsep=0pt\partopsep 0pt\vspace{-\topsep}}
\newcommand{\mypara}[1]{\vspace{2pt}\noindent\textbf{#1}}

\DeclareMathOperator{\CE}{CE}
\DeclareMathOperator{\Smooth}{Smooth}

\def\cvprPaperID{10153} %
\def\confYear{CVPR 2021}

\begin{document}

\title{Mirror3D: Depth Refinement for Mirror Surfaces}

\author{Jiaqi Tan \quad Weijie Lin \quad Angel X. Chang \quad Manolis Savva\\
Simon Fraser University\\
\url{https://3dlg-hcvc.github.io/mirror3d/}
}

\twocolumn[{%
\renewcommand\twocolumn[1][]{#1}%
\maketitle
\begin{center}
\includegraphics[width=\linewidth]{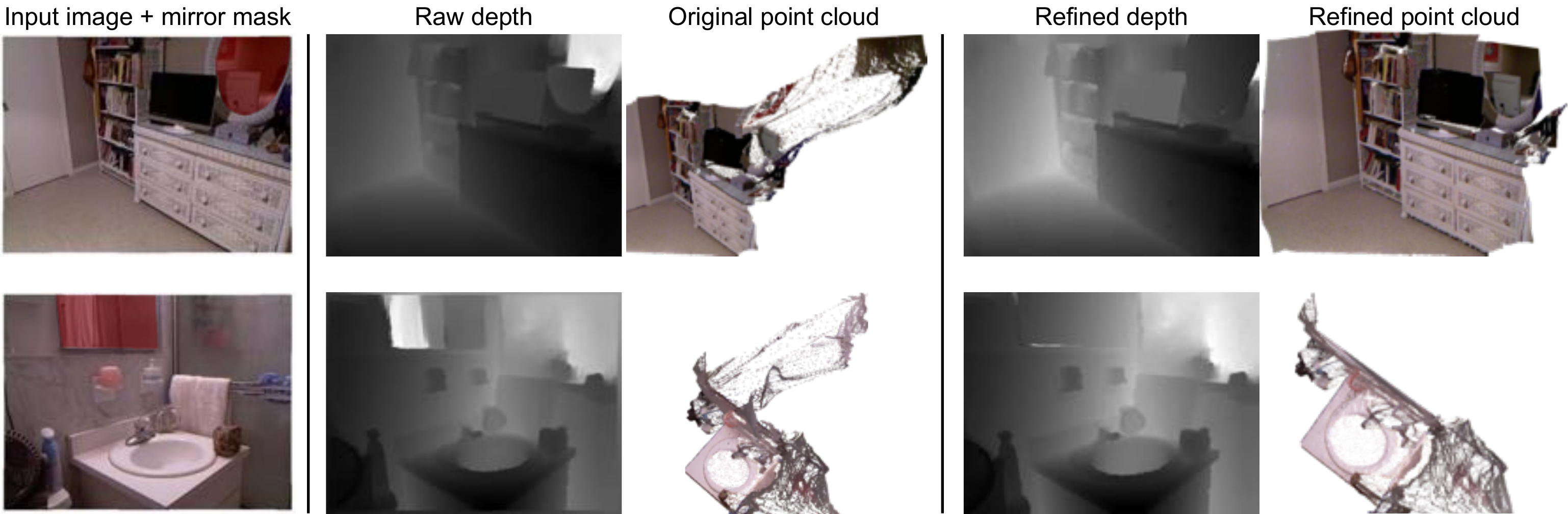}
\captionof{figure}{
We present the task of 3D mirror plane prediction and depth refinement.
First, we annotate several popular RGBD datasets (Matterport3D~\cite{matterport3d}, ScanNet~\cite{Scannet}, NYUv2~\cite{nyuv2}) with 3D mirror planes.
Our benchmarks show that both existing RGBD dataset `ground truth' raw depth data, and state-of-the-art depth estimation and depth completion methods exhibit dramatic errors on mirror surfaces.
We propose an architecture for 3D mirror plane estimation that refines depth estimates and produces more reliable reconstructions (compare left and right depth and point cloud pairs from NYUv2~\cite{nyuv2} dataset).}
\label{fig:teaser}
\end{center}
}]

\maketitle

\begin{abstract}
Despite recent progress in depth sensing and 3D reconstruction, mirror surfaces are a significant source of errors.
To address this problem, we create the Mirror3D dataset: a 3D mirror plane dataset based on three RGBD datasets (Matterport3D, NYUv2 and ScanNet) containing $7@011$ mirror instance masks and 3D planes.
We then develop \mnet: a module that refines raw sensor depth or estimated depth to correct errors on mirror surfaces.
Our key idea is to estimate the 3D mirror plane based on RGB input and surrounding depth context, and use this estimate to directly regress mirror surface depth.
Our experiments show that \mnet significantly mitigates errors from a variety of input depth data, including raw sensor depth and depth estimation or completion methods.
\end{abstract}

\section{Introduction}

Recent years have seen much progress in 3D reconstruction methods.
It is now possible to acquire 3D reconstructions of interiors with high quality geometry and texture.
However, these reconstructions fail spectacularly when used in environments with mirrors and glass, both of which are prevalent in indoor spaces.

The fragility of many reconstruction methods is due to a reliance on accurate active depth sensing.
Commodity sensors such as the Microsoft Kinect and Intel RealSense employ active infrared or time of flight depth sensing which requires strong signal return from sensed surfaces.
Unfortunately, highly glossy and reflective surfaces such as mirrors cause either no signal return or highly unreliable depth estimates.
Though it may seem that reflectors are a `corner case' with few affected pixels, the resulting errors are catastrophic to reconstruction algorithms, leading to loss of camera pose tracking and geometry artifacts (see \Cref{fig:teaser}).

Recently, \citet{whelan2018reconstructing} demonstrated that by detecting mirror planes it is possible to mitigate the above issues and achieve high-quality reconstructions in scenes with many reflectors.
However, their method relies on the use of an AprilTag~\cite{olson2011tags} attached to a custom camera rig.
The use of such custom hardware is not always feasible.
In this paper, we propose a method for identifying mirrors and estimating mirror surface depth on RGBD data collected with commodity hardware.

Our key idea is to identify mirror regions based on color information (RGB), model the mirror as a plane and use an estimated mirror normal and information from the mirror's surroundings to predict the mirror's position in 3D.
Our work is related to the recently proposed task of mirror segmentation in 2D~\cite{yang2019where, lin2020progressive}.
However, we operate in 3D and focus on using 3D mirror plane estimates to improve the reliability of depth data.

To estimate the prevalence of mirrors in 3D environments and understand the severity of the above reconstruction issues, we first annotate 3D mirror planes in three popular RGBD datasets: Matterport3D~\cite{matterport3d}, ScanNet~\cite{Scannet}, and NYUv2~\cite{nyuv2}.
These annotations create `true' ground truth for observed mirror surfaces that was previously unavailable.
We find the prominence of mirrors varies between datasets depending on their acquisition procedure, leading to a corresponding range of depth data issues and reconstruction failures.
Using this data, we introduce the task of 3D mirror detection from RGB and RGBD data, and establish initial benchmark results by: i) applying state-of-the-art depth estimation and depth completion approaches to directly estimate mirror depth values; and ii) propose a simple architecture combining a MaskRCNN~\cite{maskrcnn} module and a PlaneRCNN~\cite{liu2019planercnn} module to segment mirror surfaces, estimate mirror 3D planes and refine mirror surface depth estimates.
Overall, we make the following contributions:
\begin{itemize}\denselist
\item Introduce the task of 3D mirror plane prediction from single-view RGB and RGBD data
\item Provide 3D mirror plane annotations for three RGBD datasets (Matterport3D, NYUv2 and ScanNet)
\item Establish benchmarks for RGB and RGBD-based 3D mirror plane prediction, and evaluate depth completion and depth estimation approaches on the task
\item Present \mnet, an architecture that predicts a 3D mirror normal and mirror segmentation to refine raw sensor depth or the output of state-of-the-art depth completion and estimation methods
\end{itemize}

\section{Related work}

We summarize related work on mirror detection, 3D plane estimation, and depth estimation and completion.

\mypara{Mirror detection and correction.}
The challenges of dealing with reflective and transparent surfaces in 3D reconstruction and robotics have long been recognized.
\citet{yang2008dealing} proposed a sensor fusion technique for dealing with LiDAR sensor failures on mirror and glass surfaces.
\citet{kashammer2015mirror} detect mirrors and correct laser-scanned point clouds based on heuristics using known mirror dimensions.
The general problem of mirror surface reconstruction is addressed by work on reflectometry.
Early work focuses on the stereo image setting~\cite{balzer2011multiview}, active illumination hardware setups~\cite{balzer2014cavlectometry}, or single image input but relying on detecting reflection correspondences of reference target objects~\cite{liu2013mirror}.
In recent years, there has been renewed interest in identifying mirrors, glass, and transparent objects for improved reconstruction~\cite{whelan2018reconstructing} and for manipulation in robotics~\cite{weng2020multi}.
\citet{whelan2018reconstructing} rely on observing the reflection of an AprilTag~\cite{olson2011tags} attached to a custom scanner.
Work on mirror detection for uncontrolled RGB images using deep learning is less explored, with a few recent works.
\citet{yang2019where} introduce a network and dataset for identifying mirrors in 2D images, and \citet{lin2020progressive} extend the earlier work by extracting mirror context features to improve mirror detection.
In contrast, we study 3D mirror plane estimation in RGBD datasets without relying on custom hardware or other assumptions on the capture setup.

\mypara{3D plane detection and plane reconstruction.}
There is recent work on detecting 3D planes from single-view images using neural networks~\cite{yang2018recovering,liu2018planenet,liu2019planercnn,yu2019single}.
Unlike this work, our focus is not on creating a planar segmentation of the entire observed scene.
We focus specifically on identifying mirror regions and corresponding 3D planes.
Identifying mirrors in RGB images is challenging, as mirrors contain a reflection of the environment which can be difficult to distinguish from non-reflected regions.
Mirror detection remains challenging even with RGBD data as mirror depth values tend to be noisy and unreliable.

\mypara{Depth estimation and depth completion.}
Depth estimation and completion are recently popular tasks.
Typically, the term depth estimation is used when the input is only color (RGB) and has no depth information.
The term depth completion is used when the input is RGBD, where the D (depth) channel is noisy and may have missing values.
Existing methods for single-view depth estimation~\cite{saxena2006learning,eigen2014depth,li2017two,engel2017direct,cao2017estimating,alhashim2018high,mahjourian2018unsupervised,yin2019enforcing,lee2019big,ranftl2020} and depth completion~\cite{zhang2018deep,huang2019indoor,senushkin2020decoder,park2020non,mendes2020deep} improve depth prediction for the entire image, relying on reconstructed 3D mesh data that is assumed to provide accurate depth.
\citet{chabra2019stereodrnet} show that an exclusion mask for noisy areas such as reflective surfaces can result in better reconstruction.
We leverage 3D mirror plane estimates to improve the accuracy of existing depth estimation and completion methods.
Moreover, we note that both depth estimation and completion methods are typically evaluated on ground truth data that does not account for noise due to mirrors and glass.
Widely employed datasets such as NYUv2\cite{nyuv2}, Matterport3D\cite{matterport3d,zhang2018deep}, ScanNet~\cite{Scannet}, and SUN3D~\cite{xiao2013sun3d,halber2017fine} have noisy or missing depth for reflectors (see \Cref{fig:teaser}).
Thus, these regions are typically ignored or evaluated with incorrect values as the ground truth.
We contribute ground truth 3D mirror annotations for three RGBD datasets, allowing for correct benchmarking of mirror surface depth estimation and completion.

\begin{figure*}
  \includegraphics[width=\linewidth]{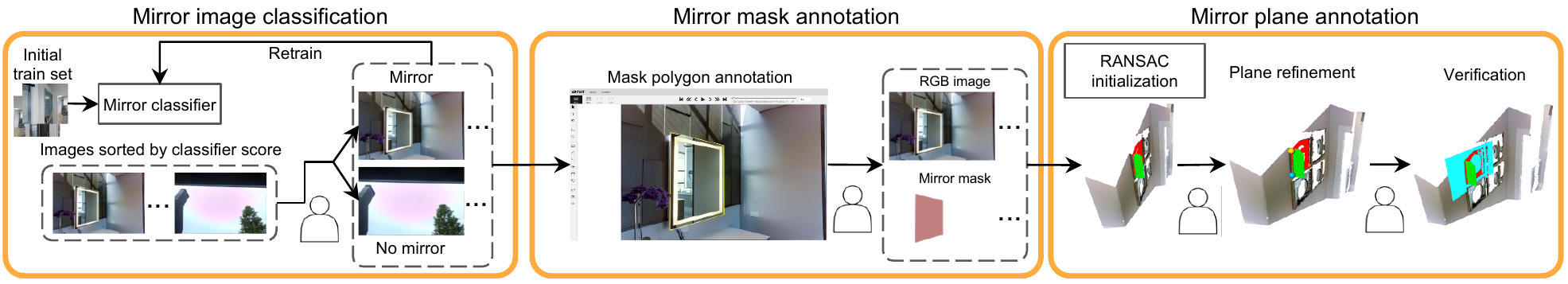}
  \caption{Our human-in-the-loop mirror mask and 3D plane annotation workflow. We leverage an iteratively trained mirror image classifier to assist a user in rapidly selecting all mirror images in an input image dataset. The images are then annotated with mirror surface polygon masks. The masks and their corresponding depth images are used to initialize mirror plane estimates with a RANSAC approach. The user then refines and verifies all 3D mirror plane estimates.}
  \label{fig:annotation-flow}
\end{figure*}

\section{Mirror3D dataset}

To enable benchmarking of 3D mirror plane prediction and surface depth estimation, we create \DATASET: the first large-scale dataset of mirror annotations for RGBD images.
We design a human-in-the-loop workflow that enables efficient iterative annotation of mirror masks and mirror 3D planes.
Using this workflow, we annotate RGBD images containing mirrors in three common RGBD datasets (NYUv2~\cite{nyuv2}, Matterport3D~\cite{matterport3d}, and ScanNet~\cite{Scannet}) to create an aggregated dataset that contains $7@011$ annotated 3D mirror planes in $5@894$ RGBD frames.

\subsection{Dataset construction}

Our annotation workflow consists of three stages: i) mirror image classification, ii) mirror mask annotation, and iii) mirror plane annotation.
\Cref{fig:annotation-flow} shows the overall pipeline.
We first pretrain a ResNet-50 classifier on a seed dataset of `mirror present' and `no mirror present' RGB images from Structured3D~\cite{Structured3D} (roughly half of the images contained mirrors, approximately $7@000$ images total).
We then sort all input RGBD dataset candidate frames by using the mirror classification score.
An annotator confirms whether images contain a mirror, splitting them into `mirror' and `no mirror' sets in batches of $100$ images.
As this verification proceeds through each batch, the classifier is finetuned on the dataset of newly annotated `mirror' and `no mirror' images to improve the efficacy of the classification score sorting whenever less than $20$ additional mirror images are added from a single $100$-image batch.
The annotator was instructed to stop looking for additional mirror images when less than 5\% of a batch contains mirrors.
This first stage was performed by one annotator and took approximately $28$ hours in total over $8$ batch iterations.

In the second stage, we used the CVAT\footnote{\url{github.com/openvinotoolkit/cvat}} annotation interface to define mirror mask polygons for all mirror images.
The annotators specified two types of mirror mask polygons: a `coarse' mask and a `precise' mask.
The coarse mask ignores occluding objects and small mirror protrusions, while the precise mask is a pixel-accurate boundary of the visible mirror surface in each image.
This stage was performed by four annotators over approximately $67$ hours.

The last stage used a 3D interface developed with Open3D~\cite{zhou2018open3d} that allows inspection of an initial 3D mirror plane normal estimate obtained by filtering mirror mask border depth values using RANSAC~\cite{fischler1981random}.
The annotators could check that the plane is correct, refine the 3D plane estimate by specifying three plane points on the point cloud and manually adjusting, or indicate that it is impossible to define a plane (in cases where the point cloud is extremely noisy).
After the 3D planes are annotated, we generated turntable videos of the point clouds with the annotated planes from a frontal and top-down view to allow for quick verification.
Planes identified as incorrect were further adjusted and corrected.
This last stage took a total of approximately $50$ hours of annotation effort by the same expert annotator who carried out the first stage.

In total, the annotators inspected approximately $30@000$ RGBD images to obtain the final $7@011$ 3D mirror plane annotations.
The mirror image classification stage took $3$s per image on average.
The mirror mask annotation stage took $25$s and $57$s per mirror instance for coarse mask and precise mask respectively.
Finally, the mirror plane annotation stage takes $20$s per mirror instance on average.

\begin{table}
\resizebox{\linewidth}{!}{
\begin{tabular}{@{}lrrrr@{}}
\toprule
 & Matterport3D & ScanNet & NYUv2 & \textbf{Total} \\
\midrule
mirror planes & $4@662$ & $2@218$ & $131$ & $7@011$ \\
RGBD images & $3@782$ & $1@987$ & $125$ & $5@894$ \\
RGBD panoramas & $2@468$ & -- & -- & $2@468$ \\
3D scenes & $79$ & $282$ & $96$ & $457$\\
\bottomrule
\end{tabular}
}
\caption{Summary statistics for our \DATASET dataset.}
\label{tab:summary-statistics}
\end{table}

\subsection{Dataset statistics}

At the end of our annotation, we end up with a total of $7@011$ mirror 3D plane and instance mask annotations across RGBD images from all three source datasets.
\Cref{tab:summary-statistics} shows a more detailed breakdown of dataset statistics.
We split this data following the standard training, validation and test splits of each source dataset.
We note that mirrors are found in roughy $22.9\%$ of Matterport3D panoramas (from a total of $79$ out of $90$ Matterport3D residences) and $96$ of $464$ ($20\%$) NYUv2 scenes, which are relatively high fractions compared to only $282$ out of approximately $1@500$ ScanNet scenes ($18.8\%$).
We hypothesize that this is due to the capture setup and methodology employed by the ScanNet authors who may have avoided scanning scene regions that contain mirrors.
\Cref{sec:supp:dataset} provides additional analysis of the annotated data in terms of mirror region location, prominence and distribution across the images.

\begin{figure*}
  \includegraphics[width=\linewidth]{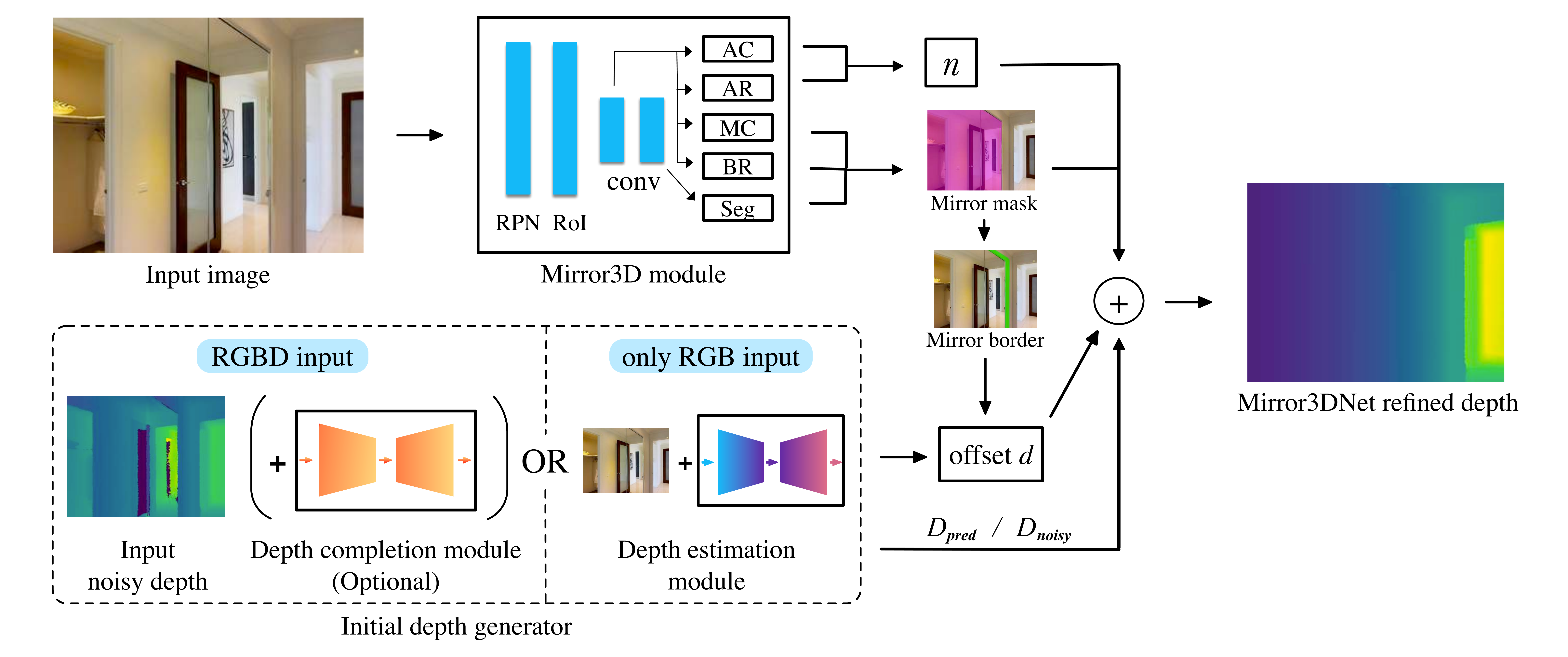}
  \caption{Overall diagram showing how the \mnet architecture can be used for either an RGB image or an RGBD image input.
  For an RGB input, we refine the depth of the predicted depth map $D_{\text{pred}}$ output by a depth estimation module.
  For RGBD input, we refine a noisy input depth $D_{\text{noisy}}$.
  The \mnet module predicts a mirror normal $n$ and mirror mask. Then, by computing an offset depth $d$ based on depth values at the mirror border, we can determine the position and orientation of the 3D mirror plane, and produce refined output depth values that improve mirror surface depth accuracy.}
  \label{fig:network-arch}
\end{figure*}

\section{\mnet: refining mirror depth}

Having constructed our dataset of 3D mirror plane annotations, we would now like to address the task of predicting 3D mirrors from a single-view RGB or RGB-D image, and using these predictions to improve estimated depth values on mirror surfaces.
Because mirrors are often planar, we make the simplifying assumption that we can model a mirror using a mirror mask and a mirror plane.
We will show that we can improve depth estimation from RGB-only image input or depth completion results from an RGBD image using the predicted 3D region and plane.
Thus, the input to our approach is either an RGB image or an RGBD image with missing or incorrect depth for mirrors.
The output is an estimated 3D mirror plane and an RGBD image with refined depth for the mirror.

Our goal is not to propose a new approach for depth estimation or depth completion, but rather to benchmark state-of-the-art approaches for both and demonstrate that we can leverage a simple mirror-aware architecture to improve depth output for both families of approaches.
To do this, we propose the \mnet architecture (see \Cref{fig:network-arch}).
This architecture uses a Mask R-CNN~\cite{maskrcnn} module and a 3D mirror plane estimation module inspired by PlaneRCNN~\cite{liu2019planercnn} to predict mirror masks and planes respectively.
A depth estimation or completion module may be used first to predict depth values given an input RGB or RGBD image, and then \mnet is applied to predict the mirror mask and mirror plane.
We evaluate this architecture's performance on depth prediction as well as mirror mask prediction and mirror normal estimation.

Thus, the overall architecture consists of three modules: mirror mask segmentation, mirror plane estimation, and depth estimation or completion.
PlaneRCNN uses a warping loss to enforce consistency of reconstructed 3D planes from nearby viewpoints.
Our problem setting only assumes a single image input, so we replace the warping loss module with a mirror plane estimation module which refines the depth on the mirror region.

\subsection{Mirror mask and plane prediction}

In this part of the architecture, we address detection and segmentation of the mirror surface.
To do this, we employ a mirror classification (MC) branch and a mirror bounding box regression (MR) branch, as well as an instance segmentation (Seg) branch, all based on the Mask R-CNN module.
We also add an anchor normal classification (AC) branch and an anchor normal regression (AR) branch after the ROI pooling layer to predict the mirror plane normal.

Since it is challenging to directly regress the mirror normal values, these latter two branches follow an approach similar to PlaneRCNN and decompose the regression into a classification phase and a residual prediction phase.
In the classification phase we obtain a coarse orientation of the mirror normal by classifying into one of a few anchor normal orientations using the AC branch.
We use $10$ mirror anchor normals from a k-means clustering of all mirror normals in the annotated Matterport3D training set.
During training, the module predicts one anchor normal for each positive proposal.
For supervision, we assign each mirror instance to its closest mirror anchor normal.

Given the anchor normal classification we then regress the residual using the AR branch to form the final normal vector.
The final normal $n$ is the sum of the anchor normal and the normal residual.
The ground truth mirror residual of an instance is the distance vector between ground truth mirror and the ground truth mirror anchor normal.

\subsection{Depth estimation and completion}

As \Cref{fig:network-arch} shows, our network architecture accommodates a depth estimation or completion module.
The details of this part depend on the input type.
That is, whether we take an RGB image or RGBD data including noisy depth.
If the input is an RGB image, we first use a depth estimation module to produce an initial depth map estimate.
Then the rest of the Mirror3D architecture refines the depth map into $D_{\text{pred}}$.
If the input is an RGBD image with a noisy depth map $D_{\text{noisy}}$, we directly carry it forward for refinement through the architecture.

To estimate the 3D mirror plane we need to combine the mirror mask and mirror normal estimates from the previous stage with a depth offset $d$ for the plane.
Since depth values on the mirror surface itself are missing or unreliable, we rely on mirror border regions which are often non-reflective materials and significantly more reliable.
Thus, we compute the average depth of points that fall on pixels within a small offset of the mirror mask border.
Given the predicted mirror segmentation mask, we create a mirror border mask region by expanding outwards by $25$ pixels.
This threshold corresponds to roughly 5\% of average image dimensions and we have empirically found it to be reasonable.
We then take the average depth of all points in this mirror mask as the offset of the mirror plane $d$.

The \mnet module is trained under three types of loss terms: i) mirror segmentation loss, ii) mirror anchor normal classification loss, and  iii) mirror anchor normal regression loss.
In mirror segmentation, we use a cross-entropy loss for mirror classification and mirror mask prediction.
We use a smooth L1 loss for mirror bounding box regression, and a cross-entropy loss for mirror anchor normal classification.
For mirror anchor normal regression, we use a smooth L1 loss.
The total loss $\mathcal{L}$ is then:
\begin{align*}
    \mathcal{L} &= \CE(a_i, a_i^*) + \Smooth_{L1}(r_i, r_i^*) \\ 
     &+ \CE(c_i, c_i^*)  + \Smooth_{L1}(b_i, b_i^*) 
     + \CE(m_i, m_i^*) 
    \label{eq:loss}
\end{align*}
where $a_i$ and $a_i^*$ are the predicted and ground truth mirror anchor normal class,
$r_i$ and $r_i^*$ are the $1 \times 3$ predicted mirror regression vector and ground truth mirror regression vector,
$c_i$ and $c_i^*$ are the predicted mirror proposal class and ground truth proposal class,
$b_i$ and $b_i^*$ are the predicted and ground truth mirror bounding box parameter,
$m_i$ and $m_i^*$ are the predicted and ground truth mirror mask.

\subsection{Implementation details}

We implement our network architecture in PyTorch~\cite{paszke2019pytorch}.
We used the Adam~\cite{kingma2014adam} optimizer with initial learning rate set to $10^{-4}$, $\beta_{1} = 0.9$ and $\beta_{2} = 0.999$ and without weight decay.
The ResNet-50~\cite{resnet50} backbone was initialized with weights pretrained on ImageNet~\cite{imagenet}.
We train our model for $50@000$ iteration with batch size $32$.
For NYUv2 data, we crop the original image by 2.5\% on each side to remove the white border and resize to $480 \times 640$.
We resize Matterport3D data to $512 \times 640$.
For depth estimation and depth refinement experiments on Matterport3D, we train with all Matterport3D train set depth data and fine-tune on Matterport3D mirror data.

\section{Experiments}

\subsection{Datasets}

We carry out our evaluation on the NYUv2 and Matterport3D datasets, following their train/val/test splits.
To show the importance of accurate mirror depth we evaluate against both original `raw depth' ground truth depth and our annotated mirror depth, computed using the `precise' masks. 
We use the following depth data in our experiments:

\mypara{\nyuraw:}
the $1449$ annotated RGBD frames from NYUv2 ($795$ frames in train and $654$ frames in test), captured with a Kinect depth sensor and exhibiting noise and missing data on mirror surfaces.

\mypara{\nyuref:}
the corrected version of NYUv2-raw, with mirror surface depth computed using our 3D mirror plane annotations.

\mypara{\mpmesh:}
depth rendered from the Matterport3D reconstructed mesh, as used by \citet{zhang2018deep}.

\mypara{\mpmeshref:}
the corrected version of the above using our 3D mirror plane annotations.

\subsection{Evaluation metrics}

\mypara{Depth estimation and completion.}
We report several standard metrics for depth estimation and completion~\cite{eigen2014depth,zhang2018deep,ranftl2020}:
root mean squared error (RMSE), scale-invariant root mean squared error (s-RMSE), mean absolute value of the relative error (AbsRel), structural similarity index measure (SSIM), and $\delta_{i}$.
RMSE is commonly used to measure disparity error, but since depth estimation methods are often limited by the size-distance ambiguity in predicting depth values, we compute s-RMSE~\cite{eigen2014depth} which assumes the optimal scale is available.
Following \citet{huang2019indoor}, we also report SSIM~\cite{wang2004image}, which measures the perceptual similarity of large-scale structures between images.
The $\delta_{i}$ metric denotes the percentage of predicted pixels where the relative error is less than a threshold $i$.
Specifically, $i$ is chosen to be equal to $1.05,1.10,1.25,1.25^{2} \text { and } 1.25^{3}$.
Here, the larger $i$ is, the more sensitive the $\delta_{i}$ metric.
Larger values of $\delta_{i}$ reflect a more accurate prediction.
We find the metrics to be largely correlated and report the RMSE and SSIM in the main paper, and provide complete results in \Cref{sec:supp:quantitative}.
Following standard practice~\cite{zhang2018deep}, we only evaluate on pixels with ground truth depths $>0.00001$ as depth of $0$ often indicates no depth reading.

\mypara{RMSE}:
$\sqrt{\nicefrac{1}{|P|} \sum_{p \in P}\left\|D^{*}(p)-D(p)\right\|^{2}}$ where $D(p)$ and $D^{*}(p)$ is depth and ground truth depth at point $p$.

\mypara{s-RMSE}:
$\sqrt{\nicefrac{1}{|P|} \sum_{p \in P}\left\|D^{*}(p)-s D(p)\right\|^{2}}$ where $s$ is the least square root of $\nicefrac{1}{n} \sum \left ( D^{*} -  sD \right )^2$.

\mypara{AbsRel}:
$\nicefrac{1}{|P|} \sum_{p \in P}\nicefrac{|D^{*}(p) - D(p)|}{D^{*}(p)}$.

\mypara{SSIM}: 
$\frac{\left(2 \mu_{D^{*}(p)} \mu_{D(p)}+c_{1}\right)\left(2 \sigma_{D^{*}(p) D(p)}+c_{2}\right)}{\left(\mu_{D^{*}(p)}^{2}+\mu_{D(p)}^{2}+c_{1}\right)\left(\sigma_{D^{*}(p)}^{2}+\sigma_{D(p)}^{2}+c_{2}\right)}$
where $c_{1}=0.0001, c_{2}=0.0009$ as in \citet{huang2019indoor}.

\mypara{$\boldsymbol{\delta_{i}}$}: percentage of pixels within error range $i$, where the error range is defined by 
$\max \left(\nicefrac{D^{*}(p)}{D(p)}, \nicefrac{D(p)}{D^{*}(p)}\right)<i$.

All the above metrics are reported separately for depth points within the ground truth mirror region, points outside the mirror, and together for all depth points in each frame.

\mypara{Mirror mask prediction and plane estimation.}
We adopt the commonly used mean average precision (mAP) to evaluate mirror mask segmentation predictions (\textbf{Seg-AP}).
We also extend the mAP to take into account whether the angle error is below $30^{\circ}$ (\textbf{$30^{\circ}$-AP}).

\mypara{Seg-AP}:
segmentation AP, with IoU threshold starting from $0.50$ to $0.95$ at $0.05$ steps.
The final AP score is the average over the $10$ threshold steps.

\mypara{30$^{\circ}$-AP}:
30 degree AP, with the same IoU threshold setting as Seg-AP and an angle error threshold of 30 degrees.
Angle error is the angle between predicted mirror normal and ground truth mirror normal.

\subsection{Qualitative evaluation}

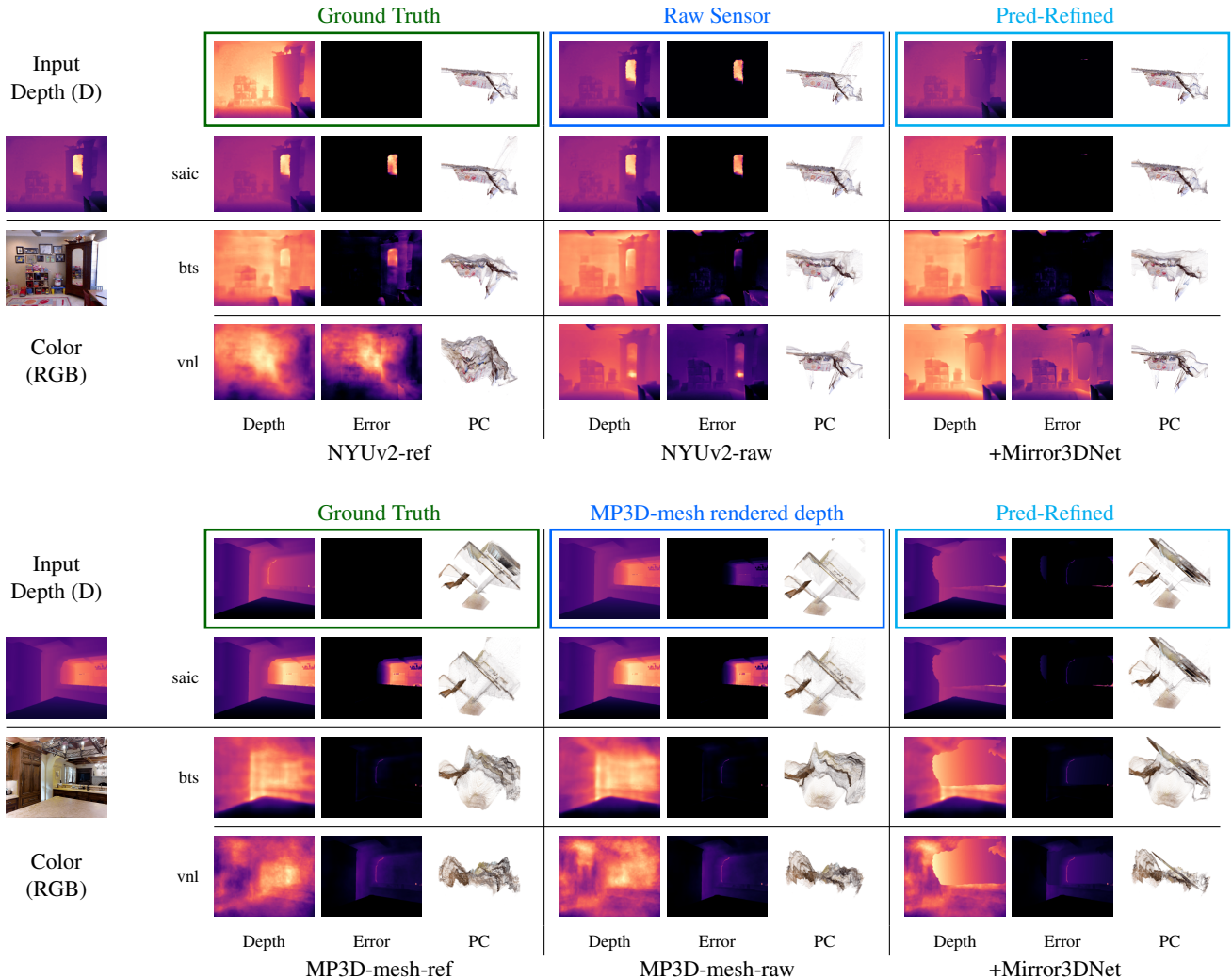
\begin{figure*}

\centering
    
\newcolumntype{C}{>{\centering\arraybackslash}X} %
\setkeys{Gin}{width=\linewidth}
\begin{tabularx}{\textwidth}{Y@{\hspace{-13mm}} Y@{\hspace{1mm}} Y@{\hspace{0mm}} Y@{\hspace{1mm}} Y@{\hspace{1mm}} Y@{\hspace{2mm}} | Y@{\hspace{1mm}} Y@{\hspace{1mm}} Y@{\hspace{2mm}} | Y@{\hspace{1mm}} Y@{\hspace{1mm}} Y@{\hspace{0mm}} }
 & & & \multicolumn{3}{c}{\textcolor{black!60!green}{\small Ground Truth}} & \multicolumn{3}{c}{\textcolor{blue!60!cyan}{\small Raw Sensor}} & \multicolumn{3}{c}{\textcolor{cyan}{\small Pred-Refined}}\\
& \small Input Depth (D) &  & \vspace{4pt}\tikzmarknode{A1}{\includegraphics{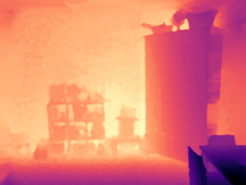}} & \vspace{4pt}\includegraphics{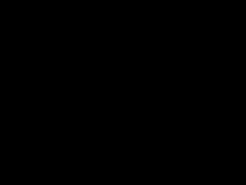} & \vspace{4pt}\tikzmarknode{A2}{\includegraphics{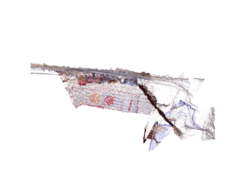}} & \vspace{4pt}\tikzmarknode{B1}{\includegraphics{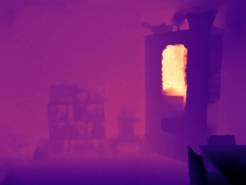}} & \vspace{4pt}\includegraphics{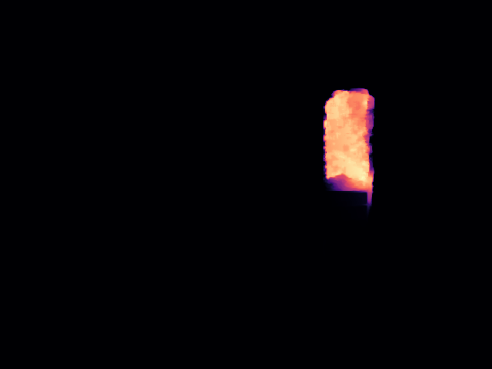} & \vspace{4pt}\tikzmarknode{B2}{\includegraphics{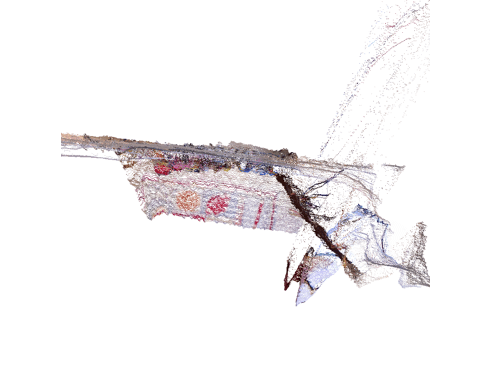}} & \vspace{4pt}\tikzmarknode{C1}{\includegraphics{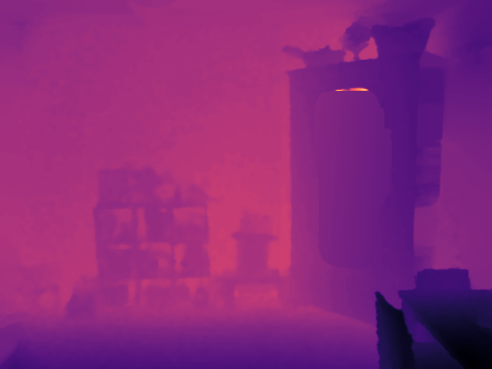}} & \vspace{4pt}\includegraphics{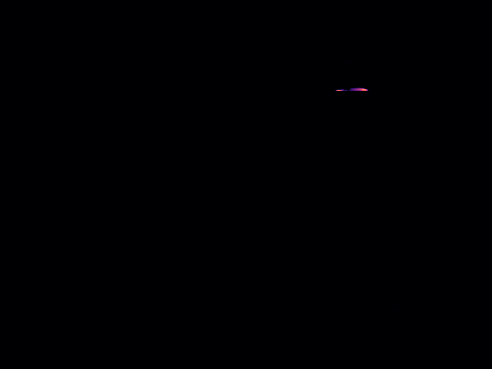} & \vspace{4pt}\tikzmarknode{C2}{\includegraphics{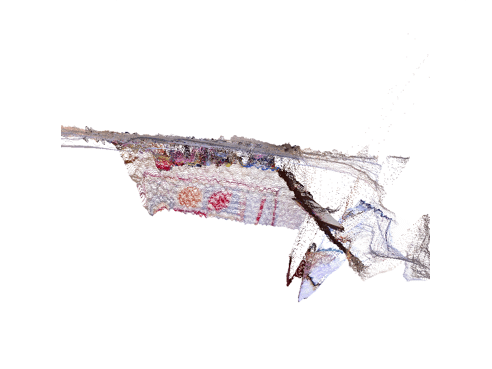}}
\\ &
\vspace{4pt}\includegraphics{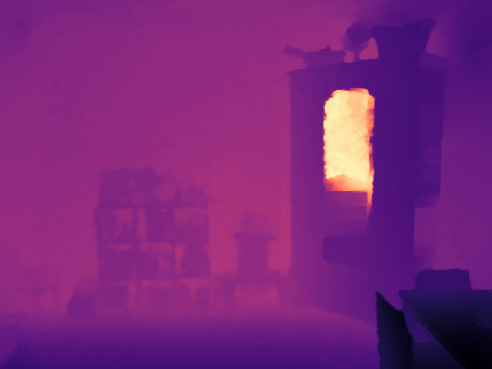} & \multicolumn{1}{r}{\scriptsize saic} & \vspace{4pt}\includegraphics{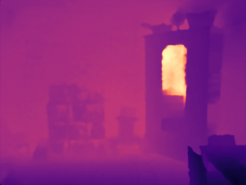} & \vspace{4pt}\includegraphics{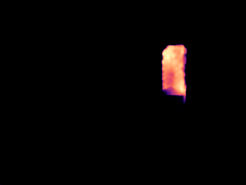} & \vspace{4pt}\includegraphics{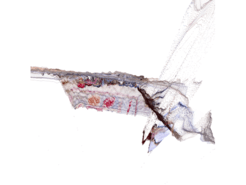} & \vspace{4pt}\includegraphics{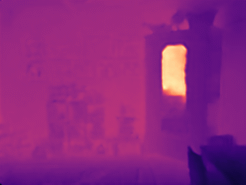} & \vspace{4pt}\includegraphics{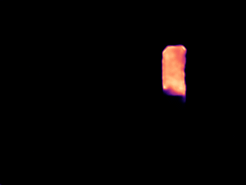} & \vspace{4pt}\includegraphics{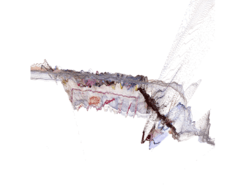} & \vspace{4pt}\includegraphics{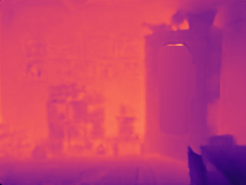} & \vspace{4pt}\includegraphics{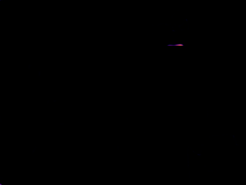} & \vspace{4pt}\includegraphics{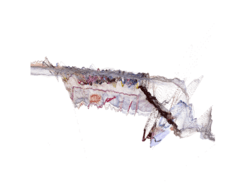} \\\cline{2-12} & \vspace{4pt}\includegraphics{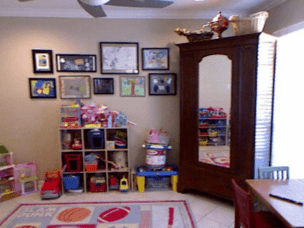} & \multicolumn{1}{r}{\scriptsize bts} & \vspace{4pt}\includegraphics{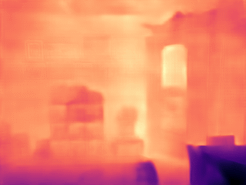} & \vspace{4pt}\includegraphics{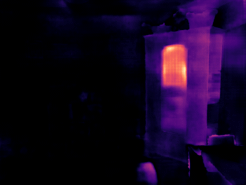} & \vspace{4pt}\includegraphics{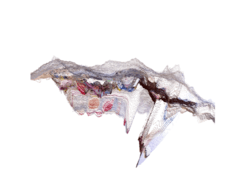} & \vspace{4pt}\includegraphics{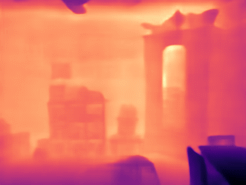} & \vspace{4pt}\includegraphics{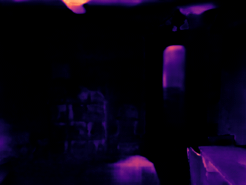} & \vspace{4pt}\includegraphics{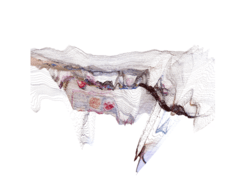} & \vspace{4pt}\includegraphics{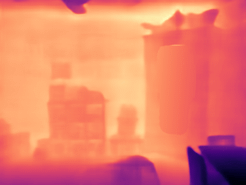} & \vspace{4pt}\includegraphics{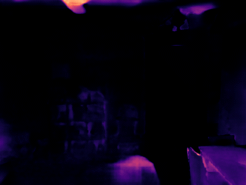} & \vspace{4pt}\includegraphics{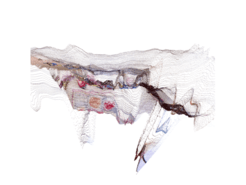} \\\cline{4-12}
& \small Color (RGB) & \multicolumn{1}{r}{\scriptsize vnl} & \vspace{4pt}\includegraphics{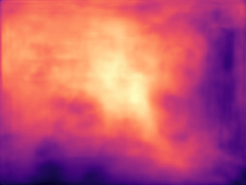} & \vspace{4pt}\includegraphics{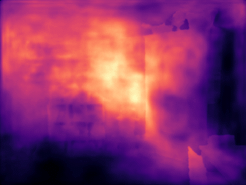} & \vspace{4pt}\includegraphics{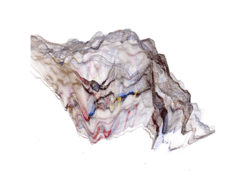} & \vspace{4pt}\includegraphics{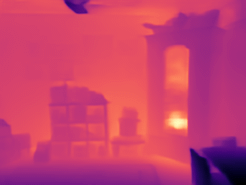} & \vspace{4pt}\includegraphics{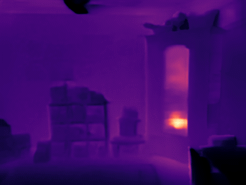} & \vspace{4pt}\includegraphics{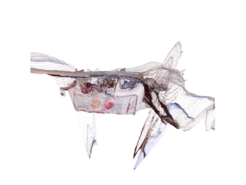} & \vspace{4pt}\includegraphics{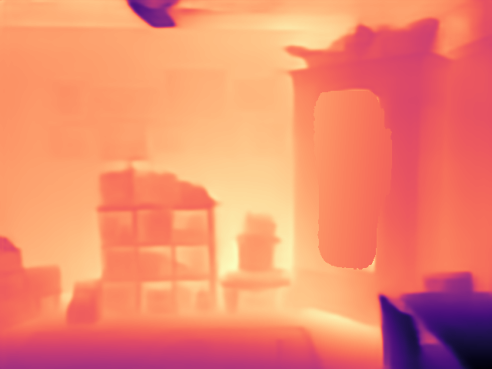} & \vspace{4pt}\includegraphics{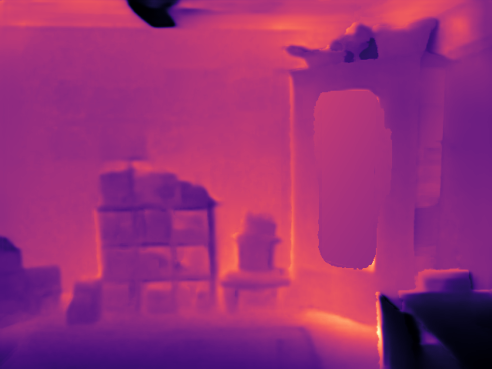} & \vspace{4pt}\includegraphics{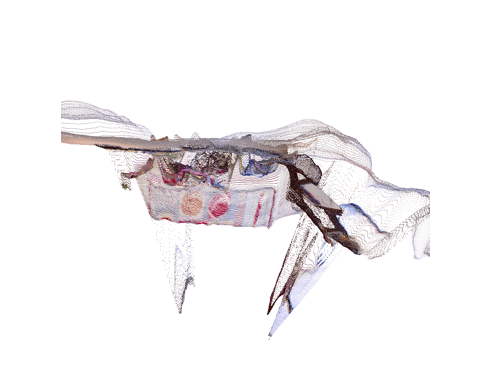} \\
 & & & \scriptsize Depth & \scriptsize Error & \scriptsize PC & \scriptsize Depth & \scriptsize Error & \scriptsize PC & \scriptsize Depth & \scriptsize Error & \scriptsize PC \\
 & & & \multicolumn{3}{c}{\small NYUv2-ref} & \multicolumn{3}{c}{\small NYUv2-raw} & \multicolumn{3}{c}{\small +Mirror3DNet}\\

\end{tabularx}
\begin{tikzpicture}[overlay,remember picture]
\node[ draw=black!60!green, line width=1pt, fit={(A1)(A2)($(A1.north west)+(+2pt,-0pt)$)($(A2.south east)+(-2pt,-0pt)$)}]{};
\node[ draw=blue!60!cyan, line width=1pt,fit={(B1)(B2)($(B1.north west)+(+2pt,-0pt)$)($(B2.south east)+(-2pt,-0pt)$)}]{};
\node[ draw=cyan, line width=1pt,fit={(C1)(C2)($(C1.north west)+(+2pt,-0pt)$)($(C2.south east)+(-2pt,-0pt)$)}]{};
\end{tikzpicture}

\vspace{10pt}

\begin{tabularx}{\textwidth}{Y@{\hspace{-13mm}} Y@{\hspace{1mm}} Y@{\hspace{0mm}} Y@{\hspace{1mm}} Y@{\hspace{1mm}} Y@{\hspace{2mm}} | Y@{\hspace{1mm}} Y@{\hspace{1mm}} Y@{\hspace{2mm}} | Y@{\hspace{1mm}} Y@{\hspace{1mm}} Y@{\hspace{0mm}} }
 & & & \multicolumn{3}{c}{\textcolor{black!60!green}{\small Ground Truth}} & \multicolumn{3}{c}{\textcolor{blue!60!cyan}{\small MP3D-mesh rendered depth}} & \multicolumn{3}{c}{\textcolor{cyan}{\small Pred-Refined}}\\
& \small Input Depth (D) &  & \vspace{4pt}\tikzmarknode{A1}{\includegraphics{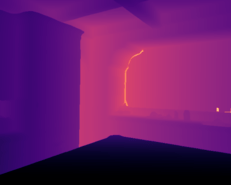}} & \vspace{4pt}\includegraphics{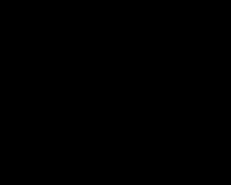} & \vspace{4pt}\tikzmarknode{A2}{\includegraphics{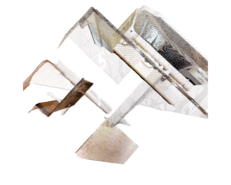}} & \vspace{4pt}\tikzmarknode{B1}{\includegraphics{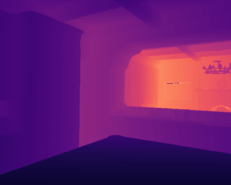}} & \vspace{4pt}\includegraphics{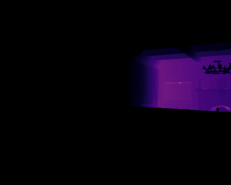} & \vspace{4pt}\tikzmarknode{B2}{\includegraphics{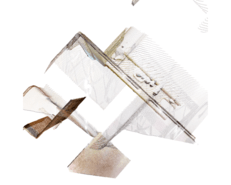}} & \vspace{4pt}\tikzmarknode{C1}{\includegraphics{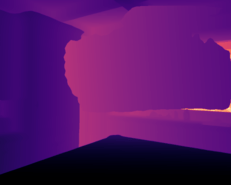}} & \vspace{4pt}\includegraphics{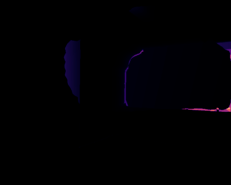} & \vspace{4pt}\tikzmarknode{C2}{\includegraphics{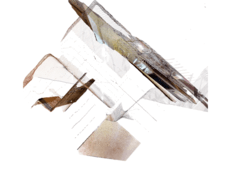}}
\\ &
\vspace{4pt}\includegraphics{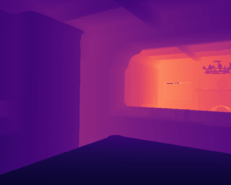} & \multicolumn{1}{r}{\scriptsize saic} & \vspace{4pt}\includegraphics{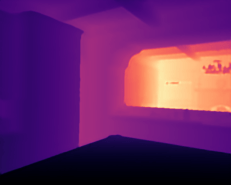} & \vspace{4pt}\includegraphics{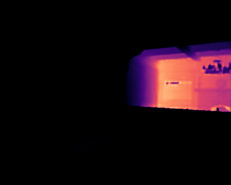} & \vspace{4pt}\includegraphics{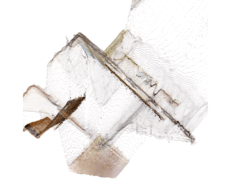} & \vspace{4pt}\includegraphics{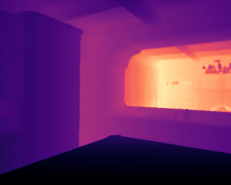} & \vspace{4pt}\includegraphics{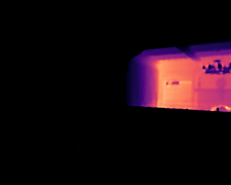} & \vspace{4pt}\includegraphics{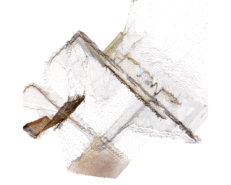} & \vspace{4pt}\includegraphics{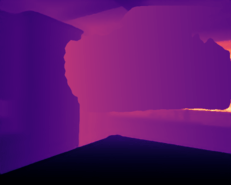} & \vspace{4pt}\includegraphics{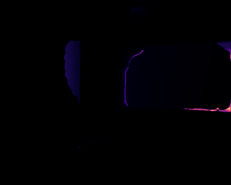} & \vspace{4pt}\includegraphics{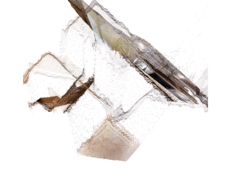} \\\cline{2-12} & \vspace{4pt}\includegraphics{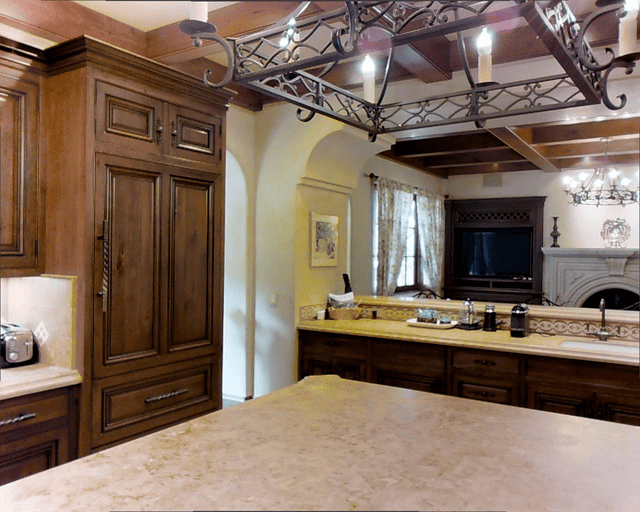} & \multicolumn{1}{r}{\scriptsize bts} & \vspace{4pt}\includegraphics{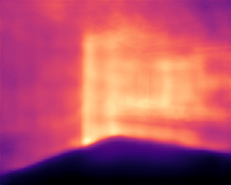} & \vspace{4pt}\includegraphics{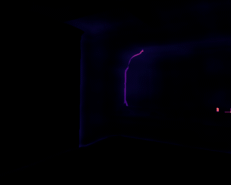} & \vspace{4pt}\includegraphics{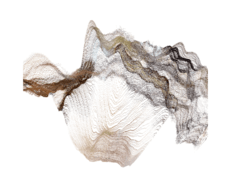} & \vspace{4pt}\includegraphics{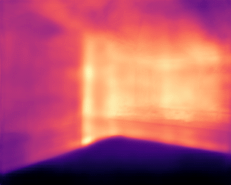} & \vspace{4pt}\includegraphics{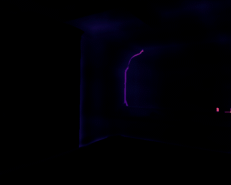} & \vspace{4pt}\includegraphics{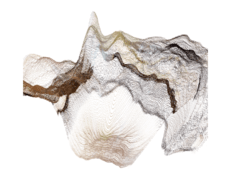} & \vspace{4pt}\includegraphics{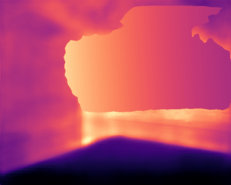} & \vspace{4pt}\includegraphics{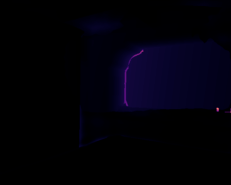} & \vspace{4pt}\includegraphics{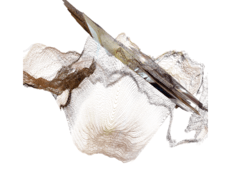} \\\cline{4-12}
& \small Color (RGB) & \multicolumn{1}{r}{\scriptsize vnl} & \vspace{4pt}\includegraphics{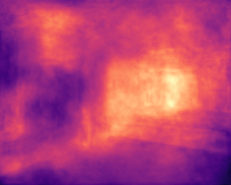} & \vspace{4pt}\includegraphics{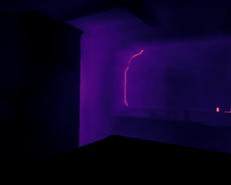} & \vspace{4pt}\includegraphics{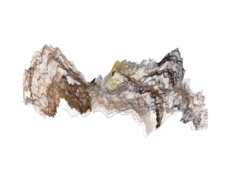} & \vspace{4pt}\includegraphics{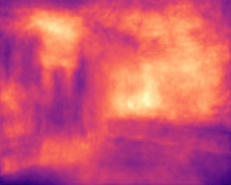} & \vspace{4pt}\includegraphics{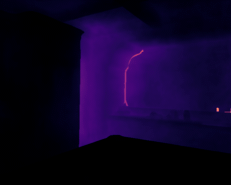} & \vspace{4pt}\includegraphics{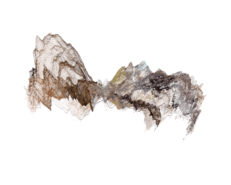} & \vspace{4pt}\includegraphics{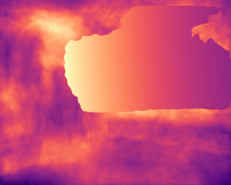} & \vspace{4pt}\includegraphics{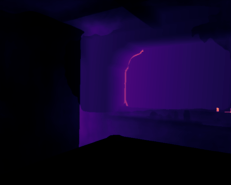} & \vspace{4pt}\includegraphics{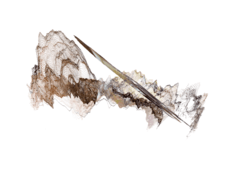} \\
 & & & \scriptsize Depth & \scriptsize Error & \scriptsize PC & \scriptsize Depth & \scriptsize Error & \scriptsize PC & \scriptsize Depth & \scriptsize Error & \scriptsize PC \\
 & & & \multicolumn{3}{c}{\small MP3D-mesh-ref} & \multicolumn{3}{c}{\small MP3D-mesh-raw} & \multicolumn{3}{c}{\small +Mirror3DNet}\\

\end{tabularx}
\begin{tikzpicture}[overlay,remember picture]
\node[ draw=black!60!green, line width=1pt, fit={(A1)(A2)($(A1.north west)+(+2pt,-0pt)$)($(A2.south east)+(-2pt,-0pt)$)}]{};
\node[ draw=blue!60!cyan, line width=1pt,fit={(B1)(B2)($(B1.north west)+(+2pt,-0pt)$)($(B2.south east)+(-2pt,-0pt)$)}]{};
\node[ draw=cyan, line width=1pt,fit={(C1)(C2)($(C1.north west)+(+2pt,-0pt)$)($(C2.south east)+(-2pt,-0pt)$)}]{};
\end{tikzpicture}
  
\caption{
Visualizations of depth frames, depth errors against ground truth (RMSE mapped to colormap), and resulting 3D point clouds (PC) for NYUv2 (top) and Matterport3D (bottom).
We compare the output depth from state-of-the-art RGB-based depth estimation approaches (bts~\cite{lee2019big}, vnl~\cite{yin2019enforcing}) and an RGB-based depth completion approach (saic~\cite{senushkin2020decoder}).
We contrast the outputs from these approaches when trained directly on the corrected datasets which leverage our 3D mirror plane annotations (\nyuref and \mpmeshref), against output of the approaches when trained on the uncorrected original datasets (\nyuraw and \mpmeshref), and against outputs after refinement using our \mnet module.
Overall, we observe that mirror depth errors are significantly reduced after applying \mnet, as seen by the reduced RMSE error in mirror regions, and the reduced prominence of depth outlier points.
}
\label{fig:exp-result-visualization}
\end{figure*}

\Cref{fig:exp-result-visualization} shows two qualitative comparison examples.
We include more qualitative examples in \Cref{sec:supp:qualitative}.
The top set of visualizations shows the improvements on \nyuraw and \nyuref data using our \mnet module to enhance results from several depth estimation and depth completion approaches.
We note that outlier depth points behind the mirror surface are significantly reduced after using \mnet (see depth RMSE images and top-down point cloud visualizations).
The improvements are consistent across input depth approach, whether RGB-based depth estimation is employed or RGBD-based depth completion is employed.
The bottom set of visualizations shows comparisons on the \mpmesh and \mpmeshref datasets.
This data exhibits fewer depth outliers due to the input being rendered from a 3D reconstructed mesh that incorporates manual mirror region correction.
Despite this, the \mnet module still improves depth accuracy on the mirror surface, in particular for RGB-based depth estimation approaches.

\subsection{Quantitative evaluation}

\begin{table}
\resizebox{\linewidth}{!}{
\begin{tabular}{@{}lll llll@{}}
\toprule
Input & Train Depth & Method & Seg-AP $\uparrow$ & $30^{\circ}$-AP $\uparrow$ \\
\midrule
RGB & * & \mnet & \best{0.464} & \best{0.256} \\
RGBD & mesh & \mnet & 0.416 & 0.255 \\
RGBD & mesh-ref & \mnet & 0.419 & 0.217 \\
RGBD & mesh & PlaneRCNN~\cite{liu2019planercnn} & 0.319 & 0.190 \\
RGBD & mesh-ref & PlaneRCNN~\cite{liu2019planercnn} & 0.334 & 0.220 \\
\bottomrule
\end{tabular}
}
\caption{
Evaluation of mirror mask segmentation and mirror normal prediction on Matterport3D validation set.
We observe that our \mnet module improves on mask segmentation, and normal estimation metrics.
}
\label{tb:mirror-prediction-results}
\end{table}

\mypara{Mirror 3D plane prediction.}
We first quantify the improvements introduced by our \mnet architecture over a baseline PlaneRCNN module for predicting the mirror 3D plane.
\Cref{tb:mirror-prediction-results} reports overall quantitative metrics for the Matterport3D dataset.
We see that the \mnet module leads to improved mirror mask segmentation and plane normal estimates.
The simplest \mnet module (without depth estimation module) gets the highest Seg-AP and $30^{\circ}$-AP.
We select a mirror anchor normal count of 10 and a mirror border width of 25 pixels based on results from ablation experiments (see \Cref{sec:supp:ablations}).

\begin{table}
\resizebox{\linewidth}{!}{
\begin{tabular}{@{}lllllllll@{}}
\toprule
& & & \multicolumn{3}{c}{RMSE $\downarrow$} & \multicolumn{3}{c}{SSIM $\uparrow$} \\\cmidrule(lr){4-6}\cmidrule(lr){7-9}
Input & Train & Method & Mirror & Other & All & Mirror & Other & All \\
\midrule
sensor-D &          * &                                             * & 1.272 &\best{0.022}& 0.424 & 0.634 &\best{0.997}& 0.946 \\
sensor-D &          * &                                         \mnet &\best{0.766}& 0.168 &\best{0.363}&\best{0.800}& 0.983 &\best{0.954}\\
\midrule
 RGBD &        ref &                                          saic~\cite{senushkin2020decoder} & 0.965 & 0.094 &\best{0.344}& 0.709 & 0.926 &\best{0.892}\\
 RGBD &        raw &                                          saic~\cite{senushkin2020decoder} & 1.245 &\best{0.070}& 0.433 & 0.641 &\best{0.930}& 0.887 \\
 RGBD &        raw &                                  saic~\cite{senushkin2020decoder} + \mnet &\best{0.749}& 0.208 & 0.378 &\best{0.803}& 0.913 & 0.891 \\
\midrule
  RGB &        ref &                                           BTS~\cite{lee2019big} &\best{0.551}& 0.466 &\best{0.494}& 0.793 & 0.798 & 0.786 \\
  RGB &        ref &                                           VNL~\cite{yin2019enforcing} & 5.956 & 5.046 & 5.204 & 0.270 & 0.291 & 0.286 \\
  RGB &        raw &                                           BTS~\cite{lee2019big} & 0.903 &\best{0.424}& 0.583 & 0.703 &\best{0.845}& 0.812 \\
  RGB &        raw &                                           VNL~\cite{yin2019enforcing} & 4.216 & 2.598 & 3.007 & 0.365 & 0.594 & 0.554 \\
  RGB &        raw &                                   BTS~\cite{lee2019big} + \mnet & 0.567 & 0.517 & 0.577 &\best{0.844}& 0.841 &\best{0.827}\\
  RGB &        raw &                                   VNL~\cite{yin2019enforcing} + \mnet & 3.245 & 2.629 & 2.822 & 0.536 & 0.591 & 0.573 \\
\bottomrule\end{tabular}}
\caption{Depth prediction evaluation on \nyuref test set, for images containing mirrors.}
\label{tb:nyuv2-refined-results}
\end{table}

\begin{table}
\resizebox{\linewidth}{!}{
\begin{tabular}{@{}lllllllll@{}}
\toprule
& & & \multicolumn{3}{c}{RMSE $\downarrow$} & \multicolumn{3}{c}{SSIM $\uparrow$} \\\cmidrule(lr){4-6}\cmidrule(lr){7-9}
Input & Train & Method & Mirror & Other & All & Mirror & Other & All \\
\midrule
sensor-D &          * &                                             * &\best{0.000}&\best{0.000}&\best{0.000}&\best{1.000}&\best{1.000}&\best{1.000}\\
sensor-D &          * &                                         \mnet & 0.784 & 0.158 & 0.369 & 0.772 & 0.986 & 0.958 \\
\midrule
 RGBD &        ref &                                          saic~\cite{senushkin2020decoder} & 0.429 & 0.090 & 0.205 & 0.815 & 0.928 & 0.907 \\
 RGBD &        raw &                                          saic~\cite{senushkin2020decoder} &\best{0.107}&\best{0.062}&\best{0.067}&\best{0.890}&\best{0.932}&\best{0.924}\\
 RGBD &        raw &                                  saic~\cite{senushkin2020decoder} + \mnet & 0.825 & 0.204 & 0.398 & 0.734 & 0.915 & 0.888 \\
\midrule
  RGB &        ref &                                           BTS~\cite{lee2019big} & 1.093 & 0.466 & 0.631 & 0.625 & 0.799 & 0.768 \\
  RGB &        ref &                                           VNL~\cite{yin2019enforcing} & 4.915 & 5.046 & 5.052 & 0.306 & 0.291 & 0.294 \\
  RGB &        raw &                                           BTS~\cite{lee2019big} &\best{0.971}&\best{0.423}&\best{0.529}& 0.618 &\best{0.846}&\best{0.813}\\
  RGB &        raw &                                           VNL~\cite{yin2019enforcing} & 3.316 & 2.597 & 2.794 & 0.413 & 0.594 & 0.567 \\
  RGB &        raw &                                   BTS~\cite{lee2019big} + \mnet & 1.155 & 0.516 & 0.676 &\best{0.646}& 0.841 & 0.810 \\
  RGB &        raw &                                   VNL~\cite{yin2019enforcing} + \mnet & 2.584 & 2.628 & 2.693 & 0.501 & 0.592 & 0.576 \\
\bottomrule
\end{tabular}
}
\caption{Depth prediction evaluation on \nyuraw test set, for images containing mirrors.}
\label{tb:nyuv2-raw-results}
\end{table}

\begin{table}
\resizebox{\linewidth}{!}{
\begin{tabular}{@{}lllllllll@{}}
\toprule
& & & \multicolumn{3}{c}{RMSE $\downarrow$} & \multicolumn{3}{c}{SSIM $\uparrow$} \\\cmidrule(lr){4-6}\cmidrule(lr){7-9}
Input & Train & Method & Mirror & Other & All & Mirror & Other & All \\
\midrule
sensor-D &          * &                                             * & 2.676 & 0.959 & 1.325 & 0.287 & 0.787 & 0.707 \\
mesh-D &          * &                                             * & 0.661 &\best{0.000}&\best{0.155}& 0.841 &\best{1.000}&\best{0.980}\\
sensor-D &          * &                                         \mnet & 1.249 & 0.949 & 1.017 & 0.738 & 0.785 & 0.772 \\
mesh-D &          * &                                         \mnet &\best{0.477}& 0.156 & 0.226 &\best{0.893}& 0.979 & 0.964 \\
\midrule
 RGBD &   mesh-ref &                                          saic~\cite{senushkin2020decoder} &\best{0.312}&\best{0.098}&\best{0.144}& 0.891 &\best{0.961}&\best{0.950}\\
 RGBD &       mesh &                                          saic~\cite{senushkin2020decoder} & 0.475 & 0.111 & 0.183 & 0.867 & 0.956 & 0.943 \\
 RGBD &       mesh &                                  saic~\cite{senushkin2020decoder} + \mnet & 0.433 & 0.221 & 0.265 &\best{0.901}& 0.939 & 0.928 \\
\midrule
  RGB &   mesh-ref &                                           BTS~\cite{lee2019big} &\best{0.581}&\best{0.603}&\best{0.628}& 0.837 &\best{0.800}&\best{0.792}\\
  RGB &   mesh-ref &                                           VNL~\cite{yin2019enforcing} & 1.478 & 1.439 & 1.435 & 0.700 & 0.679 & 0.677 \\
  RGB &       mesh &                                           BTS~\cite{lee2019big} & 0.664 & 0.634 & 0.676 & 0.797 & 0.790 & 0.775 \\
  RGB &       mesh &                                           VNL~\cite{yin2019enforcing} & 1.459 & 1.438 & 1.432 & 0.702 & 0.678 & 0.677 \\
  RGB &       mesh &                                   BTS~\cite{lee2019big} + \mnet & 0.685 & 0.675 & 0.698 &\best{0.859}& 0.783 & 0.782 \\
  RGB &       mesh &                                   VNL~\cite{yin2019enforcing} + \mnet & 1.516 & 1.449 & 1.446 & 0.740 & 0.680 & 0.687 \\
\bottomrule
\end{tabular}
}
\caption{Depth prediction evaluation on \mpmeshref test set, for images containing mirrors.}
\label{tb:m3d-refined-results}
\end{table}

\mypara{Mirror depth refinement.}
We conducted a series of experiments on the NYUv2 and Matterport3D test sets to quantify the improvements in depth value prediction offered by our \mnet architecture, and summarize overall trends.
Please refer to \Cref{sec:supp:quantitative} for additional results and metrics.
We used three different datasets as the ground truth for the purposes of the evaluation reported here: \nyuref, \nyuraw, and \mpmeshref.
Results with the first dataset (\nyuref) treated as the ground truth are in \Cref{tb:nyuv2-refined-results}.
These results show how far from the `correct ground truth' the various depth estimation and prediction methods are, and how much of an improvement \mnet can provide.
We note that methods trained on the corrected \nyuref depth have significantly better performance on mirror area depth prediction.
The \mnet module trained on Matterport3D can generalize to NYUv2 and helps improve depth accuracy for mirrors significantly when applied to methods trained on \nyuraw, which would be the practical input at capture time.

We contrast the above set of results against the results reported in \Cref{tb:nyuv2-raw-results}, where the ground truth is assumed to be the original, uncorrected \nyuraw dataset.
We make the observation that evaluating on raw depth as ground truth gives a different (and inaccurate) ranking of methods compared to the corrected ground truth in \nyuref.
Not surprisingly, the ranking on \nyuraw gives an edge to the methods that were trained on the raw depth.
This is however, a misleading result, as the `assumed to be correct ground truth' is highly inaccurate for mirror regions.

Lastly, in \Cref{tb:m3d-refined-results} we see results using the \mpmeshref dataset as the ground truth.
The overall trends remain the same, but we note that on this data there are less pronounced differences in performance (i.e. it is still better to train on refined depth).
We hypothesize that this is due to mesh cleanup and post-processing relying on human-provided annotation of some reflective and transparent surfaces.
In other words, the Matterport reconstruction pipeline that was used to produce the mesh from which \mpmesh data is rendered already incorporates a degree of mirror surface correction.
While this annotation is not explicitly specified by the Matterport3D dataset, the mesh reconstruction itself makes use of it, and the rendered mesh depth will thus have cleaner depth than without human intervention.
Our goal with \mnet is to automate this depth refinement so that human intervention is not required at capture time or mesh reconstruction post-processing.

\subsection{Limitations and future work}

Our mirror depth refinement approach is a simple first step but it is subject to several limitations that suggest future work directions.
We assume planar mirrors and use a fixed border width to estimate the mirror depth offset.
Both assumptions can be addressed using more advanced mirror detection approaches.
Moreover, in this paper we only focused on using 3D mirror plane estimates to refine depth values at the mirror surface.
Observations of reflected objects in mirrors can be leveraged to further improve reconstruction of other surfaces beyond the mirror.
The estimated 3D mirror planes can also be used to create more realistic rendered visuals from 3D reconstructions, by simulating reflections and light propagation from the mirror surfaces.

\section{Conclusion}

In this paper, we tackled the problem of 3D mirror plane prediction and mirror depth refinement.
We created Mirror3D: a large-scale dataset of 3D mirror plane annotations based on three popular RGBD datasets which we use to obtain corrected ground truth depth on mirror surfaces.
Using this data, we develop \mnet: a mirror depth refinement architecture that can be used to refine depth estimation or depth completion output.
Our experiments show that mirror depth errors in popular RGBD datasets are prevalent, and that treating existing depth data as ground truth can misrepresent depth prediction method performance.
Moreover, we show that our \mnet architecture helps to improve mirror depth estimates from depth estimation and depth completion approaches, significantly mitigating 3D reconstruction artifacts due to mirror surfaces.

\vspace{1em}
\mypara{Acknowledgements.}
Many thanks to Yiming Zhang for help in developing annotation verification and visualization tools, and the dataset website.
We also thank Hanxiao Jiang, Yongsen Mao and Yiming Zhang for their help with dataset annotation.
We thank Shitao Tang for helpful early conversations on the Mirror3DNet neural architecture used in our work.
We are grateful to the anonymous reviewers for their helpful suggestions.
This research was enabled in part by support provided by \href{www.westgrid.ca}{WestGrid} and \href{www.computecanada.ca}{Compute Canada}.
Angel X. Chang is supported by a Canada CIFAR AI Chair, and Manolis Savva by a Canada Research Chair and NSERC Discovery Grant.

\appendix
\begin{table*}[!htp]
\centering
\begin{tabular}{@{}lrrrr rrrr rrrr@{}}
\toprule
& \multicolumn{4}{c}{NYUv2} & \multicolumn{4}{c}{Matterport3D} & \multicolumn{4}{c}{ScanNet}
\\\cmidrule(lr){2-5}\cmidrule(lr){6-9}\cmidrule(lr){10-13}
 & Train & Val & Test & \textbf{Total} & Train & Val & Test & \textbf{Total} & Train & Val & Test & \textbf{Total} \\
\midrule
Mirror planes & $73$ & -- & $58$ & $131$ & $3@435$ & $340$ & $887$ & $4@662$ & $1@619$ & $241$ & $358$ & $2@218$ \\
RGBD images & $73$ & -- & $52$ & $125$ & $2@851$ & $282$ & $649$ & $3@782$ & $1@409$ & $229$ & $349$ & $1@987$ \\
RGBD panoramas & -- & -- & -- & -- & $1@840$ & $198$ & $430$ & $2@468$ & -- & -- & -- & -- \\
3D scenes & $54$ & -- & $42$ & $96$ & $56$ & $8$ & $15$ & $79$ & $192$ & $35$ & $55$ & $282$ \\
\bottomrule
\end{tabular}
\caption{Breakdown of number of mirror planes, RGBD images, panoramas, and 3D scenes across splits for \DATASET.}
\label{tab:detailed-statistics}
\end{table*}
\section{Mirror3D Dataset Statistics}
\label{sec:supp:dataset}
Here, we provide additional summary statistics for the Mirror3D dataset that we constructed based on RGBD frames from NYUv2~\cite{nyuv2}, Matterport3D~\cite{matterport3d}, and ScanNet~\cite{Scannet}.
We also provide several examples of annotated 3D mirror planes to illustrate the diversity of scenarios in which they occur.

\begin{figure*}
  \subfigure[NYUv2]{
    \begin{minipage}[t]{0.23\linewidth}
      \includegraphics[width=\textwidth]{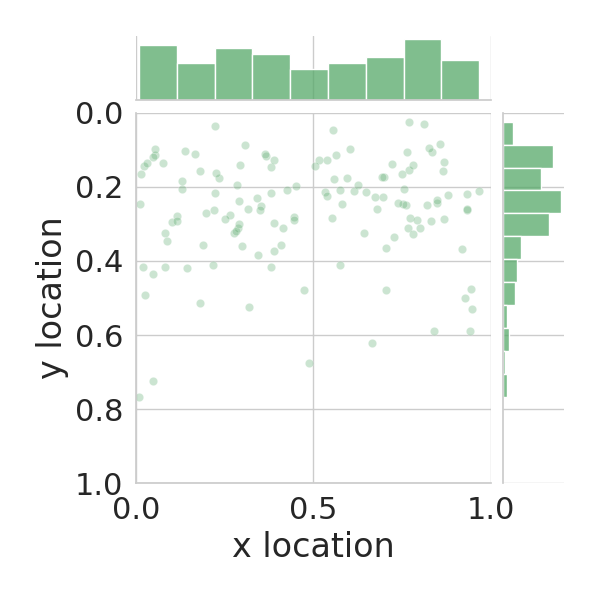}
    \end{minipage}
  }
  \subfigure[Matterport3D]{
    \begin{minipage}[t]{0.23\linewidth}
      \includegraphics[width=\textwidth]{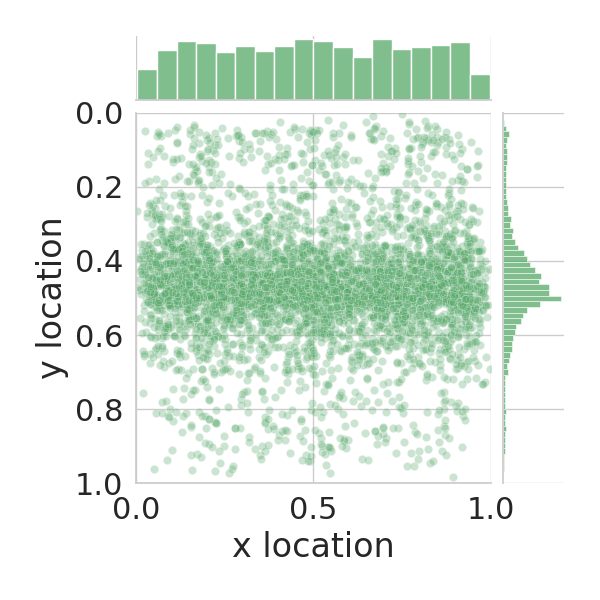}
    \end{minipage}
  }
  \subfigure[ScanNet]{
    \begin{minipage}[t]{0.23\linewidth}
      \includegraphics[width=\textwidth]{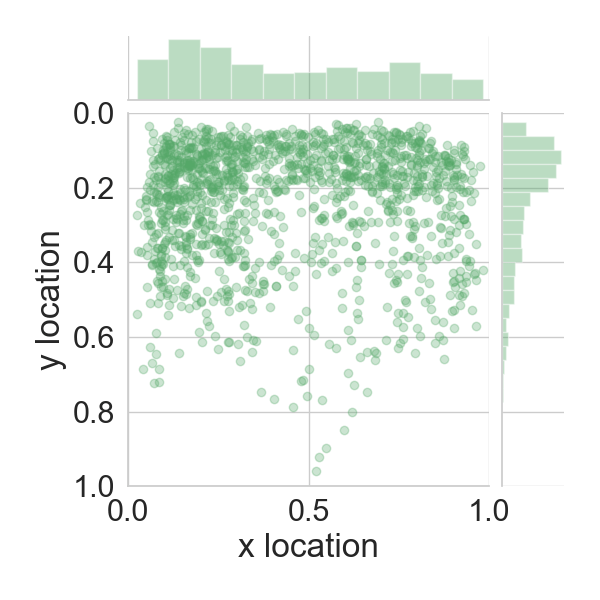}
    \end{minipage}
  }
  \subfigure[Overall]{
    \begin{minipage}[t]{0.23\linewidth}
      \includegraphics[width=\textwidth]{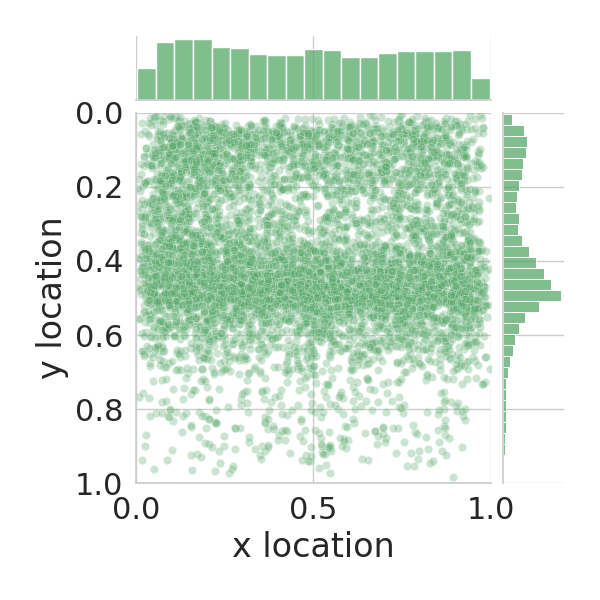}
    \end{minipage}
  }
  \caption{Distribution of mirror mask centroids in image plane, plotted by source RGBD dataset (a through c) and overall (d). Each point corresponds to a mirror instance mask. Matterport3D is collected using a static tripod-based sensor and exhibits strong vertical centering of the mirror centroid point. In contrast, ScanNet and NYUv2 were collected with hand-held depth sensors which were frequently held in a slightly downwards facing angle, and avoided capturing mirrors `head on'. The overall dataset therefore exhibits a somewhat bimodal distribution of mirror centroid along the vertical axis.}
 \label{fig:mirror-location-dis}
\end{figure*}

\mypara{Train/test/val split.}
\Cref{tab:detailed-statistics} provides the detailed breakdown by number of planes, images, panoramas, and scenes across train/val/test for the Mirror3D dataset.
Note that NYUv2 does not have a validation split.

\mypara{Mirror in-frame location distribution.}
\Cref{fig:mirror-location-dis} plots the distribution of mirror mask centroids within the image plane per source RGBD dataset and also overall for the entirety of Mirror3D.
Overall, mirror centroid points tend to cluster around the upper part of the image.
This is expected as mirrors are usually placed at human eye level.
We do note clear differences in the distributions for different datasets, which combine to form an overall distribution with fairly dense coverage of the image frame.

\begin{figure*}
  \centering
  \subfigure[NYUv2]{
    \begin{minipage}[t]{0.23\linewidth}
      \includegraphics[width=\textwidth]{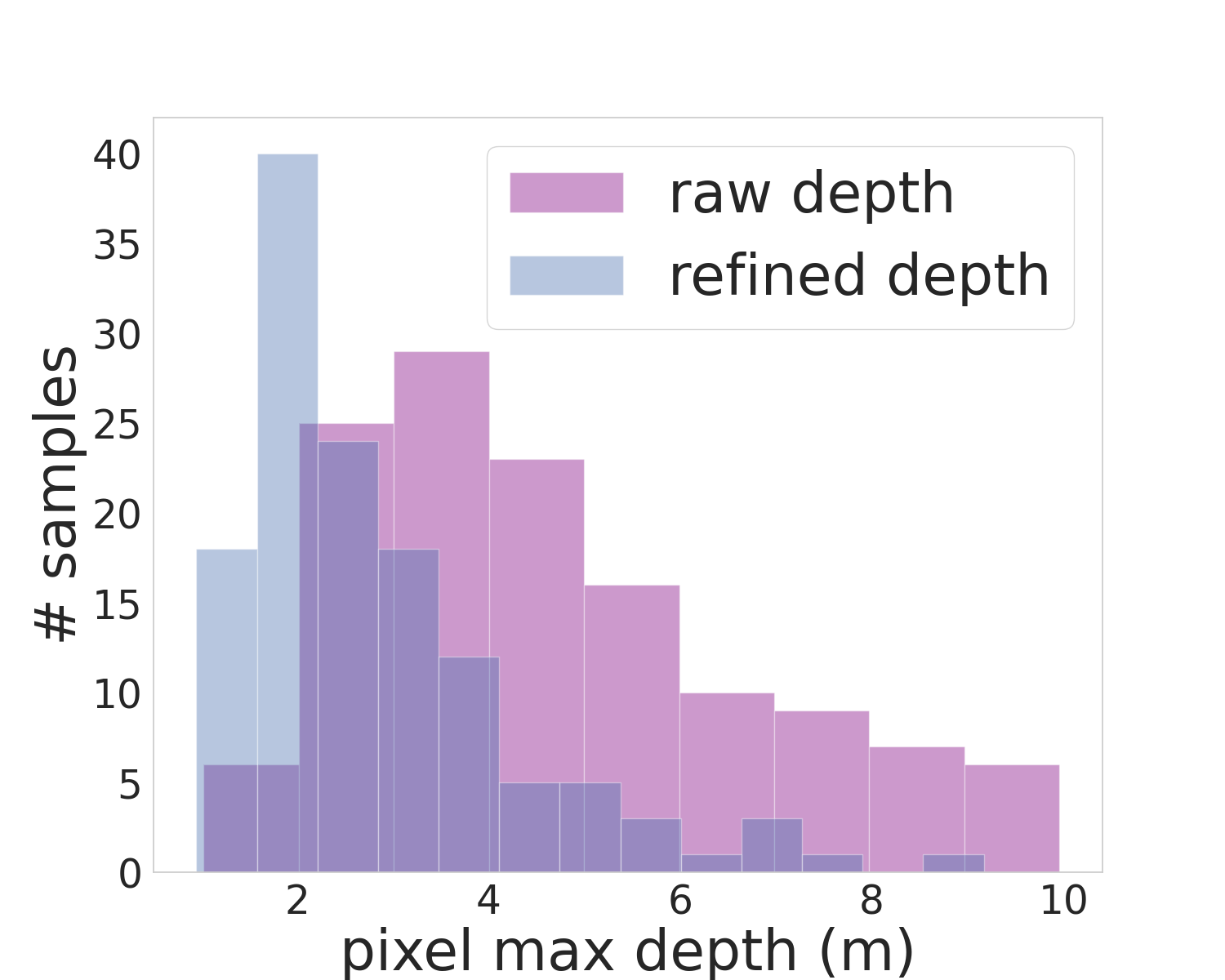}
    \end{minipage}
  }
  \subfigure[Matterport3D]{
    \begin{minipage}[t]{0.23\linewidth}
      \includegraphics[width=\textwidth]{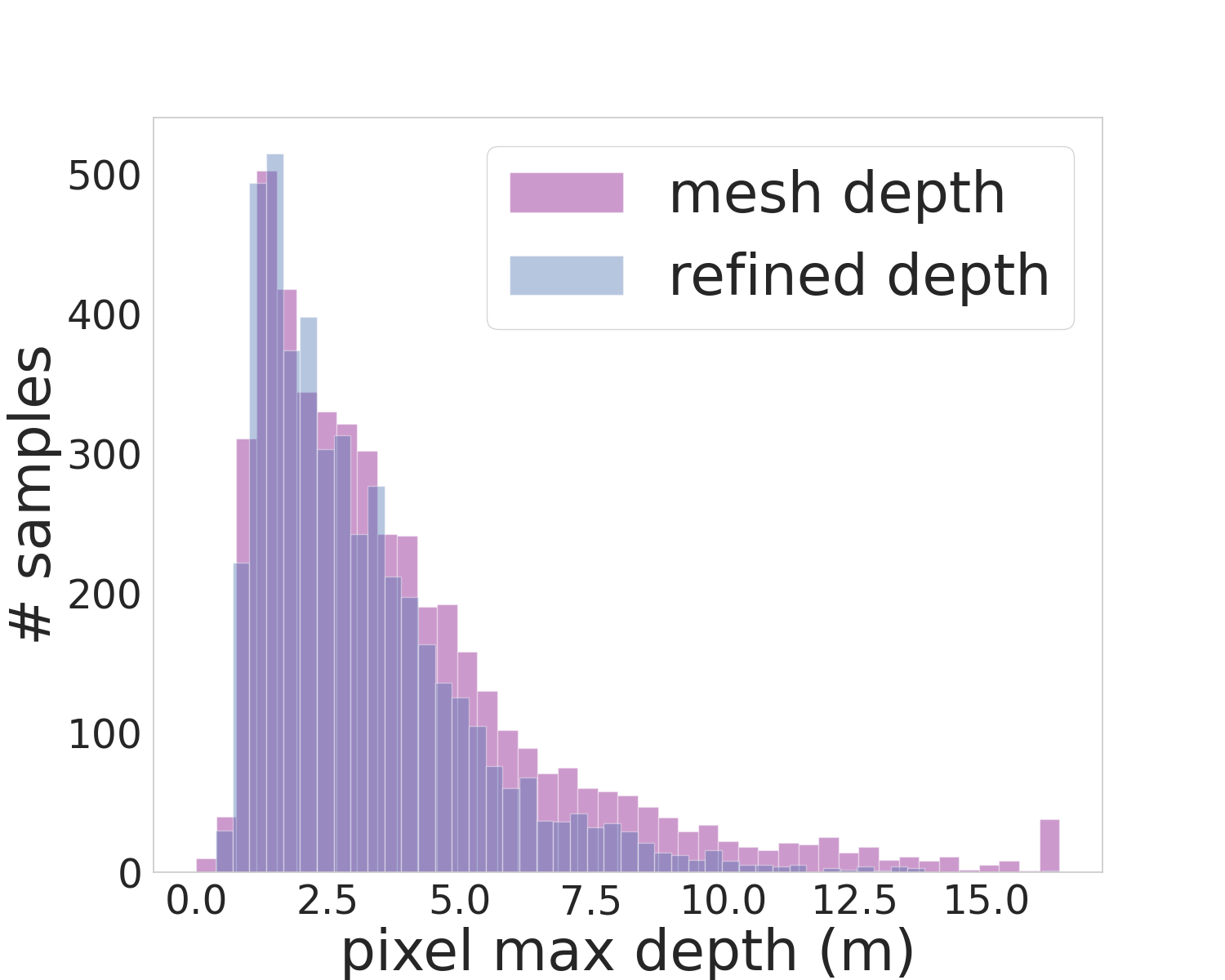}
    \end{minipage}
  }
  \subfigure[ScanNet]{
    \begin{minipage}[t]{0.23\linewidth}
      \includegraphics[width=\textwidth]{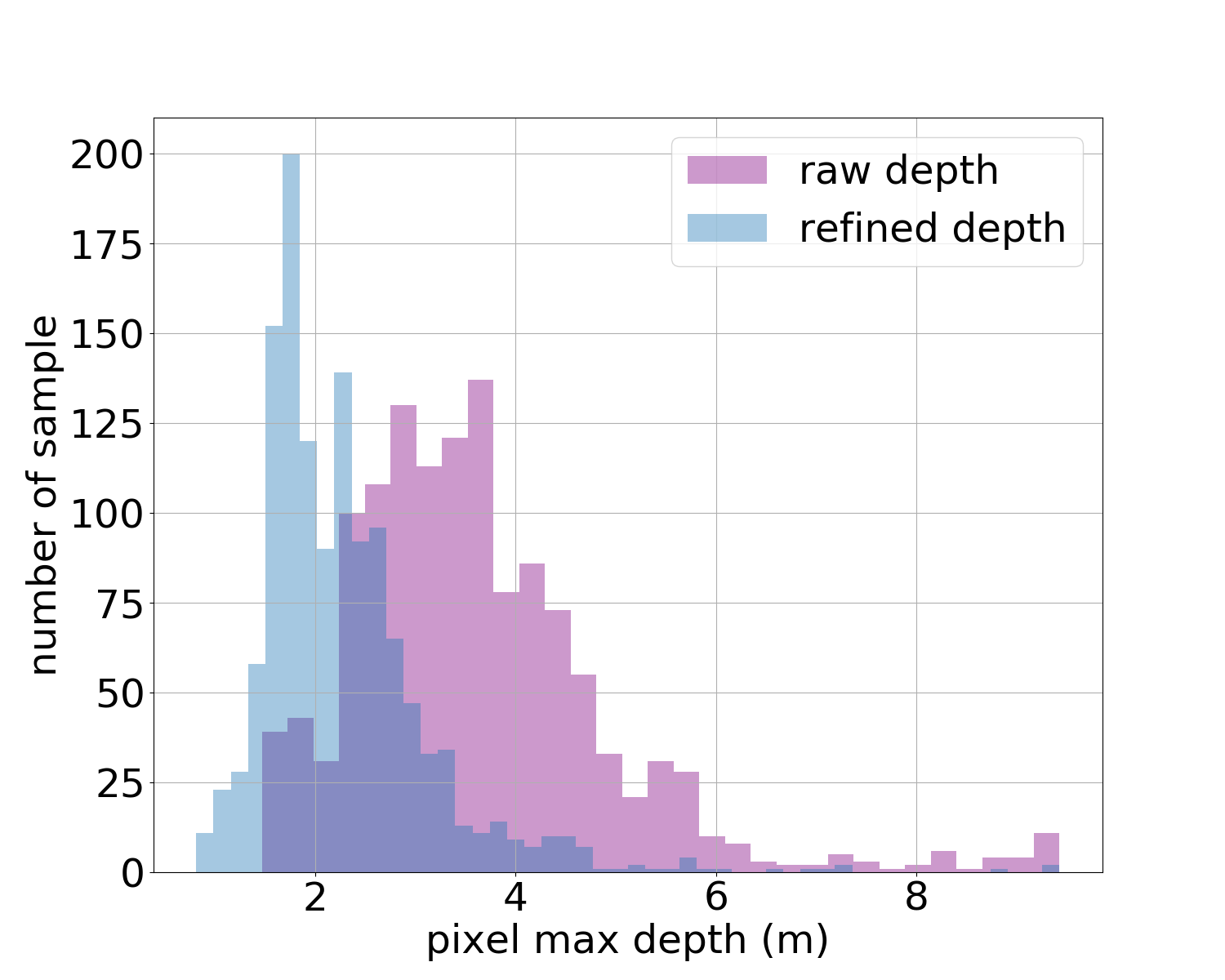}
    \end{minipage}
  }
  \subfigure[Overall]{
    \begin{minipage}[t]{0.23\linewidth}
      \includegraphics[width=\textwidth]{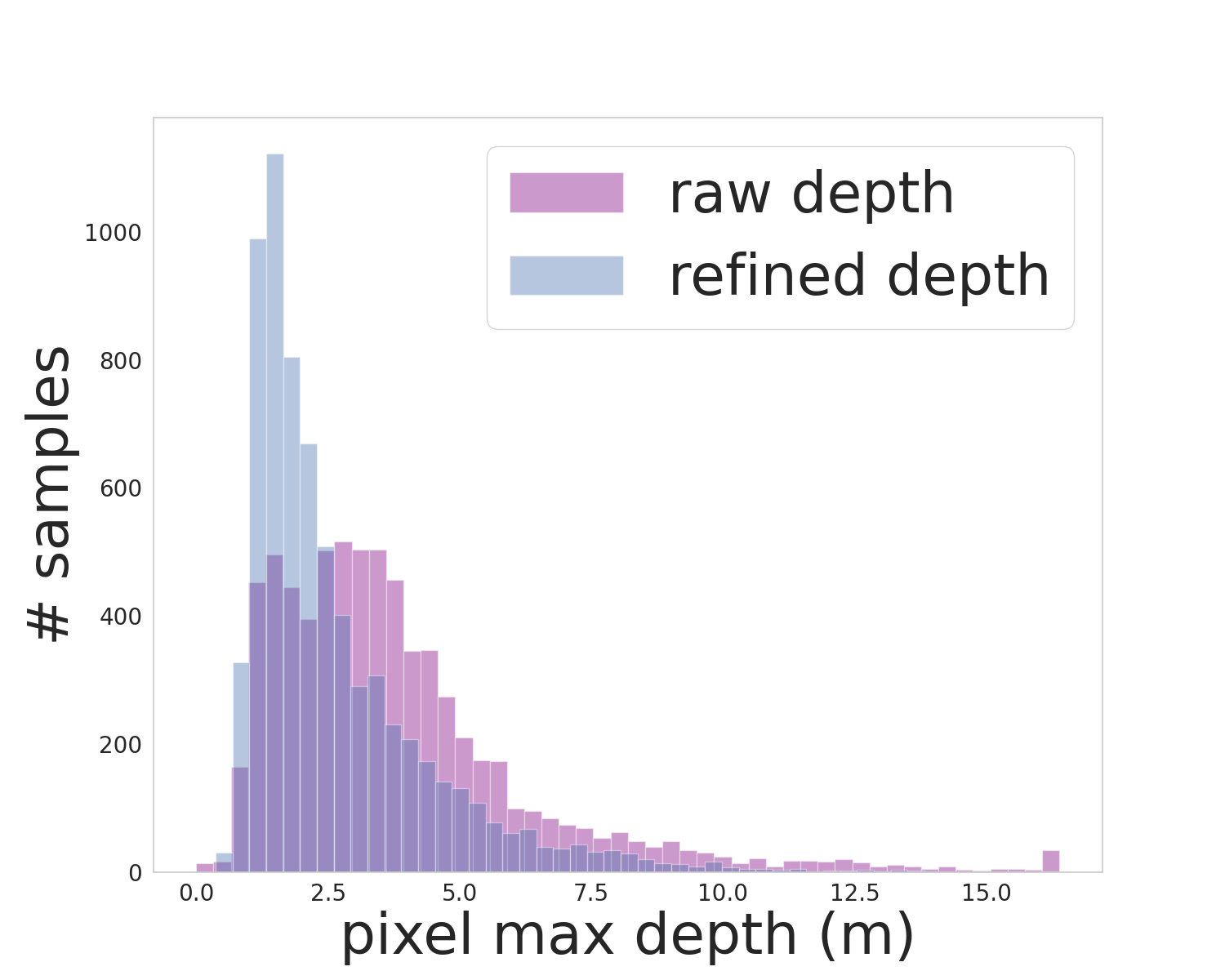}
    \end{minipage}
  }

  \caption{Distribution of the maximum depth value per frame by source RGBD dataset (a through c) and overall (d) for both the raw and refined depth in images with mirrors. After correcting the noisy depth values using our 3D mirror plane annotations, the maximum depth is consistently reduced for all datasets showing the reduction of depth outlier points at far distances (usually associated with `floating points' behind mirror surfaces).}
  \label{fig:mirror-area-depth-dis}
\end{figure*}

\mypara{Mirror depth distribution.}
The distance from the camera of the mirror surface is an important characteristic influencing the reliability of the surrounding depth points, and correspondingly the challenge of performing depth estimation or completion on images including mirrors.
\Cref{fig:mirror-area-depth-dis} plots histograms of the maximum mirror depth value for each mirror instance in our dataset, again broken down by source RGBD dataset and overall.
We show histograms for both the raw original depth values and the corrected depth values after annotation of the 3D mirror planes.
We note that the corrected depth value distributions are tighter, with fewer outliers at larger depth values.
This indicates that the annotation mitigates many of the `floating depth noise' artifacts in the raw data.

\begin{figure*}
  \centering
  \subfigure[NYUv2]{
    \begin{minipage}[t]{0.23\linewidth}
      \includegraphics[width=\textwidth]{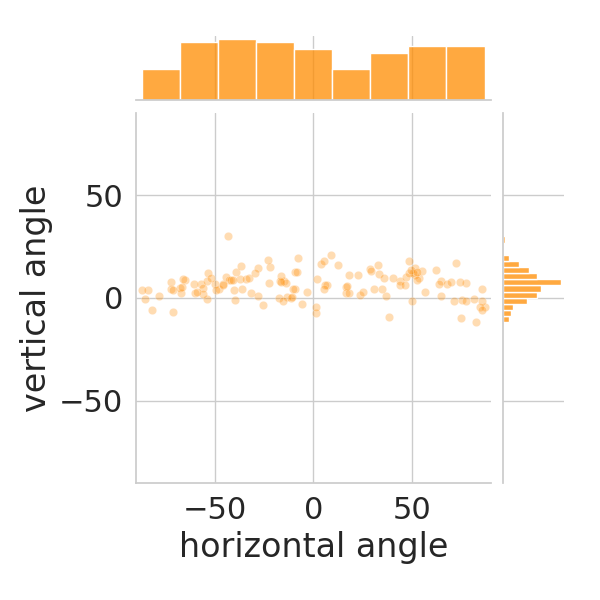}
    \end{minipage}
  }
  \subfigure[Matterport3D]{
    \begin{minipage}[t]{0.23\linewidth}
      \includegraphics[width=\textwidth]{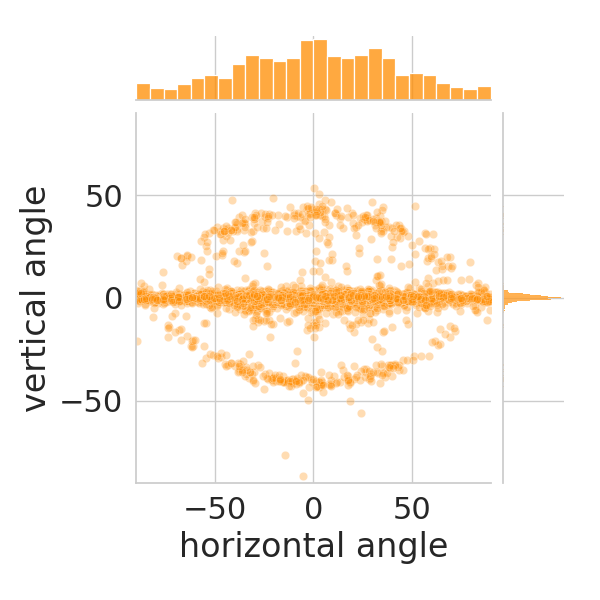}
    \end{minipage}
  }
  \subfigure[ScanNet]{
    \begin{minipage}[t]{0.23\linewidth}
      \centering
      \includegraphics[width=\textwidth]{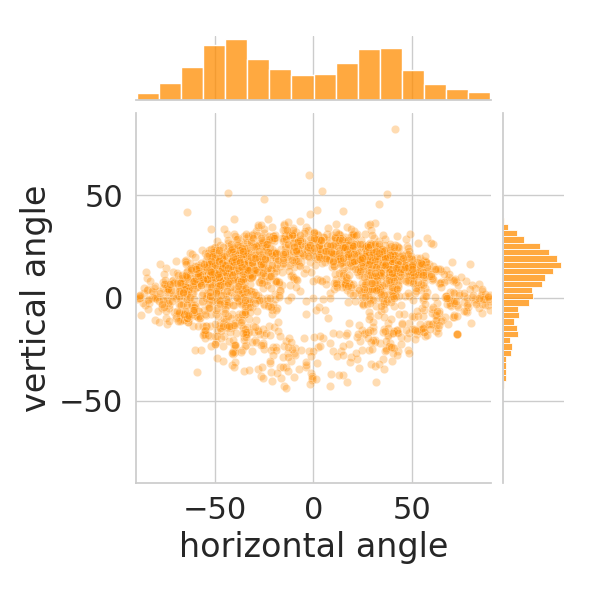}
    \end{minipage}
  }
  \subfigure[Overall]{
    \begin{minipage}[t]{0.23\linewidth}
      \includegraphics[width=\textwidth]{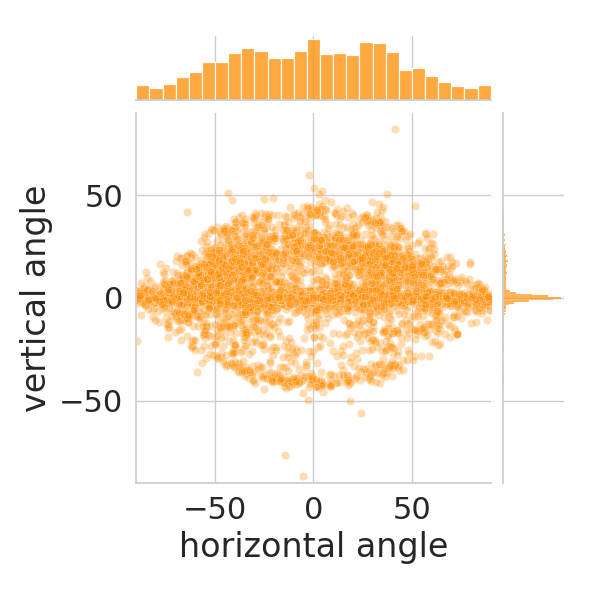}
    \end{minipage}
  }
  \caption{Distribution of mirror normals by source RGBD dataset (a through c) and overall (d). Each point corresponds to the normal of a 3D mirror plane instance.  The overall distribution reveals that many normals are at angles that are close to `facing towards' the camera with small horizontal deviations. We note interesting differences between the distributions for different source datasets.  The Matterport3D frames exhibit a tight horizontal vertical angle distribution, which we hypothesize is due to the tripod-based sensor. This is in contrast to both NYUv2 and ScanNet which were both captured with handheld devices and exhibit broader distributions.}
  \label{fig:mirror-normal-dis}
\end{figure*}

\mypara{Mirror normal distribution.}
In \Cref{fig:mirror-normal-dis} we plot the distribution of mirror 3D plane normal direction relative to the camera viewing direction.
Most mirror normals are clustered around facing the camera center and have relatively small angle deviations from that orientation.
There are again interesting differences between the source dataset distributions that we attribute to the use of tripod-based equipment with fixed tilt angles (Matterport3D) vs the use of handheld scanning sensors (NYUv2 and ScanNet).

\begin{figure*}
  \centering
  \subfigure[NYUv2]{
    \begin{minipage}[t]{0.23\linewidth}
      \includegraphics[width=\textwidth]{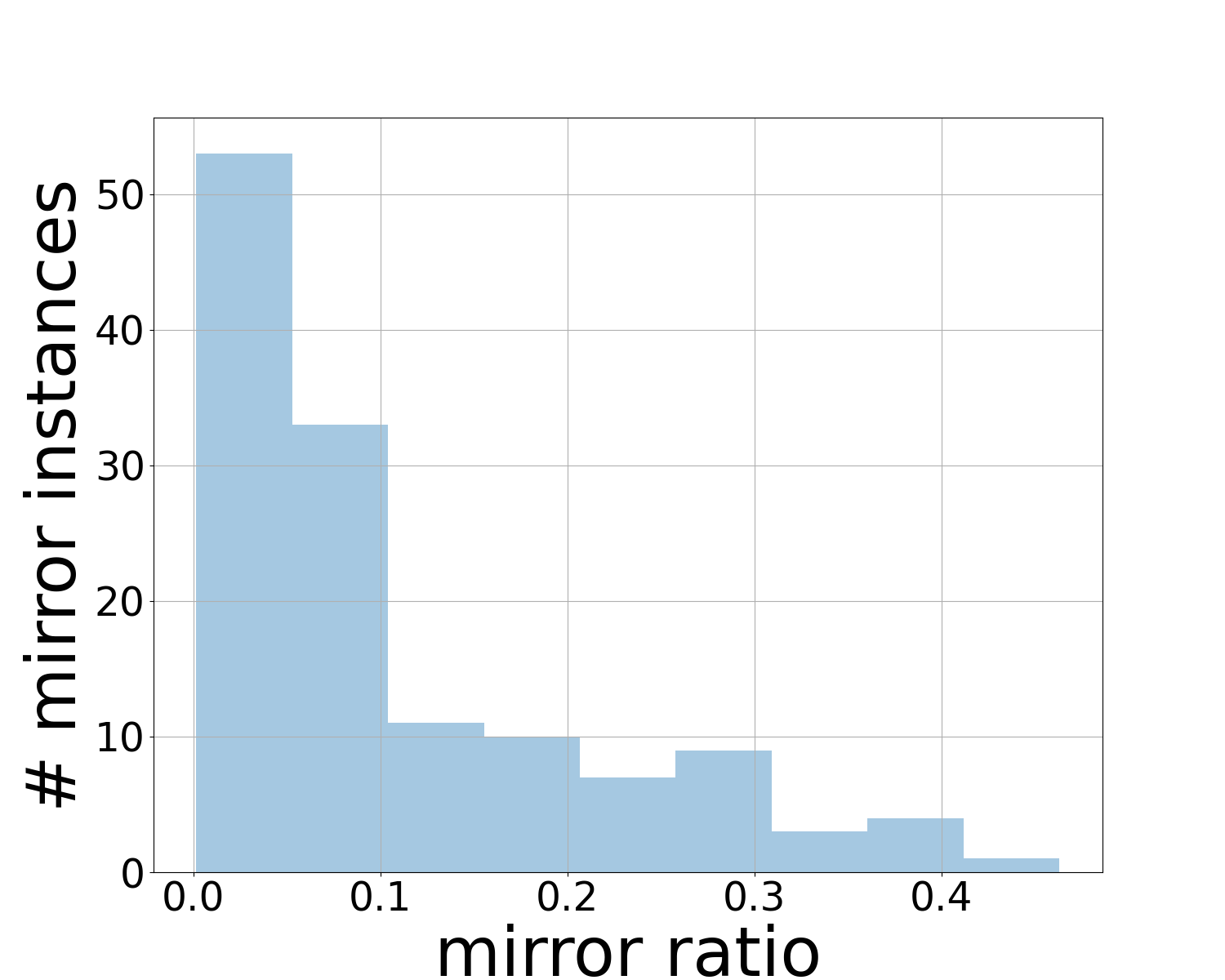}
    \end{minipage}
  }
  \subfigure[Matterport3D]{
    \begin{minipage}[t]{0.23\linewidth}
      \includegraphics[width=\textwidth]{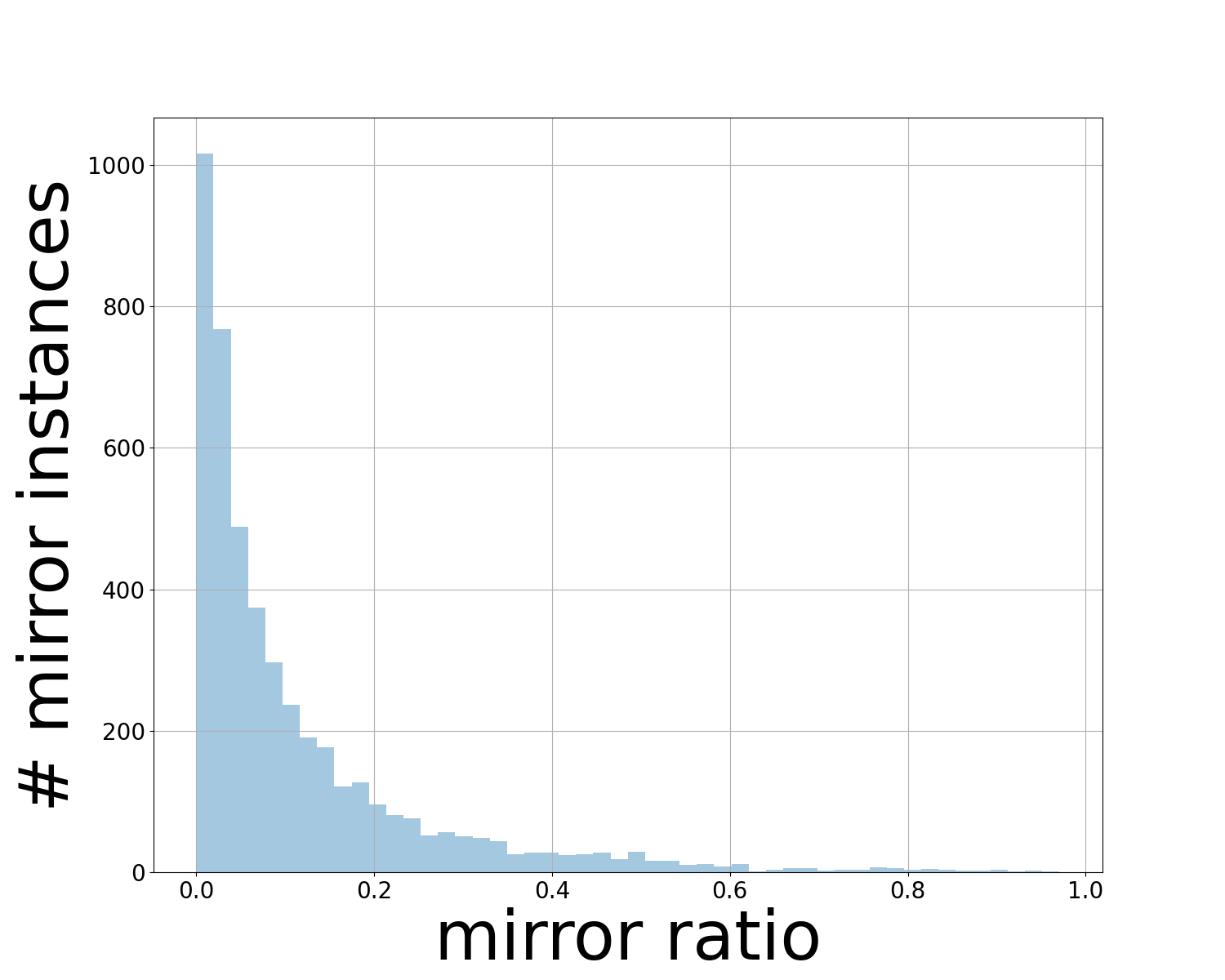}
    \end{minipage}
  }
  \subfigure[ScanNet]{
    \begin{minipage}[t]{0.23\linewidth}
      \centering
      \includegraphics[width=\textwidth]{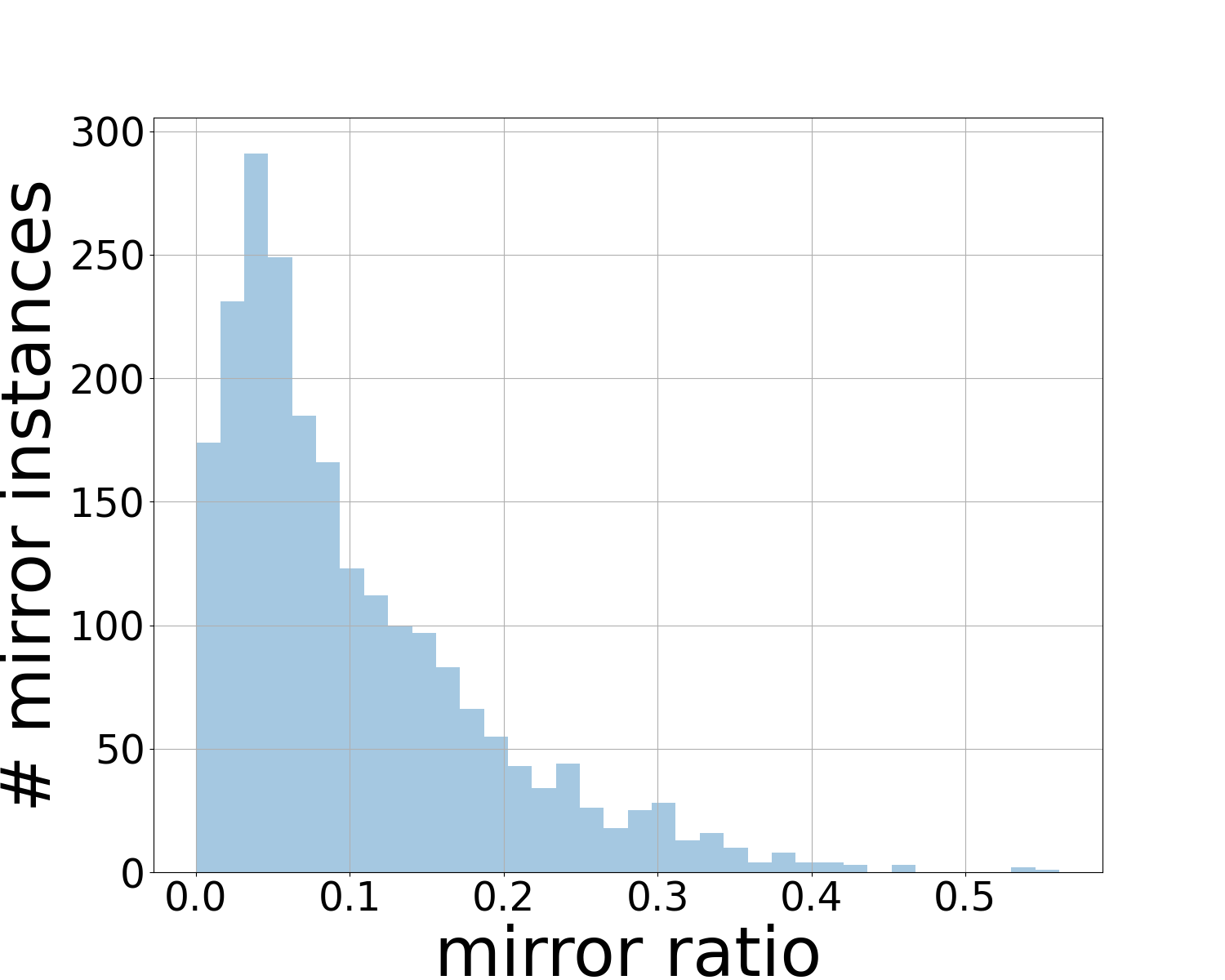}
    \end{minipage}
  }
  \subfigure[Overall]{
    \begin{minipage}[t]{0.23\linewidth}
      \includegraphics[width=\textwidth]{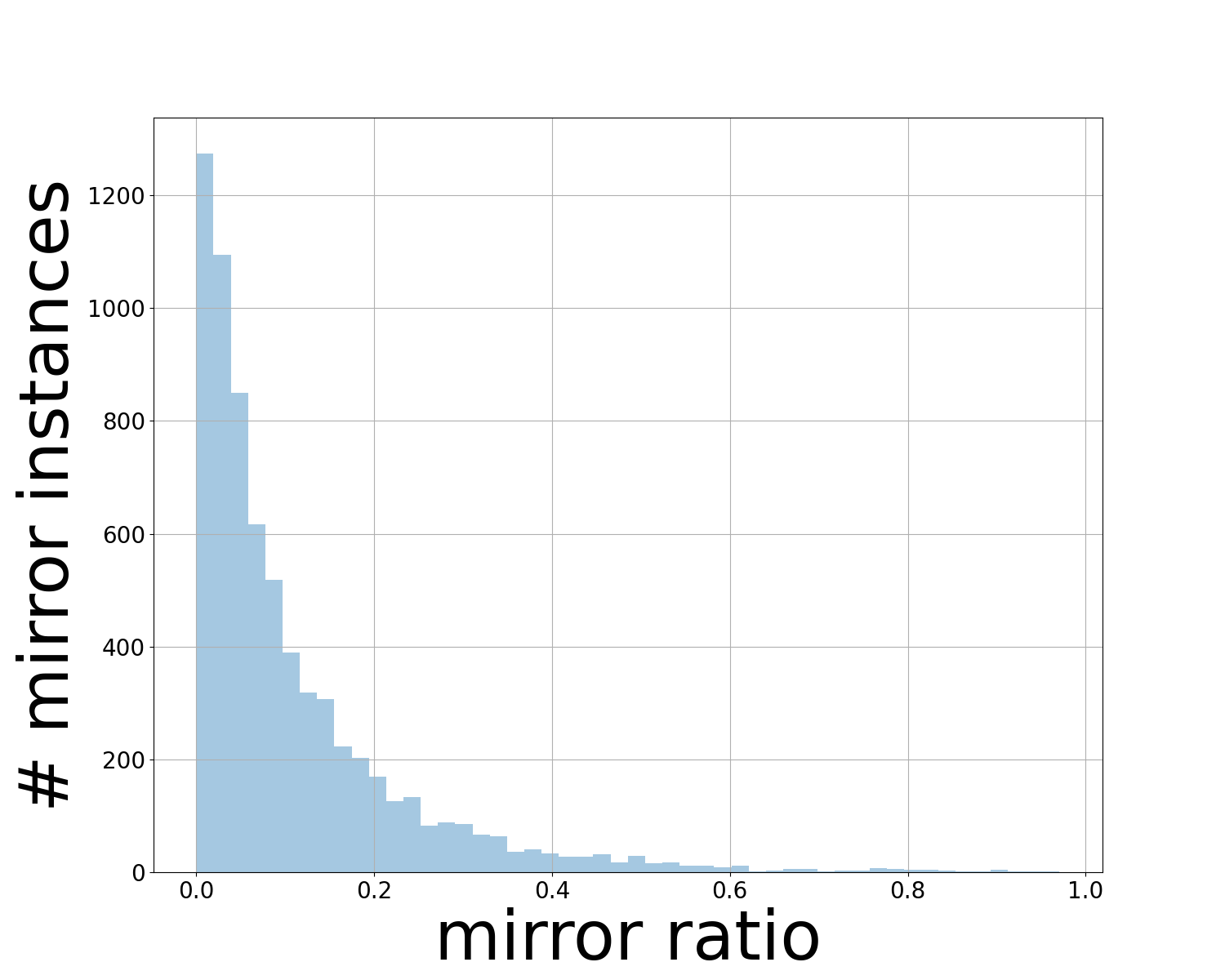}
    \end{minipage}
  }
  \caption{Distribution of the ratio of mirror pixels to total pixels, by source RGBD dataset (a through c) and overall (d).}
  \label{fig:mirror-ratio-dis}
\end{figure*}

\mypara{Mirror pixel ratio distribution.}
We define the \emph{mirror ratio} as the fraction of image pixels that belong to a mirror instance.
In \Cref{fig:mirror-ratio-dis}, we plot the mirror ratio distribution across source RGBD datasets, and for the overall Mirror3D dataset.
We note that there is a relatively broad range of mirror ratios, though few images have ratios higher than $0.5$.
This is understandable, as that would correspond to scenarios where the mirror takes up most of the image and would thus show the equipment or the operator.

\begin{figure*}
\includegraphics[width=\textwidth]{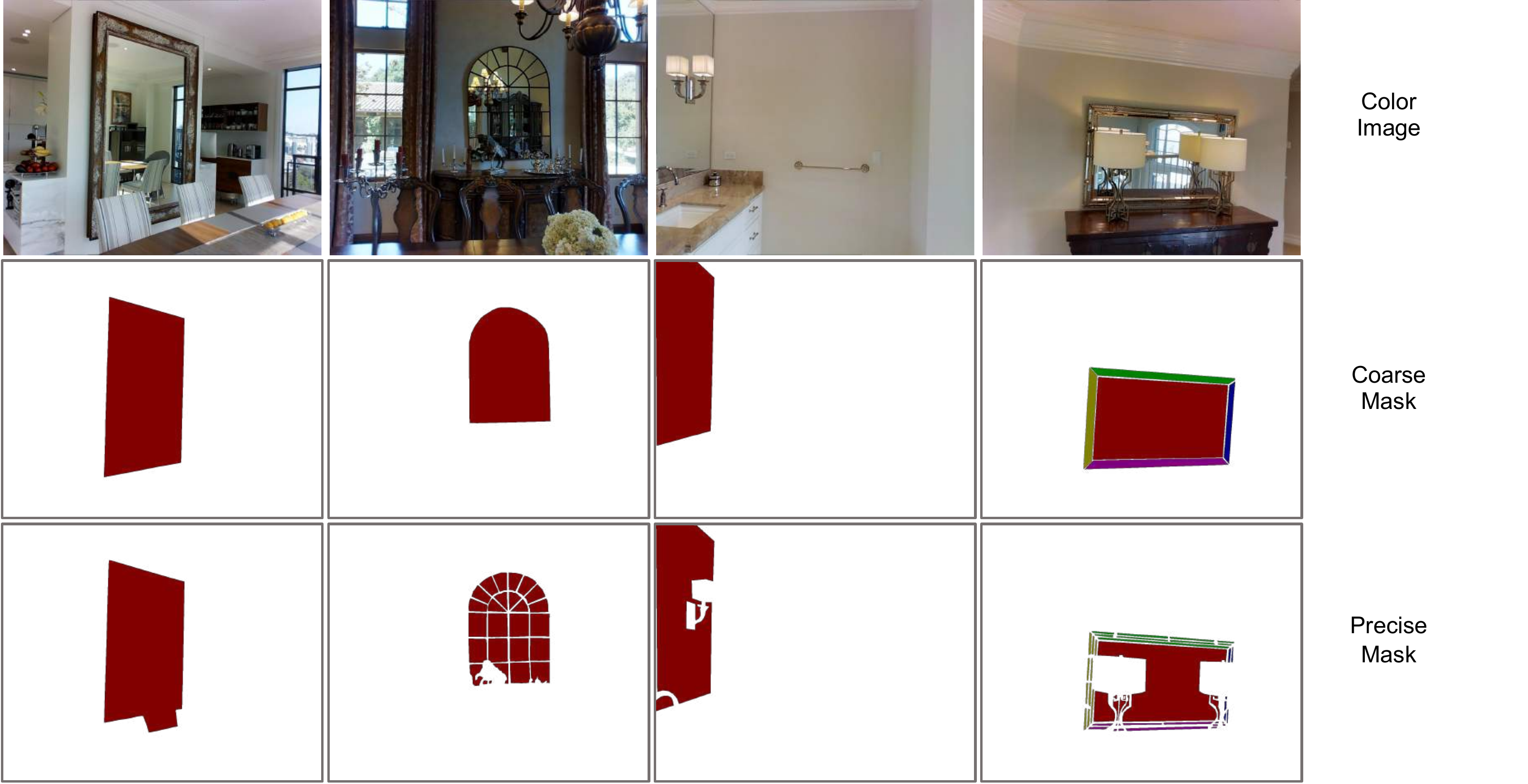}
\caption{Example of `coarse' and `precise' masks in our Mirror3D dataset, images from the Matterport3D~\cite{matterport3d} dataset.  The `precise' mask provides a pixel-accurate boundary of the visible mirror surface in each image.}
\label{fig:annotations_coarse_precise}
\end{figure*}

\mypara{Coarse vs precise masks.}
During our annotation process, we annotate both a `coarse' and a `precise' mask for each mirror plane (see \Cref{fig:annotations_coarse_precise}).  The `coarse' mask serves as an indication of the entire mirror surface (including occluded areas), while the `precise' mask provides a pixel-accurate boundary of the visible mirror surface in each image.

\begin{figure*}
\centering
\setkeys{Gin}{width=\linewidth}
\noindent
\begin{tabularx}{\textwidth}{Y@{\hspace{1.2mm}} Y@{\hspace{1.2mm}} Y@{\hspace{1.2mm}} Y@{\hspace{1.2mm}}}
\includegraphics{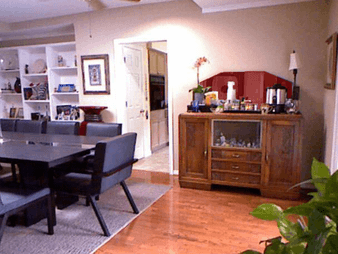} & \includegraphics{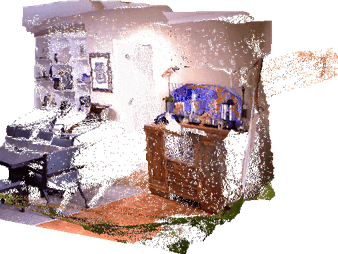} & \includegraphics{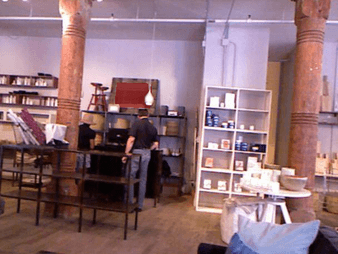} & \includegraphics{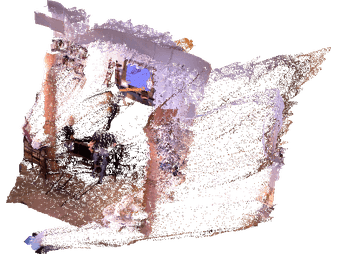} \\
\includegraphics{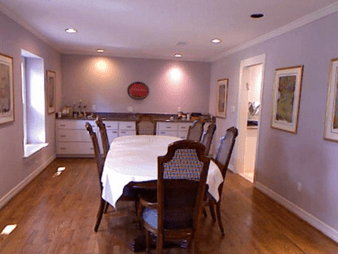} & \includegraphics{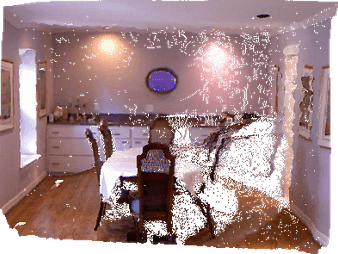} & \includegraphics{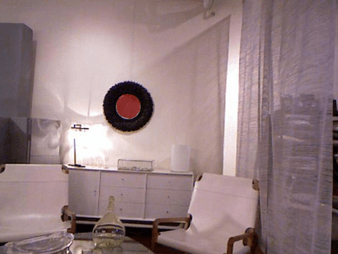} & \includegraphics{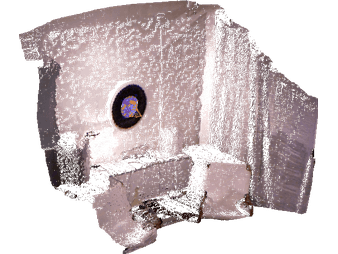} \\
\includegraphics{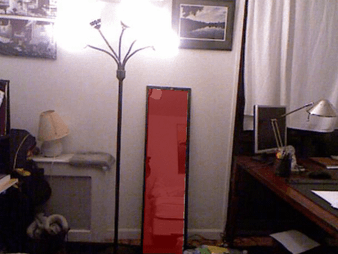} & \includegraphics{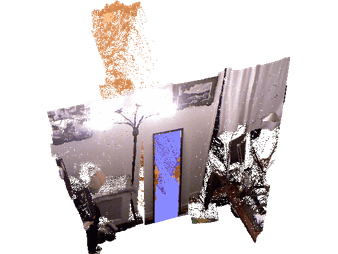} & \includegraphics{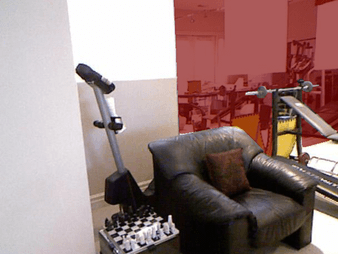} & \includegraphics{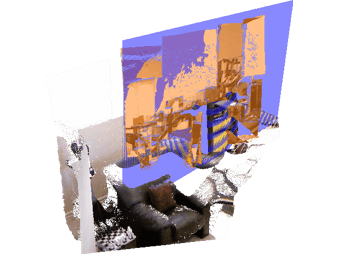} \\
\includegraphics{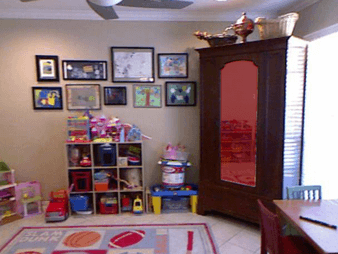} & \includegraphics{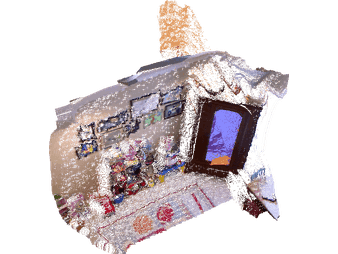} & \includegraphics{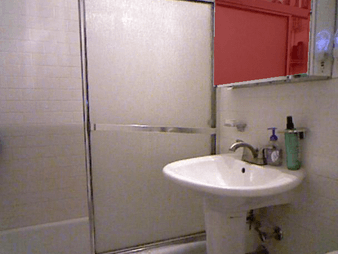} & \includegraphics{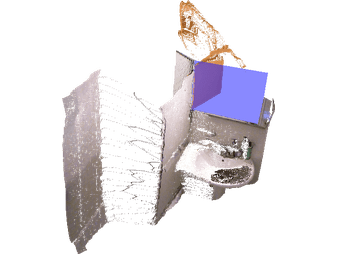} \\
\includegraphics{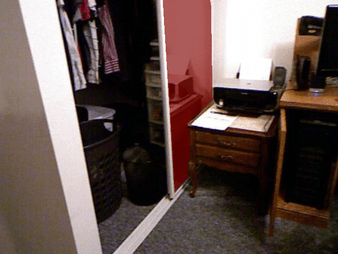} & \includegraphics{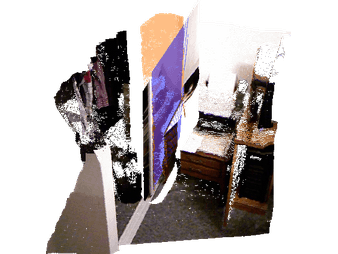} & \includegraphics{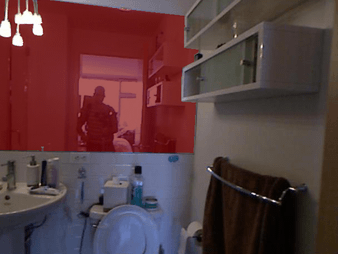} & \includegraphics{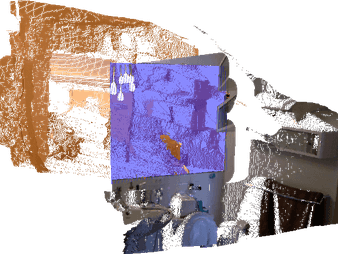} \\
\includegraphics{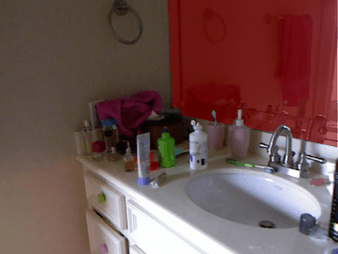} & \includegraphics{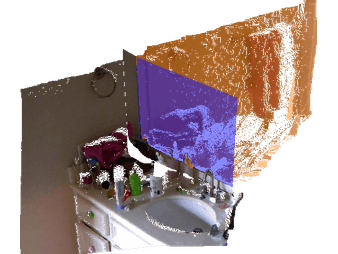} & \includegraphics{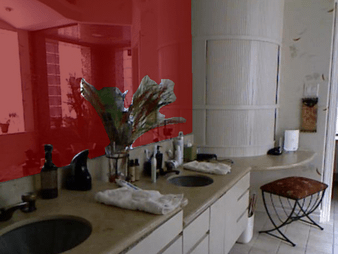} & \includegraphics{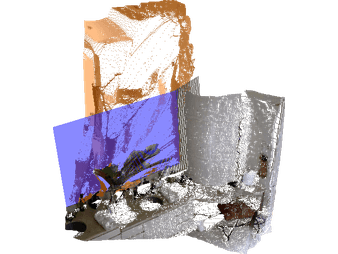} \\
\end{tabularx}
\caption{Example 3D mirror plane annotations in our Mirror3D dataset. Source RGBD data from the NYUv2~\cite{nyuv2} dataset. In each image pair, the mirror mask is shown as a transparent red on the RGB image, the mirror plane is in light blue on the point cloud, and erroneous depth points that are incorrectly behind the mirror plane in the raw depth are shaded in orange.}
\label{fig:annotations_nyu}
\end{figure*}

\begin{figure*}
\centering
\setkeys{Gin}{width=\linewidth}
\noindent
\begin{tabularx}{\textwidth}{Y@{\hspace{1.2mm}} Y@{\hspace{1.2mm}} Y@{\hspace{1.2mm}} Y@{\hspace{1.2mm}}}
\includegraphics{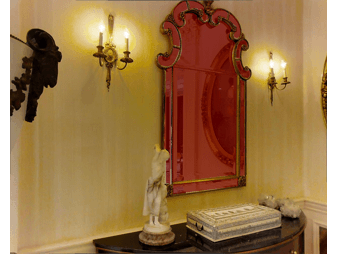} & \includegraphics{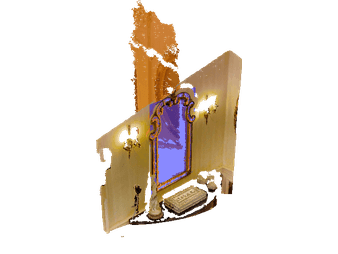} & \includegraphics{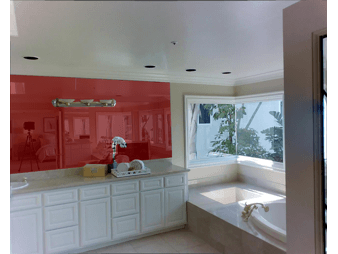} & \includegraphics{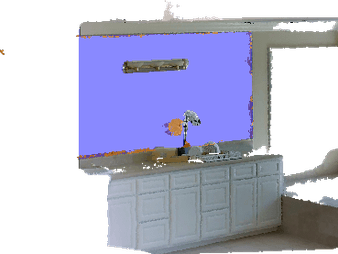} \\
\includegraphics{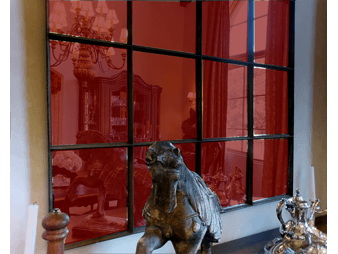} & \includegraphics{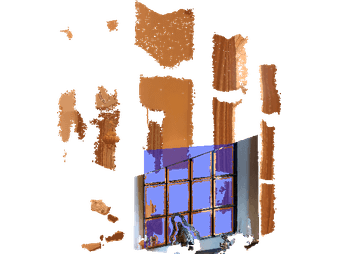} & \includegraphics{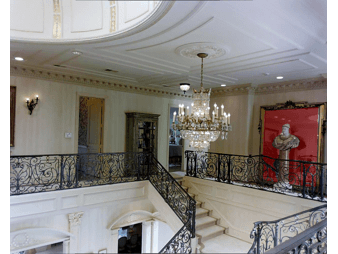} & \includegraphics{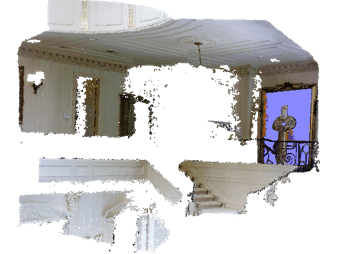} \\
\includegraphics{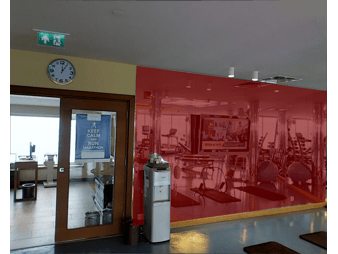} & \includegraphics{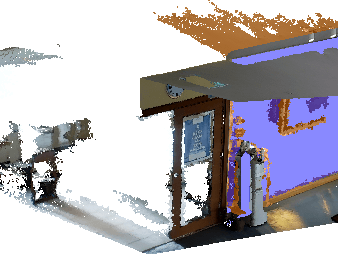} & \includegraphics{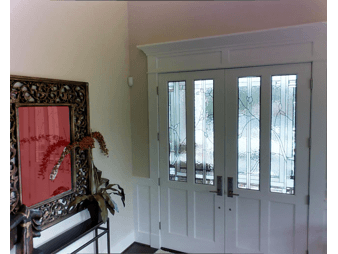} & \includegraphics{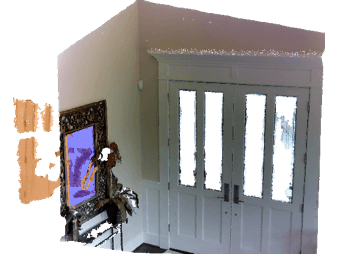} \\
\includegraphics{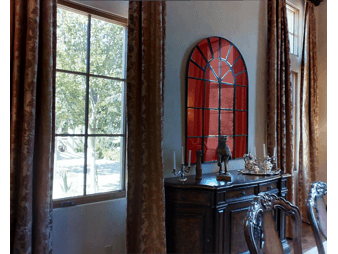} & \includegraphics{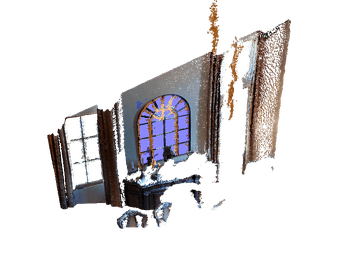} & \includegraphics{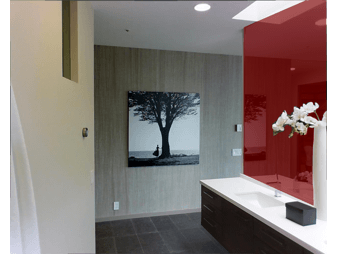} & \includegraphics{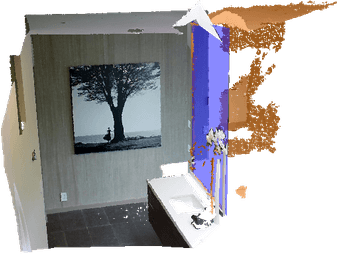} \\
\includegraphics{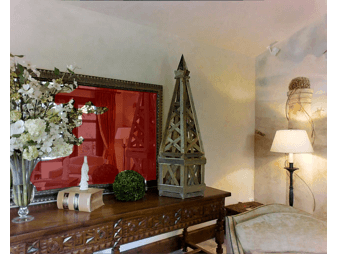} & \includegraphics{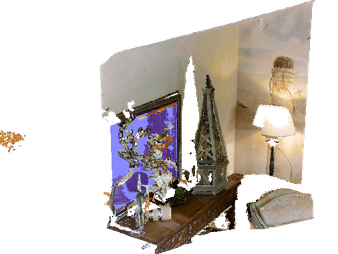} & \includegraphics{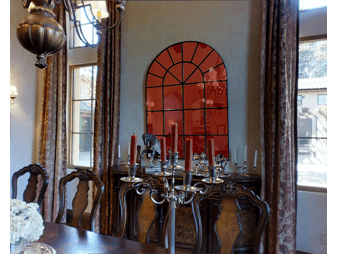} & \includegraphics{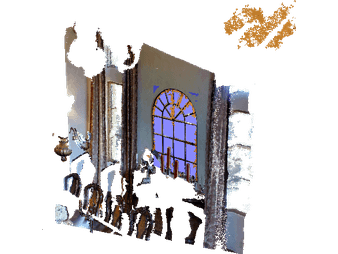} \\
\includegraphics{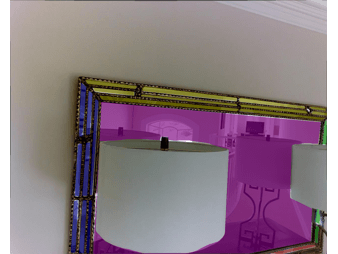} & \includegraphics{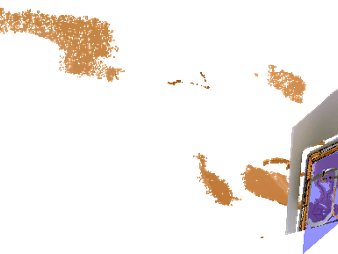} & \includegraphics{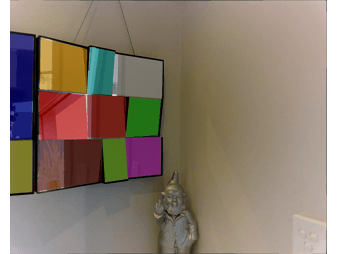} & \includegraphics{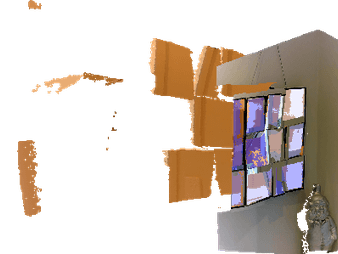} \\
\end{tabularx}
\caption{Example 3D mirror plane annotations in our Mirror3D dataset. Source RGBD data from the Matterport3D~\cite{matterport3d} dataset. In each image pair, the mirror mask is shown as a transparent polygon (with different color signifying each mirror instance) on the RGB image, the mirror plane is in light blue on the point cloud, and erroneous depth points that are incorrectly behind the mirror plane in the raw depth are shaded in orange.}
\label{fig:annotations_mp3d}
\end{figure*}

\begin{figure*}
\centering
\setkeys{Gin}{width=\linewidth}
\noindent
\begin{tabularx}{\textwidth}{Y@{\hspace{1.2mm}} Y@{\hspace{1.2mm}} Y@{\hspace{1.2mm}} Y@{\hspace{1.2mm}}}
\includegraphics{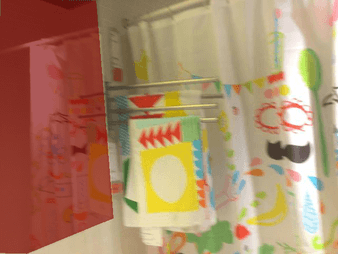} & \includegraphics{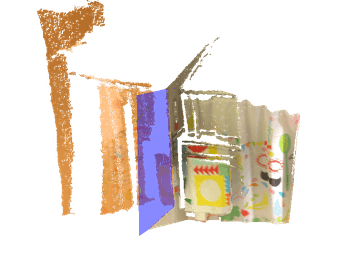} & \includegraphics{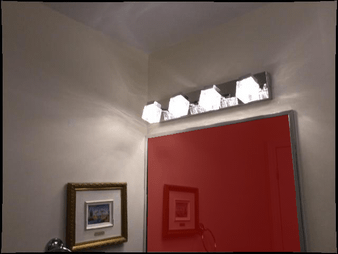} & \includegraphics{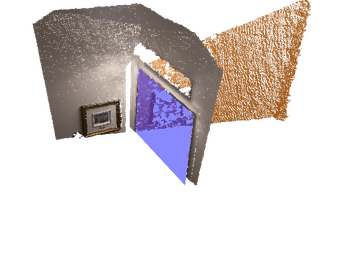} \\
\includegraphics{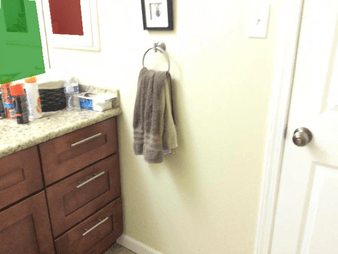} & \includegraphics{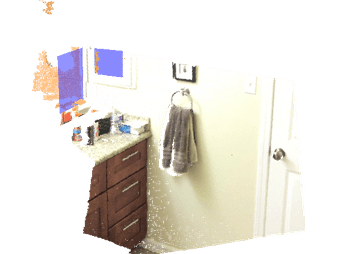} & \includegraphics{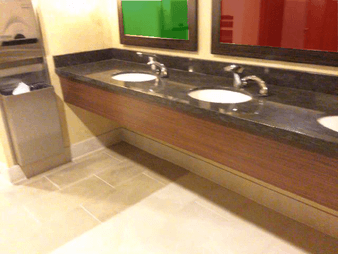} & \includegraphics{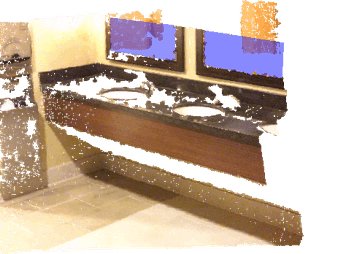} \\
\includegraphics{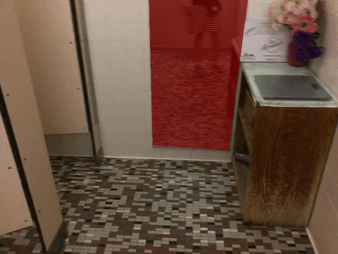} & \includegraphics{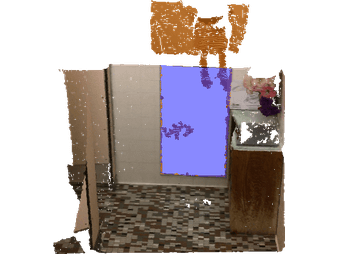} & \includegraphics{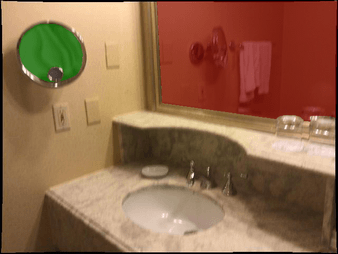} & \includegraphics{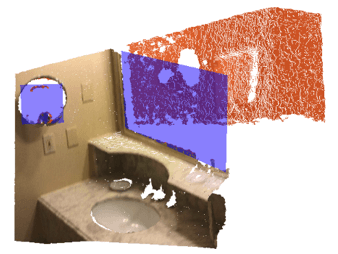} \\
\includegraphics{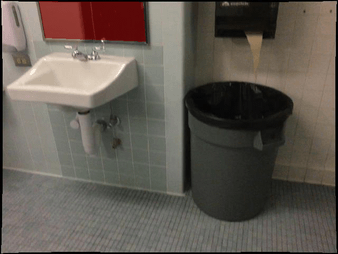} & \includegraphics{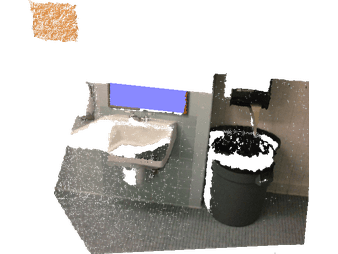} & \includegraphics{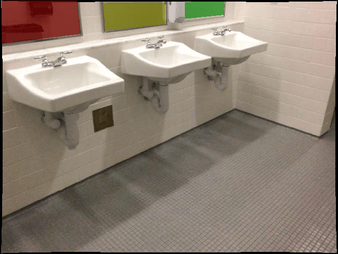} & \includegraphics{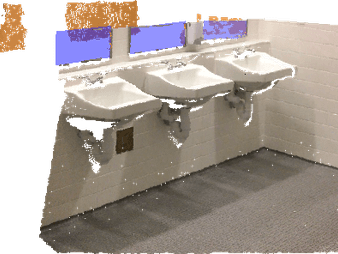} \\
\includegraphics{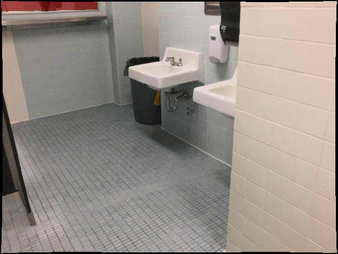} & \includegraphics{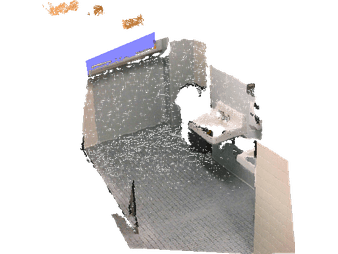} & \includegraphics{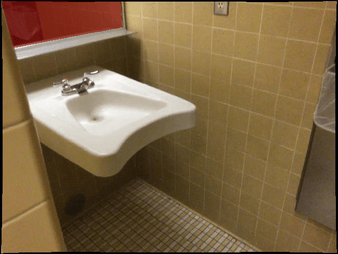} & \includegraphics{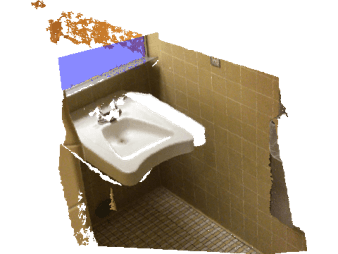} \\
\includegraphics{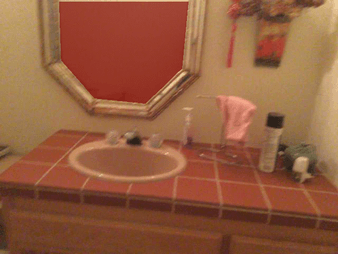} & \includegraphics{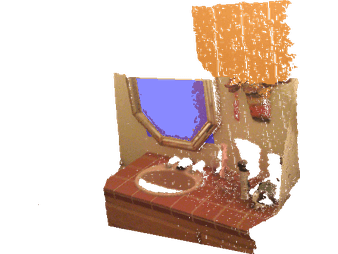} & \includegraphics{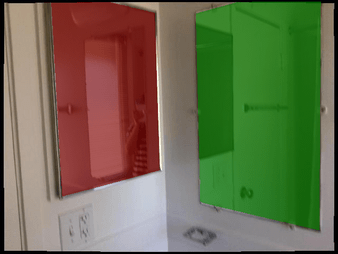} & \includegraphics{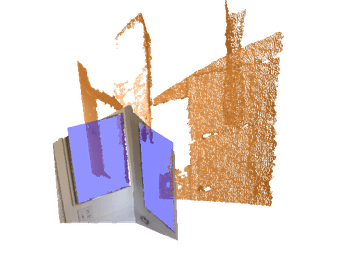} \\
\end{tabularx}
\caption{Example 3D mirror plane annotations in our Mirror3D dataset. Source RGBD data from the ScanNet~\cite{Scannet} dataset. In each image pair, the mirror mask is shown as a transparent polygon (with different color signifying each mirror instance) on the RGB image, the mirror plane is in light blue on the point cloud, and erroneous depth points that are incorrectly behind the mirror plane in the raw depth are shaded in orange.}
\label{fig:annotations_scannet}
\end{figure*}

\mypara{Example annotations.}
In \Cref{fig:annotations_nyu,fig:annotations_mp3d,fig:annotations_scannet} we provide several examples of mirror mask and 3D mirror plane annotations from our Mirror3D dataset across the three source RGBD datasets.
We show image pairs, with one image showing the mirror masks on the RGB frame, and the other visualizing the 3D mirror plane and depth points `behind' the mirror plane in each point cloud.
Mirrors occur in a variety of scenarios, and the outlier depth points cause many `floating artifacts' that our mirror plane annotations allow us to mitigate.

\clearpage
\clearpage

\section{Ablations}
\label{sec:supp:ablations}

We present ablations for two aspects of our approach: the choice of mirror anchor normal count and the mirror border threshold.
For our ablation studies, we train our model on Matterport3D train set and evaluate on the validation set.

\begin{table}
\centering
\begin{tabular}{@{} r rrr @{}}
\toprule
\# anchors & Seg-AP $\uparrow$ & $30^{\circ}$-AP $\uparrow$  \\
\midrule
3  & 0.396 & 0.253 \\
5  & 0.419 & 0.228 \\
7  & 0.434 & 0.224 \\
10 & \best{0.464} & \best{0.256} \\
12 & 0.386 & 0.199 \\
\bottomrule
\end{tabular}
\caption{
Ablation on anchor normal count.
We evaluate a spectrum of anchor normal counts in terms of mirror normal estimation. Using $10$ anchor normals gives the highest segmentation AP and 30 degree AP.
}
\label{tb:ablation-anchor}
\end{table}

\mypara{Impact of anchor normal count.}
We evaluate performance on a spectrum of anchor normal counts (i.e. different values of $k$ for the $k$-means clustering that we use to select anchor normals in the training set).
\Cref{tb:ablation-anchor} shows that using $10$ anchor normals achieves the best performance on both mirror segmentation and mirror normal estimation metrics on the \mpmesh (Matterport3D mesh-based depth) dataset.

\begin{table}
   \resizebox{\linewidth}{!}{
    \begin{tabular}{@{} l lll lll @{}}
    \toprule
     & \multicolumn{3}{c}{RMSE $\downarrow$} & \multicolumn{3}{c}{SSIM $\uparrow$} \\\cmidrule(lr){2-4}\cmidrule(lr){5-7}
    Border width & Mirror & Other & All & Mirror & Other & All \\
    \midrule
    
    15 & 0.851 & \best{0.798} & \best{0.858} & 0.823 & \best{0.798} & \best{0.791} \\
    25 & \best{0.839} & 0.800 & \best{0.858} & \best{0.825} & \best{0.798} & \best{0.791} \\
    35 & 0.842 & 0.801 & 0.859 & \best{0.825} & \best{0.798} & 0.784 \\
    \bottomrule
    \end{tabular}
    }
    \caption{Ablations on mirror border width. We tested performance on three difference border widths (number of pixels expanded outwards from mirror mask region) on the \mpmeshref dataset. We find that a border width of $25$ produces the lowest RMSE on mirror regions.}
    \label{tb:ablation border}
\end{table}

\mypara{Impact of mirror border width.} 
We experiment with different mirror border width values for estimating the mirror plane offset on the \mpmeshref dataset.
\Cref{tb:ablation border} summarizes the results.
The mirror border width has a relatively small impact on the accuracy of the estimated depth, with a mirror border width of $25$ pixels giving the best results.

\section{Additional qualitative results}
\label{sec:supp:qualitative}

We provide more qualitative results presented in a similar layout as the qualitative results figure in the main paper.
\Cref{fig:supp-qualitative} shows two qualitative comparison results from NYUv2 (top two sets) and six from Matterport3D (bottom six sets).
Note that by using \mnet to refine depth output from various depth estimation and completion approaches we can significantly reduce the depth error against the corrected ground truth depth (lower error values, particularly in mirror regions shown in the rightmost RMSE visualization column).

\begin{figure*}
\centering
\setkeys{Gin}{width=\linewidth}

\begin{tabularx}{\textwidth}{Y@{\hspace{-13mm}} Y@{\hspace{1mm}} Y@{\hspace{0mm}} Y@{\hspace{1mm}} Y@{\hspace{1mm}} Y@{\hspace{2mm}} | Y@{\hspace{1mm}} Y@{\hspace{1mm}} Y@{\hspace{2mm}} | Y@{\hspace{1mm}} Y@{\hspace{1mm}} Y@{\hspace{0mm}} }
 & & & \multicolumn{3}{c}{\textcolor{black!60!green}{\small Ground Truth}} & \multicolumn{3}{c}{\textcolor{blue!60!cyan}{\small Raw Sensor}} & \multicolumn{3}{c}{\textcolor{cyan}{\small Pred-Refined}}\\
& \small Input Depth (D) &  & \vspace{4pt}\tikzmarknode{A1}{\includegraphics{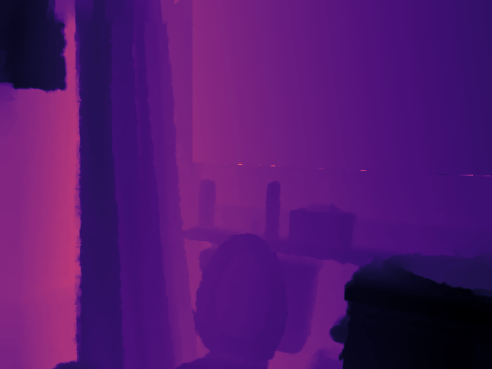}} & \vspace{4pt}\includegraphics{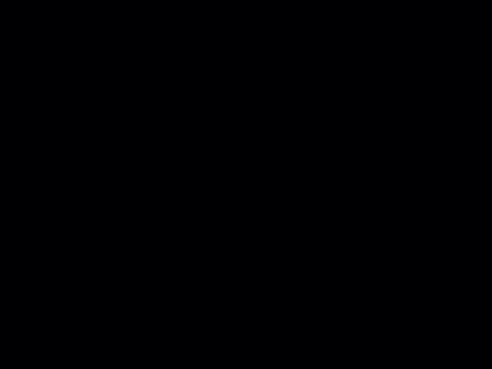} & \vspace{4pt}\tikzmarknode{A2}{\includegraphics{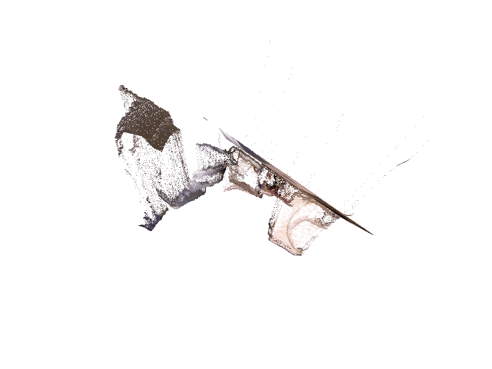}} & \vspace{4pt}\tikzmarknode{B1}{\includegraphics{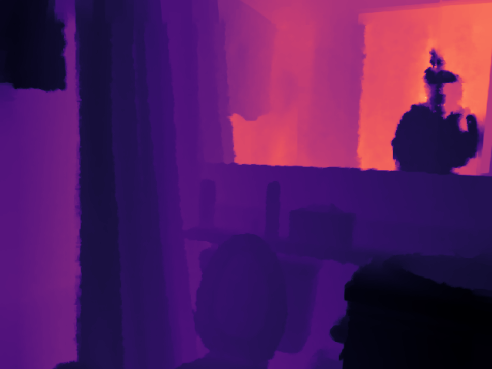}} & \vspace{4pt}\includegraphics{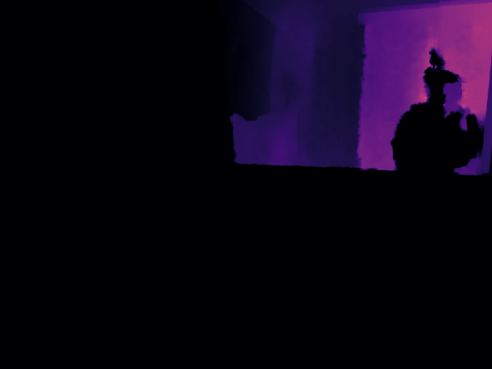} & \vspace{4pt}\tikzmarknode{B2}{\includegraphics{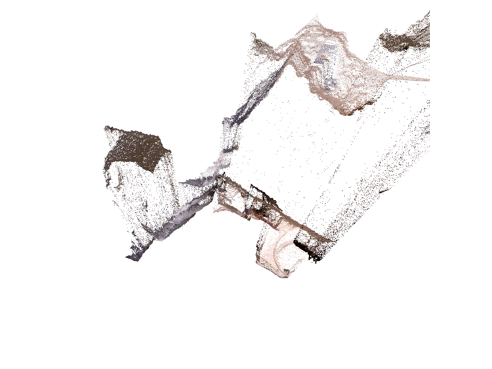}} & \vspace{4pt}\tikzmarknode{C1}{\includegraphics{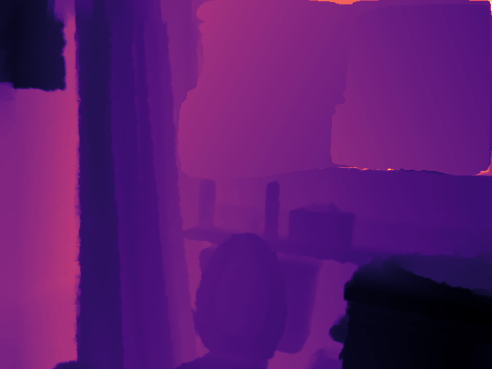}} & \vspace{4pt}\includegraphics{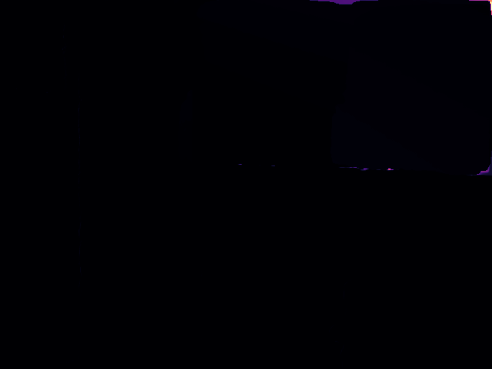} & \vspace{4pt}\tikzmarknode{C2}{\includegraphics{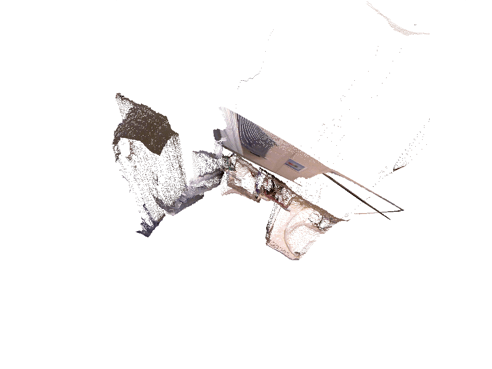}}
\\ &
\vspace{4pt}\includegraphics{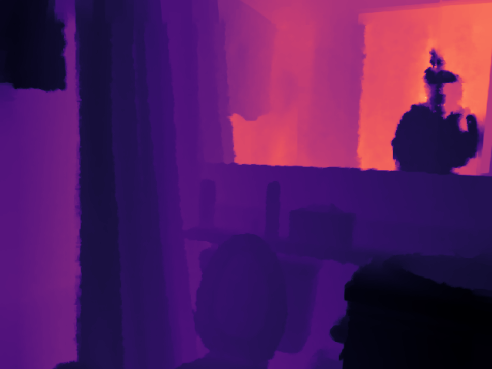} & \multicolumn{1}{r}{\scriptsize saic} & \vspace{4pt}\includegraphics{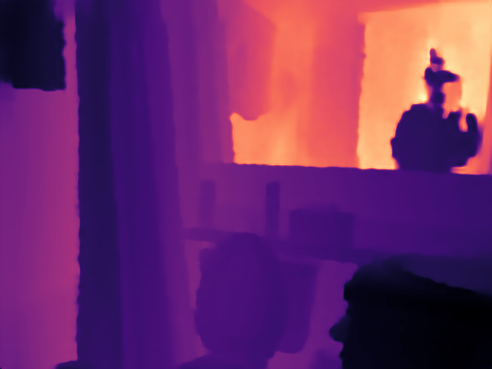} & \vspace{4pt}\includegraphics{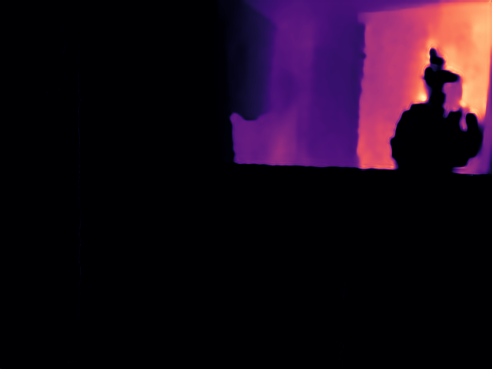} & \vspace{4pt}\includegraphics{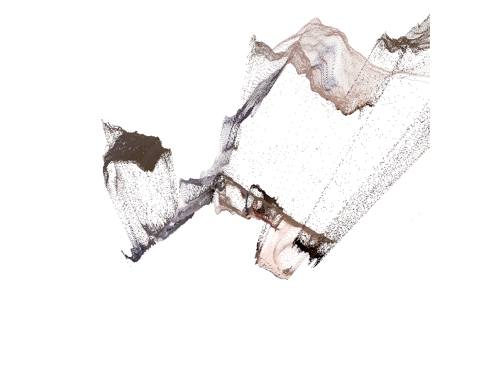} & \vspace{4pt}\includegraphics{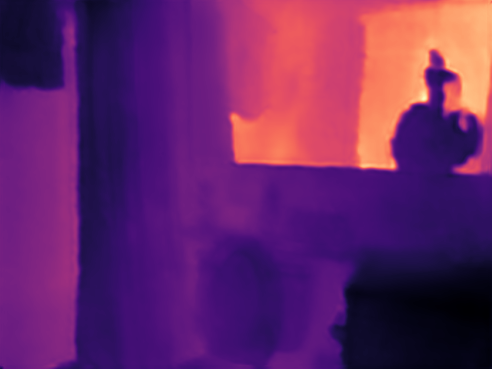} & \vspace{4pt}\includegraphics{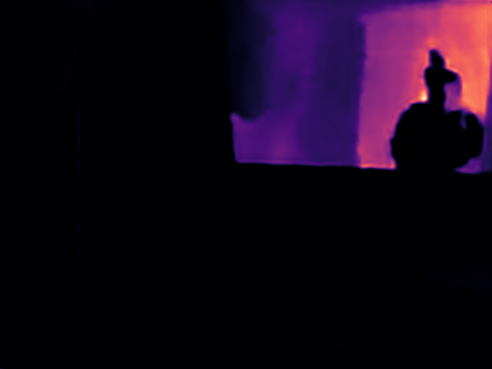} & \vspace{4pt}\includegraphics{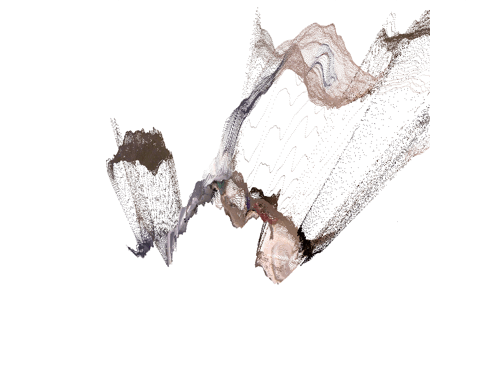} & \vspace{4pt}\includegraphics{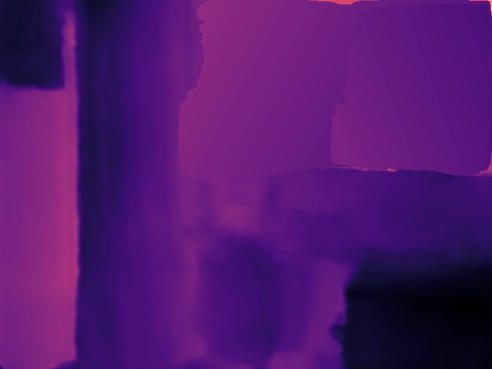} & \vspace{4pt}\includegraphics{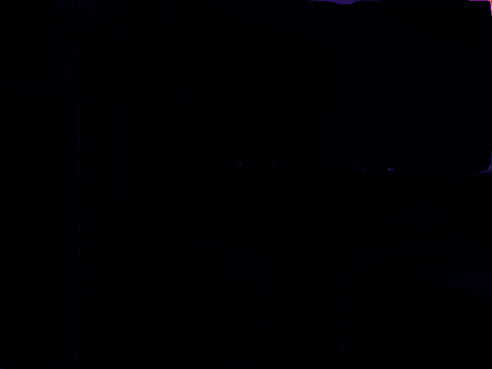} & \vspace{4pt}\includegraphics{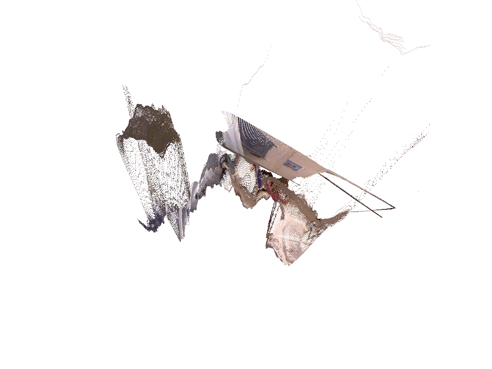} \\
\end{tabularx}

\vspace{11pt}

\begin{tabularx}{\textwidth}{Y@{\hspace{-13mm}} Y@{\hspace{1mm}} Y@{\hspace{0mm}} Y@{\hspace{1mm}} Y@{\hspace{1mm}} Y@{\hspace{2mm}} | Y@{\hspace{1mm}} Y@{\hspace{1mm}} Y@{\hspace{2mm}} | Y@{\hspace{1mm}} Y@{\hspace{1mm}} Y@{\hspace{0mm}} }
 & & & \multicolumn{3}{c}{\textcolor{black!60!green}{\small Ground Truth}} & \multicolumn{3}{c}{\textcolor{blue!60!cyan}{\small Raw Sensor}} & \multicolumn{3}{c}{\textcolor{cyan}{\small Pred-Refined}}\\
& \small Input Depth (D) &  & \vspace{4pt}\tikzmarknode{A1}{\includegraphics{figure/new_result_vis/nyu_440_0.png}} & \vspace{4pt}\includegraphics{figure/new_result_vis/nyu_440_1.png} & \vspace{4pt}\tikzmarknode{A2}{\includegraphics{figure/new_result_vis/nyu_440_2.png}} & \vspace{4pt}\tikzmarknode{B1}{\includegraphics{figure/new_result_vis/nyu_440_3.png}} & \vspace{4pt}\includegraphics{figure/new_result_vis/nyu_440_4.png} & \vspace{4pt}\tikzmarknode{B2}{\includegraphics{figure/new_result_vis/nyu_440_5.png}} & \vspace{4pt}\tikzmarknode{C1}{\includegraphics{figure/new_result_vis/nyu_440_6.png}} & \vspace{4pt}\includegraphics{figure/new_result_vis/nyu_440_7.png} & \vspace{4pt}\tikzmarknode{C2}{\includegraphics{figure/new_result_vis/nyu_440_8.png}}
\\ &
\vspace{4pt}\includegraphics{figure/new_result_vis/nyu_440_9.png} & \multicolumn{1}{r}{\scriptsize saic} & \vspace{4pt}\includegraphics{figure/new_result_vis/nyu_440_10.png} & \vspace{4pt}\includegraphics{figure/new_result_vis/nyu_440_11.png} & \vspace{4pt}\includegraphics{figure/new_result_vis/nyu_440_12.png} & \vspace{4pt}\includegraphics{figure/new_result_vis/nyu_440_13.png} & \vspace{4pt}\includegraphics{figure/new_result_vis/nyu_440_14.png} & \vspace{4pt}\includegraphics{figure/new_result_vis/nyu_440_15.png} & \vspace{4pt}\includegraphics{figure/new_result_vis/nyu_440_16.png} & \vspace{4pt}\includegraphics{figure/new_result_vis/nyu_440_17.png} & \vspace{4pt}\includegraphics{figure/new_result_vis/nyu_440_18.png} \\
\end{tabularx}

\vspace{11pt}

\begin{tabularx}{\textwidth}{Y@{\hspace{-13mm}} Y@{\hspace{1mm}} Y@{\hspace{0mm}} Y@{\hspace{1mm}} Y@{\hspace{1mm}} Y@{\hspace{2mm}} | Y@{\hspace{1mm}} Y@{\hspace{1mm}} Y@{\hspace{2mm}} | Y@{\hspace{1mm}} Y@{\hspace{1mm}} Y@{\hspace{0mm}} }
 & & & \multicolumn{3}{c}{\textcolor{black!60!green}{\small Ground Truth}} & \multicolumn{3}{c}{\textcolor{blue!60!cyan}{\small MP3D-Mesh Rendered Depth}} & \multicolumn{3}{c}{\textcolor{cyan}{\small Pred-Refined}}\\
& \small Input Depth (D) &  & \vspace{4pt}\tikzmarknode{A1}{\includegraphics{figure/new_result_vis/mp3d_2389e105b2584b3997a819cfe0b99d30_i1_1_0.png}} & \vspace{4pt}\includegraphics{figure/new_result_vis/mp3d_2389e105b2584b3997a819cfe0b99d30_i1_1_1.png} & \vspace{4pt}\tikzmarknode{A2}{\includegraphics{figure/new_result_vis/mp3d_2389e105b2584b3997a819cfe0b99d30_i1_1_2.png}} & \vspace{4pt}\tikzmarknode{B1}{\includegraphics{figure/new_result_vis/mp3d_2389e105b2584b3997a819cfe0b99d30_i1_1_3.png}} & \vspace{4pt}\includegraphics{figure/new_result_vis/mp3d_2389e105b2584b3997a819cfe0b99d30_i1_1_4.png} & \vspace{4pt}\tikzmarknode{B2}{\includegraphics{figure/new_result_vis/mp3d_2389e105b2584b3997a819cfe0b99d30_i1_1_5.png}} & \vspace{4pt}\tikzmarknode{C1}{\includegraphics{figure/new_result_vis/mp3d_2389e105b2584b3997a819cfe0b99d30_i1_1_6.png}} & \vspace{4pt}\includegraphics{figure/new_result_vis/mp3d_2389e105b2584b3997a819cfe0b99d30_i1_1_7.png} & \vspace{4pt}\tikzmarknode{C2}{\includegraphics{figure/new_result_vis/mp3d_2389e105b2584b3997a819cfe0b99d30_i1_1_8.png}}
\\
\end{tabularx}

\end{figure*}

\begin{figure*}
\centering
\setkeys{Gin}{width=\linewidth}
\begin{tabularx}{\textwidth}{Y@{\hspace{-13mm}} Y@{\hspace{1mm}} Y@{\hspace{0mm}} Y@{\hspace{1mm}} Y@{\hspace{1mm}} Y@{\hspace{2mm}} | Y@{\hspace{1mm}} Y@{\hspace{1mm}} Y@{\hspace{2mm}} | Y@{\hspace{1mm}} Y@{\hspace{1mm}} Y@{\hspace{0mm}} }
 & & & \multicolumn{3}{c}{\textcolor{black!60!green}{\small Ground Truth}} & \multicolumn{3}{c}{\textcolor{blue!60!cyan}{\small MP3D-Mesh Rendered Depth}} & \multicolumn{3}{c}{\textcolor{cyan}{\small Pred-Refined}}\\
& \small Input Depth (D) &  & \vspace{4pt}\tikzmarknode{A1}{\includegraphics{figure/new_result_vis/mp3d_fa1e045efce4481f9b398c97366a5dcc_i1_4_0.png}} & \vspace{4pt}\includegraphics{figure/new_result_vis/mp3d_fa1e045efce4481f9b398c97366a5dcc_i1_4_1.png} & \vspace{4pt}\tikzmarknode{A2}{\includegraphics{figure/new_result_vis/mp3d_fa1e045efce4481f9b398c97366a5dcc_i1_4_2.png}} & \vspace{4pt}\tikzmarknode{B1}{\includegraphics{figure/new_result_vis/mp3d_fa1e045efce4481f9b398c97366a5dcc_i1_4_3.png}} & \vspace{4pt}\includegraphics{figure/new_result_vis/mp3d_fa1e045efce4481f9b398c97366a5dcc_i1_4_4.png} & \vspace{4pt}\tikzmarknode{B2}{\includegraphics{figure/new_result_vis/mp3d_fa1e045efce4481f9b398c97366a5dcc_i1_4_5.png}} & \vspace{4pt}\tikzmarknode{C1}{\includegraphics{figure/new_result_vis/mp3d_fa1e045efce4481f9b398c97366a5dcc_i1_4_6.png}} & \vspace{4pt}\includegraphics{figure/new_result_vis/mp3d_fa1e045efce4481f9b398c97366a5dcc_i1_4_7.png} & \vspace{4pt}\tikzmarknode{C2}{\includegraphics{figure/new_result_vis/mp3d_fa1e045efce4481f9b398c97366a5dcc_i1_4_8.png}}
\\
\end{tabularx}

\vspace{11pt}
\begin{tabularx}{\textwidth}{Y@{\hspace{-13mm}} Y@{\hspace{1mm}} Y@{\hspace{0mm}} Y@{\hspace{1mm}} Y@{\hspace{1mm}} Y@{\hspace{2mm}} | Y@{\hspace{1mm}} Y@{\hspace{1mm}} Y@{\hspace{2mm}} | Y@{\hspace{1mm}} Y@{\hspace{1mm}} Y@{\hspace{0mm}} }
 & & & \multicolumn{3}{c}{\textcolor{black!60!green}{\small Ground Truth}} & \multicolumn{3}{c}{\textcolor{blue!60!cyan}{\small MP3D-Mesh Rendered Depth}} & \multicolumn{3}{c}{\textcolor{cyan}{\small Pred-Refined}}\\
& \small Input Depth (D) &  & \vspace{4pt}\tikzmarknode{A1}{\includegraphics{figure/new_result_vis/mp3d_46b120dde4004bf9ada90cf967c0585b_i1_2_0.png}} & \vspace{4pt}\includegraphics{figure/new_result_vis/mp3d_46b120dde4004bf9ada90cf967c0585b_i1_2_1.png} & \vspace{4pt}\tikzmarknode{A2}{\includegraphics{figure/new_result_vis/mp3d_46b120dde4004bf9ada90cf967c0585b_i1_2_2.png}} & \vspace{4pt}\tikzmarknode{B1}{\includegraphics{figure/new_result_vis/mp3d_46b120dde4004bf9ada90cf967c0585b_i1_2_3.png}} & \vspace{4pt}\includegraphics{figure/new_result_vis/mp3d_46b120dde4004bf9ada90cf967c0585b_i1_2_4.png} & \vspace{4pt}\tikzmarknode{B2}{\includegraphics{figure/new_result_vis/mp3d_46b120dde4004bf9ada90cf967c0585b_i1_2_5.png}} & \vspace{4pt}\tikzmarknode{C1}{\includegraphics{figure/new_result_vis/mp3d_46b120dde4004bf9ada90cf967c0585b_i1_2_6.png}} & \vspace{4pt}\includegraphics{figure/new_result_vis/mp3d_46b120dde4004bf9ada90cf967c0585b_i1_2_7.png} & \vspace{4pt}\tikzmarknode{C2}{\includegraphics{figure/new_result_vis/mp3d_46b120dde4004bf9ada90cf967c0585b_i1_2_8.png}}
\\
\end{tabularx}

\vspace{11pt}

\begin{tabularx}{\textwidth}{Y@{\hspace{-13mm}} Y@{\hspace{1mm}} Y@{\hspace{0mm}} Y@{\hspace{1mm}} Y@{\hspace{1mm}} Y@{\hspace{2mm}} | Y@{\hspace{1mm}} Y@{\hspace{1mm}} Y@{\hspace{2mm}} | Y@{\hspace{1mm}} Y@{\hspace{1mm}} Y@{\hspace{0mm}} }
 & & & \multicolumn{3}{c}{\textcolor{black!60!green}{\small Ground Truth}} & \multicolumn{3}{c}{\textcolor{blue!60!cyan}{\small MP3D-Mesh Rendered Depth}} & \multicolumn{3}{c}{\textcolor{cyan}{\small Pred-Refined}}\\
& \small Input Depth (D) &  & \vspace{4pt}\tikzmarknode{A1}{\includegraphics{figure/new_result_vis/mp3d_e5adc5c026db41378897c0b8393c2361_i1_4_0.png}} & \vspace{4pt}\includegraphics{figure/new_result_vis/mp3d_e5adc5c026db41378897c0b8393c2361_i1_4_1.png} & \vspace{4pt}\tikzmarknode{A2}{\includegraphics{figure/new_result_vis/mp3d_e5adc5c026db41378897c0b8393c2361_i1_4_2.png}} & \vspace{4pt}\tikzmarknode{B1}{\includegraphics{figure/new_result_vis/mp3d_e5adc5c026db41378897c0b8393c2361_i1_4_3.png}} & \vspace{4pt}\includegraphics{figure/new_result_vis/mp3d_e5adc5c026db41378897c0b8393c2361_i1_4_4.png} & \vspace{4pt}\tikzmarknode{B2}{\includegraphics{figure/new_result_vis/mp3d_e5adc5c026db41378897c0b8393c2361_i1_4_5.png}} & \vspace{4pt}\tikzmarknode{C1}{\includegraphics{figure/new_result_vis/mp3d_e5adc5c026db41378897c0b8393c2361_i1_4_6.png}} & \vspace{4pt}\includegraphics{figure/new_result_vis/mp3d_e5adc5c026db41378897c0b8393c2361_i1_4_7.png} & \vspace{4pt}\tikzmarknode{C2}{\includegraphics{figure/new_result_vis/mp3d_e5adc5c026db41378897c0b8393c2361_i1_4_8.png}}
\\
\end{tabularx}

\end{figure*}

\begin{figure*}


\centering
\setkeys{Gin}{width=\linewidth}

\begin{tabularx}{\textwidth}{Y@{\hspace{-13mm}} Y@{\hspace{1mm}} Y@{\hspace{0mm}} Y@{\hspace{1mm}} Y@{\hspace{1mm}} Y@{\hspace{2mm}} | Y@{\hspace{1mm}} Y@{\hspace{1mm}} Y@{\hspace{2mm}} | Y@{\hspace{1mm}} Y@{\hspace{1mm}} Y@{\hspace{0mm}} }
 & & & \multicolumn{3}{c}{\textcolor{black!60!green}{\small Ground Truth}} & \multicolumn{3}{c}{\textcolor{blue!60!cyan}{\small MP3D-Mesh Rendered Depth}} & \multicolumn{3}{c}{\textcolor{cyan}{\small Pred-Refined}}\\
& \small Input Depth (D) &  & \vspace{4pt}\tikzmarknode{A1}{\includegraphics{figure/new_result_vis/mp3d_02e0ac86ffc647e48a677f327431e4b7_i1_4_0.png}} & \vspace{4pt}\includegraphics{figure/new_result_vis/mp3d_02e0ac86ffc647e48a677f327431e4b7_i1_4_1.png} & \vspace{4pt}\tikzmarknode{A2}{\includegraphics{figure/new_result_vis/mp3d_02e0ac86ffc647e48a677f327431e4b7_i1_4_2.png}} & \vspace{4pt}\tikzmarknode{B1}{\includegraphics{figure/new_result_vis/mp3d_02e0ac86ffc647e48a677f327431e4b7_i1_4_3.png}} & \vspace{4pt}\includegraphics{figure/new_result_vis/mp3d_02e0ac86ffc647e48a677f327431e4b7_i1_4_4.png} & \vspace{4pt}\tikzmarknode{B2}{\includegraphics{figure/new_result_vis/mp3d_02e0ac86ffc647e48a677f327431e4b7_i1_4_5.png}} & \vspace{4pt}\tikzmarknode{C1}{\includegraphics{figure/new_result_vis/mp3d_02e0ac86ffc647e48a677f327431e4b7_i1_4_6.png}} & \vspace{4pt}\includegraphics{figure/new_result_vis/mp3d_02e0ac86ffc647e48a677f327431e4b7_i1_4_7.png} & \vspace{4pt}\tikzmarknode{C2}{\includegraphics{figure/new_result_vis/mp3d_02e0ac86ffc647e48a677f327431e4b7_i1_4_8.png}}
\\
\end{tabularx}

\vspace{11pt}

\begin{tabularx}{\textwidth}{Y@{\hspace{-13mm}} Y@{\hspace{1mm}} Y@{\hspace{0mm}} Y@{\hspace{1mm}} Y@{\hspace{1mm}} Y@{\hspace{2mm}} | Y@{\hspace{1mm}} Y@{\hspace{1mm}} Y@{\hspace{2mm}} | Y@{\hspace{1mm}} Y@{\hspace{1mm}} Y@{\hspace{0mm}} }
 & & & \multicolumn{3}{c}{\textcolor{black!60!green}{\small Ground Truth}} & \multicolumn{3}{c}{\textcolor{blue!60!cyan}{\small MP3D-Mesh Rendered Depth}} & \multicolumn{3}{c}{\textcolor{cyan}{\small Pred-Refined}}\\
& \small Input Depth (D) &  & \vspace{4pt}\tikzmarknode{A1}{\includegraphics{figure/new_result_vis/mp3d_dd1ed8e5c888446a9c00ecc18c151311_i1_0_0.png}} & \vspace{4pt}\includegraphics{figure/new_result_vis/mp3d_dd1ed8e5c888446a9c00ecc18c151311_i1_0_1.png} & \vspace{4pt}\tikzmarknode{A2}{\includegraphics{figure/new_result_vis/mp3d_dd1ed8e5c888446a9c00ecc18c151311_i1_0_2.png}} & \vspace{4pt}\tikzmarknode{B1}{\includegraphics{figure/new_result_vis/mp3d_dd1ed8e5c888446a9c00ecc18c151311_i1_0_3.png}} & \vspace{4pt}\includegraphics{figure/new_result_vis/mp3d_dd1ed8e5c888446a9c00ecc18c151311_i1_0_4.png} & \vspace{4pt}\tikzmarknode{B2}{\includegraphics{figure/new_result_vis/mp3d_dd1ed8e5c888446a9c00ecc18c151311_i1_0_5.png}} & \vspace{4pt}\tikzmarknode{C1}{\includegraphics{figure/new_result_vis/mp3d_dd1ed8e5c888446a9c00ecc18c151311_i1_0_6.png}} & \vspace{4pt}\includegraphics{figure/new_result_vis/mp3d_dd1ed8e5c888446a9c00ecc18c151311_i1_0_7.png} & \vspace{4pt}\tikzmarknode{C2}{\includegraphics{figure/new_result_vis/mp3d_dd1ed8e5c888446a9c00ecc18c151311_i1_0_8.png}}
\\
\end{tabularx}%
\caption{Additional qualitative results. The top two samples are results from NYUv2~\cite{nyuv2}, the bottom six samples are from Matterport3D~\cite{matterport3d}.}
\label{fig:supp-qualitative}
\end{figure*}

\begin{figure*}
\centering
\setkeys{Gin}{width=\linewidth}
\noindent

\begin{tabularx}{\textwidth}{Y@{\hspace{1.5mm}} Y@{\hspace{1.5mm}} Y@{\hspace{1.5mm}} Y@{\hspace{1.5mm}} Y@{\hspace{0mm}}}
 Mirror3DNet & Mirror3DNet-mesh & Mirror3DNet-mesh-ref & PlaneRCNN-mesh & PlaneRCNN-mesh-ref  \\
\includegraphics{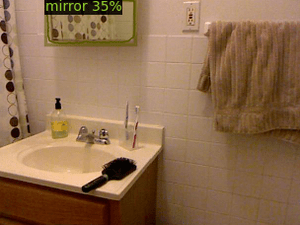} &
\includegraphics{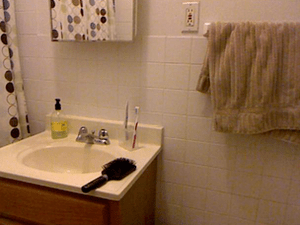} & \includegraphics{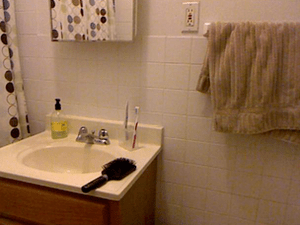} &
\includegraphics{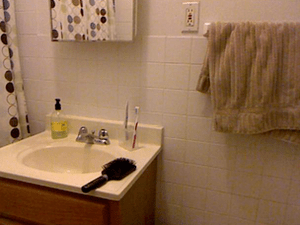} & \includegraphics{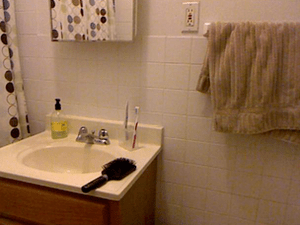}
\\
\includegraphics{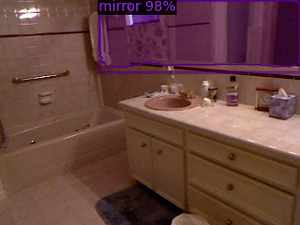} &
\includegraphics{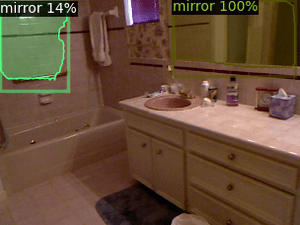} & \includegraphics{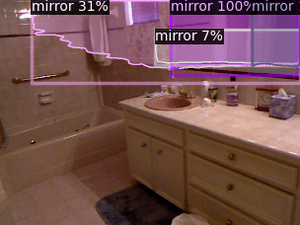} &
\includegraphics{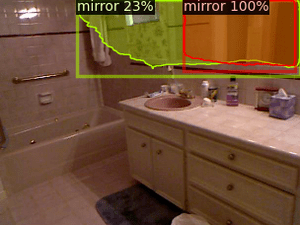} & \includegraphics{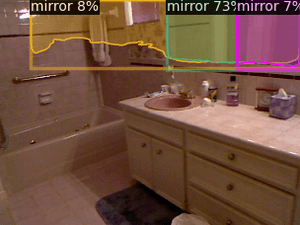} \\
\includegraphics{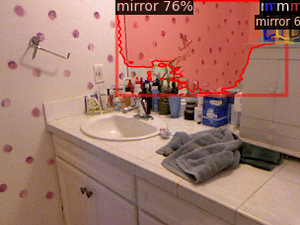} &
\includegraphics{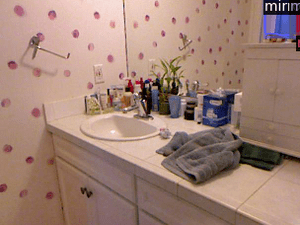} & \includegraphics{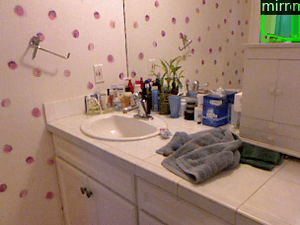} &
\includegraphics{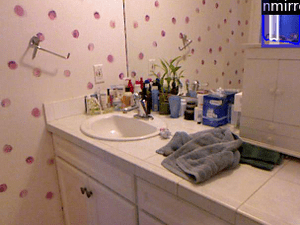} & \includegraphics{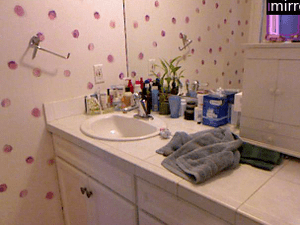} \\
\end{tabularx}
\centering
\caption{
Mirror segmentation results for NYUv2.
Models were trained on Matterport3D mirror data We see that \mnet trained on Matterport3D generalizes to NYUv2 mirror data and gives the best segmentation results.
The predicted mirror mask is shown as a transparent polygon with a confidence score on the left top corner in each image.
}
\label{fig:mirror-seg-nyuv2}

\vspace{\baselineskip}

\begin{tabularx}{\textwidth}{Y@{\hspace{1.5mm}} Y@{\hspace{1.5mm}} Y@{\hspace{1.5mm}} Y@{\hspace{1.5mm}} Y@{\hspace{0mm}}}
 Mirror3DNet & Mirror3DNet-mesh & Mirror3DNet-mesh-ref & PlaneRCNN-mesh & PlaneRCNN-mesh-ref \\
\includegraphics{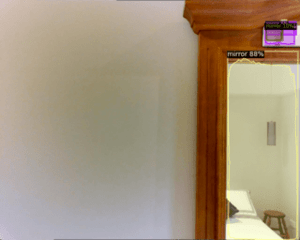} &
\includegraphics{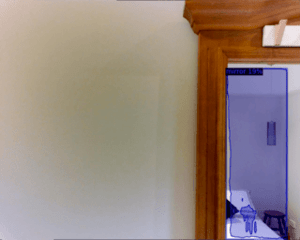} & \includegraphics{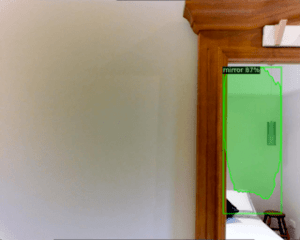} &
\includegraphics{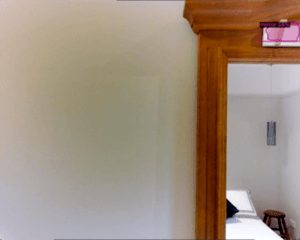} & \includegraphics{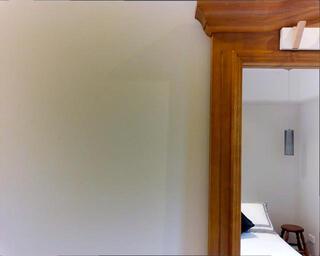}
\\
\includegraphics{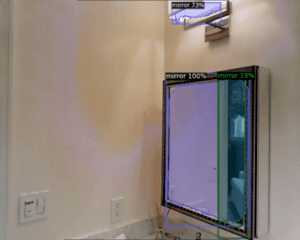} &
\includegraphics{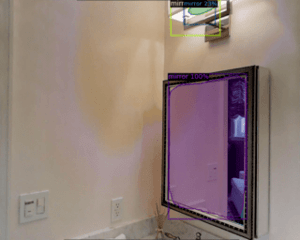} & \includegraphics{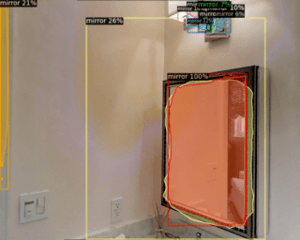} &
\includegraphics{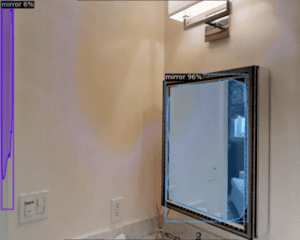} & \includegraphics{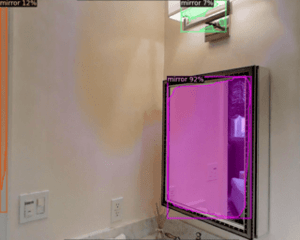}
\\
\includegraphics{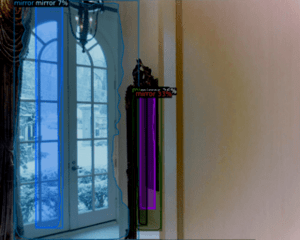} &
\includegraphics{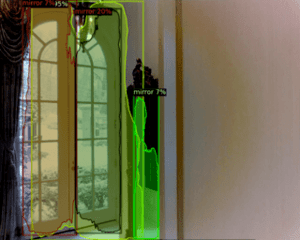} & \includegraphics{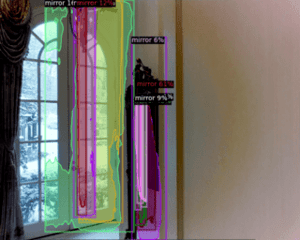} &
\includegraphics{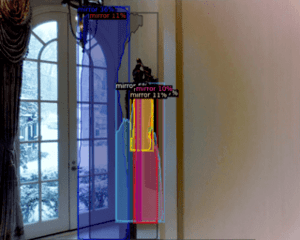} & \includegraphics{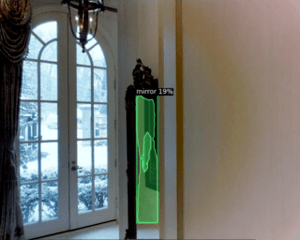} \\
\end{tabularx}
\centering
\caption{
Mirror segmentation results for Matterport3D.
Overall, we see that \mnet gives the best results.
}
\label{fig:mirror-seg-mp3d}
\end{figure*}

We also provide qualitative examples of predicted mirror segmentations for NYUv2 (\Cref{fig:mirror-seg-nyuv2}) and Matterport3D (\Cref{fig:mirror-seg-mp3d}).
We train \mnet and PlaneRCNN on the Matterport3D train split, and evaluate on NYUv2 and Matterport3D test splits.
We note that \mnet trained on just RGB, no depth (first column) provides better detection of mirrors than when trained with depth (other columns).
Due to the noisy nature of the depth values around mirror regions, depth is not a reliable signal for the presence of a mirror for both \mnet and PlaneRCNN.

\section{Additional quantitative results}
\label{sec:supp:quantitative}

Since pixels with depth value $<0.00001$ are filtered out during evaluation, we provide statistics analyzing what percentage of pixels are filtered (\Cref{tab:filtered-pixel-statistics}).
For NYUv2, there are no pixels that are filtered.
In contrast, we see that a large fraction of mirror pixels are invalid for raw sensor depth from Matterport3D and ScanNet.
For depth derived from the Matterport3D mesh reconstruction, there are fewer mirror depth values that are filtered out during evaluation.
Overall, the percent of pixels that are filtered out during evaluation is low.

We also compile a complete set of result tables including all the quantitative evaluation metrics we defined in the main paper.
We train separate models for NYUv2 and Matterport.
For each dataset, models are trained on the train split and evaluated on the test split.
We train all models three times with different random seeds and report the average and the standard error across the three runs.
In \Cref{tab:nyuv2-ref-results-complete} we report the full set of metrics for experiments using the \nyuref dataset.
Contrast these results with the ones obtained when using the \nyuraw dataset, shown in \Cref{tab:nyuv2-raw-results-complete} to see the impact of using the original raw depth as the ground truth for evaluation.
Similarly, in \Cref{tab:m3d-refined-results-complete} we report complete metrics for the experiments using \mpmeshref, and contrast these metrics against the original dataset depth being used as ground truth in \Cref{tab:m3d-raw-results-complete}.

\clearpage
\clearpage

\begin{table*}[!htp]
\centering
\resizebox{\linewidth} {!} {
\begin{tabular}{@{}lrrrr rrrr rrrr@{}}
\toprule
& \multicolumn{4}{c}{Matterport3D-sensor} & \multicolumn{4}{c}{Matterport3D-mesh} & \multicolumn{4}{c}{ScanNet-sensor}
\\\cmidrule(lr){2-5}\cmidrule(lr){6-9} \cmidrule(lr){10-13}
 & Train & Val & Test & \textbf{All} & Train & Val & Test & \textbf{All} & Train & Val & Test & \textbf{All}\\
\midrule
\% mirror pixels over frames & $39.89$ & $34.00$ & $40.31$ & $40.10$ & $6.00$ & $5.93$ & $6.75$ & $6.34$ & $30.04$ & $32.20$ & $33.74$ & $31.89$ \\
\% mirror pixels over mirror pixels & $29.05$ & $20.68$ & $28.20$ & $24.87$ & $4.75$ & $2.51$ & $3.48$ & $3.58$ & $24.34$ & $26.00$ & $26.69$ & $25.52$ \\
\% pixels over all pixels     & $3.96$  & $2.90$  & $4.51$  & $4.24$ & $0.65$ & $0.35$ & $0.56$ & $0.52$ & $2.94$  & $3.18$ & $2.30$ & $3.06$ \\
\bottomrule
\end{tabular}
}

\caption{
Breakdown of percent of pixels with original depth values $< 0.00001$ across train/val/test splits for Matterport3D and ScanNet.
NYUv2-sensor does not have any pixels with depth value $< 0.00001$.
Note that during evaluation (on val/test), these pixels are filtered out.
We see that a large fraction of mirror pixels are invalid for raw sensor depth from Matterport3D and ScanNet. For depth derived from Matterport3D mesh reconstruction, there are fewer mirror pixel values that are filtered out during evaluation.
Overall, the percent of pixels that are filtered out during evaluation is low.
After refining the depth for mirrors, our \DATASET dataset does not contain any depth value $< 0.00001$ and so all pixels are included in the evaluation.}
\label{tab:filtered-pixel-statistics}
\end{table*}

\begin{table*}
\resizebox{\linewidth}{!}{
\begin{tabular}{@{} lll ccc ccc ccc ccc @{}}
\toprule
& & & \multicolumn{3}{c}{RMSE $\downarrow$} & \multicolumn{3}{c}{s-RMSE $\downarrow$} & \multicolumn{3}{c}{AbsRel $\downarrow$}  & \multicolumn{3}{c}{SSIM $\uparrow$}
\\\cmidrule(lr){4-6}\cmidrule(lr){7-9}\cmidrule(lr){10-12}\cmidrule(lr){13-15}
Input & Train & Method & Mirror & Other & All & Mirror & Other & All & Mirror & Other & All & Mirror & Other & All   \\
\midrule
sensor-D &          * &                                             * & 1.272 $\pm$ 0.113 &\best{0.022 } $\pm$  0.004& 0.424 $\pm$ 0.046 & 0.941 $\pm$ 0.082 & 0.199 $\pm$ 0.026 & 0.342 $\pm$ 0.033 & 0.670 $\pm$ 0.069 &\best{0.001 } $\pm$  0.000& 0.099 $\pm$ 0.017 & 0.634 $\pm$ 0.024 &\best{0.997 } $\pm$  0.000& 0.946 $\pm$ 0.006\\
sensor-D &          * &                                         \mnet &\best{0.766 } $\pm$  0.058& 0.168 $\pm$ 0.056 &\best{0.363 } $\pm$  0.058&\best{0.617 } $\pm$  0.037&\best{0.189 } $\pm$  0.023&\best{0.271 } $\pm$  0.023&\best{0.401 } $\pm$  0.032& 0.015 $\pm$ 0.005 &\best{0.073 } $\pm$  0.009&\best{0.800 } $\pm$  0.013& 0.983 $\pm$ 0.002 &\best{0.954 } $\pm$  0.003\\
\midrule
 RGBD &        ref &                                          saic~\cite{senushkin2020decoder} & 0.965 $\pm$ 0.063 & 0.094 $\pm$ 0.008 &\best{0.344 } $\pm$  0.021& 0.760 $\pm$ 0.048 &\best{0.184 } $\pm$  0.012& 0.293 $\pm$ 0.017 & 0.477 $\pm$ 0.034 &\best{0.020 } $\pm$  0.001&\best{0.081 } $\pm$  0.006& 0.709 $\pm$ 0.014 & 0.926 $\pm$ 0.005 &\best{0.892 } $\pm$  0.005\\
 RGBD &        raw &                                          saic~\cite{senushkin2020decoder} & 1.245 $\pm$ 0.066 &\best{0.070 } $\pm$  0.004& 0.433 $\pm$ 0.026 & 0.918 $\pm$ 0.048 & 0.220 $\pm$ 0.014 & 0.350 $\pm$ 0.019 & 0.655 $\pm$ 0.040 & 0.026 $\pm$ 0.002 & 0.119 $\pm$ 0.010 & 0.641 $\pm$ 0.014 &\best{0.930 } $\pm$  0.003& 0.887 $\pm$ 0.003\\
\bottomrule\end{tabular}}
\resizebox{\linewidth}{!}{
\begin{tabular}{@{} lll ccc ccc ccc ccc ccc @{}}
\toprule
& & & \multicolumn{3}{c}{1.05 $\uparrow$} & \multicolumn{3}{c}{1.10 $\uparrow$} & \multicolumn{3}{c}{1.25 $\uparrow$} & \multicolumn{3}{c}{${1.25}^2 \uparrow$} & \multicolumn{3}{c}{ ${1.25}^3\uparrow$}
\\\cmidrule(lr){4-6}\cmidrule(lr){7-9}\cmidrule(lr){10-12}\cmidrule(lr){13-15}\cmidrule(lr){16-18}
Input & Train & Method & Mirror & Other & All & Mirror & Other & All & Mirror & Other & All & Mirror & Other & All & Mirror & Other & All  \\
\midrule
sensor-D &          * &                                             * & 0.168 $\pm$ 0.037 &\best{0.996 } $\pm$  0.001&\best{0.894 } $\pm$  0.013& 0.227 $\pm$ 0.041 &\best{0.998 } $\pm$  0.001&\best{0.901 } $\pm$  0.013& 0.353 $\pm$ 0.048 &\best{0.999 } $\pm$  0.000& 0.912 $\pm$ 0.012 & 0.527 $\pm$ 0.047 &\best{0.999 } $\pm$  0.000& 0.931 $\pm$ 0.011 & 0.695 $\pm$ 0.043 &\best{1.000 } $\pm$  0.000& 0.950 $\pm$ 0.010\\
sensor-D &          * &                                         \mnet &\best{0.206 } $\pm$  0.025& 0.969 $\pm$ 0.004 & 0.873 $\pm$ 0.008 &\best{0.290 } $\pm$  0.029& 0.977 $\pm$ 0.004 & 0.888 $\pm$ 0.008 &\best{0.495 } $\pm$  0.031& 0.987 $\pm$ 0.003 &\best{0.917 } $\pm$  0.007&\best{0.745 } $\pm$  0.025& 0.994 $\pm$ 0.002 &\best{0.957 } $\pm$  0.006&\best{0.890 } $\pm$  0.017& 0.998 $\pm$ 0.001 &\best{0.980 } $\pm$  0.004\\
\midrule
 RGBD &        ref &                                          saic~\cite{senushkin2020decoder} & 0.176 $\pm$ 0.021 &\best{0.917 } $\pm$  0.010&\best{0.825 } $\pm$  0.010& 0.267 $\pm$ 0.025 &\best{0.969 } $\pm$  0.005&\best{0.881 } $\pm$  0.007& 0.474 $\pm$ 0.029 &\best{0.995 } $\pm$  0.001&\best{0.927 } $\pm$  0.006& 0.685 $\pm$ 0.027 & 0.998 $\pm$ 0.001 &\best{0.956 } $\pm$  0.005& 0.828 $\pm$ 0.021 & 0.999 $\pm$ 0.000 & 0.975 $\pm$ 0.004\\
\bottomrule
\end{tabular}
}
\caption{Additional quantitative metrics for experiments on \nyuref test set images containing mirrors (NYUv2~\cite{nyuv2} with ground truth depth refined using 3D mirror plane annotations).}
\label{tab:nyuv2-ref-results-complete}
\end{table*}

\begin{table*}
\resizebox{\linewidth}{!}{
\begin{tabular}{@{} lll ccc ccc ccc ccc @{}}
\toprule
& & & \multicolumn{3}{c}{RMSE $\downarrow$} & \multicolumn{3}{c}{s-RMSE $\downarrow$} & \multicolumn{3}{c}{AbsRel $\downarrow$}  & \multicolumn{3}{c}{SSIM $\uparrow$}
\\\cmidrule(lr){4-6}\cmidrule(lr){7-9}\cmidrule(lr){10-12}\cmidrule(lr){13-15}
Input & Train & Method & Mirror & Other & All & Mirror & Other & All & Mirror & Other & All & Mirror & Other & All   \\
\midrule
sensor-D &          * &                                             * &\best{0.000 } $\pm$  0.000&\best{0.000 } $\pm$  0.000&\best{0.000 } $\pm$  0.000&\best{0.000 } $\pm$  0.000&\best{0.000 } $\pm$  0.000&\best{0.000 } $\pm$  0.000&\best{0.000 } $\pm$  0.000&\best{0.000 } $\pm$  0.000&\best{0.000 } $\pm$  0.000&\best{1.000 } $\pm$  0.000&\best{1.000 } $\pm$  0.000&\best{1.000 } $\pm$  0.000\\
sensor-D &          * &                                         \mnet & 0.784 $\pm$ 0.073 & 0.158 $\pm$ 0.056 & 0.369 $\pm$ 0.059 & 0.709 $\pm$ 0.056 & 0.162 $\pm$ 0.023 & 0.295 $\pm$ 0.027 & 0.165 $\pm$ 0.013 & 0.014 $\pm$ 0.005 & 0.034 $\pm$ 0.006 & 0.772 $\pm$ 0.016 & 0.986 $\pm$ 0.002 & 0.958 $\pm$ 0.003\\
\midrule
 RGBD &        ref &                                          saic~\cite{senushkin2020decoder} & 0.429 $\pm$ 0.039 & 0.090 $\pm$ 0.008 & 0.205 $\pm$ 0.017 & 0.378 $\pm$ 0.033 & 0.112 $\pm$ 0.009 & 0.184 $\pm$ 0.016 & 0.107 $\pm$ 0.010 &\best{0.020 } $\pm$  0.001& 0.035 $\pm$ 0.002 & 0.815 $\pm$ 0.011 & 0.928 $\pm$ 0.005 & 0.907 $\pm$ 0.005\\
 RGBD &        raw &                                          saic~\cite{senushkin2020decoder} &\best{0.107 } $\pm$  0.007&\best{0.062 } $\pm$  0.003&\best{0.067 } $\pm$  0.003&\best{0.104 } $\pm$  0.007&\best{0.058 } $\pm$  0.003&\best{0.064 } $\pm$  0.003&\best{0.027 } $\pm$  0.003& 0.025 $\pm$ 0.002 &\best{0.024 } $\pm$  0.002&\best{0.890 } $\pm$  0.004&\best{0.932 } $\pm$  0.003&\best{0.924 } $\pm$  0.003\\
\bottomrule\end{tabular}}
\resizebox{\linewidth}{!}{
\begin{tabular}{@{} lll ccc ccc ccc ccc ccc @{}}
\toprule
& & & \multicolumn{3}{c}{1.05 $\uparrow$} & \multicolumn{3}{c}{1.10 $\uparrow$} & \multicolumn{3}{c}{1.25 $\uparrow$} & \multicolumn{3}{c}{${1.25}^2 \uparrow$} & \multicolumn{3}{c}{ ${1.25}^3\uparrow$}
\\\cmidrule(lr){4-6}\cmidrule(lr){7-9}\cmidrule(lr){10-12}\cmidrule(lr){13-15}\cmidrule(lr){16-18}
Input & Train & Method & Mirror & Other & All & Mirror & Other & All & Mirror & Other & All & Mirror & Other & All & Mirror & Other & All  \\
\midrule
sensor-D &          * &                                             * &\best{1.000 } $\pm$  0.000&\best{1.000 } $\pm$  0.000&\best{1.000 } $\pm$  0.000&\best{1.000 } $\pm$  0.000&\best{1.000 } $\pm$  0.000&\best{1.000 } $\pm$  0.000&\best{1.000 } $\pm$  0.000&\best{1.000 } $\pm$  0.000&\best{1.000 } $\pm$  0.000&\best{1.000 } $\pm$  0.000&\best{1.000 } $\pm$  0.000&\best{1.000 } $\pm$  0.000&\best{1.000 } $\pm$  0.000&\best{1.000 } $\pm$  0.000&\best{1.000 } $\pm$  0.000\\
sensor-D &          * &                                         \mnet & 0.508 $\pm$ 0.031 & 0.972 $\pm$ 0.004 & 0.915 $\pm$ 0.007 & 0.555 $\pm$ 0.031 & 0.979 $\pm$ 0.004 & 0.924 $\pm$ 0.007 & 0.638 $\pm$ 0.028 & 0.988 $\pm$ 0.003 & 0.941 $\pm$ 0.006 & 0.799 $\pm$ 0.022 & 0.995 $\pm$ 0.002 & 0.970 $\pm$ 0.004 & 0.932 $\pm$ 0.012 & 0.998 $\pm$ 0.001 & 0.990 $\pm$ 0.002\\
\midrule
 RGBD &        ref &                                          saic~\cite{senushkin2020decoder} & 0.628 $\pm$ 0.029 &\best{0.920 } $\pm$  0.010& 0.874 $\pm$ 0.011 & 0.710 $\pm$ 0.027 &\best{0.970 } $\pm$  0.005& 0.928 $\pm$ 0.007 & 0.807 $\pm$ 0.024 &\best{0.995 } $\pm$  0.001& 0.963 $\pm$ 0.005 & 0.880 $\pm$ 0.019 & 0.998 $\pm$ 0.001 & 0.976 $\pm$ 0.004 & 0.944 $\pm$ 0.013 & 0.999 $\pm$ 0.000 & 0.987 $\pm$ 0.003\\
\bottomrule
\end{tabular}}
\caption{Additional quantitative metrics for experiments on \nyuraw test set images containing mirrors (NYUv2~\cite{nyuv2} using original raw depth as ground truth). Compare to \Cref{tab:nyuv2-ref-results-complete} and note the incorrect ranking of methods with this imperfect ground truth.}
    \label{tab:nyuv2-raw-results-complete}
\end{table*}

\clearpage
\clearpage

\begin{table*}
\resizebox{\linewidth}{!}{
\begin{tabular}{@{} lll ccc ccc ccc ccc @{}}
\toprule
& & & \multicolumn{3}{c}{RMSE $\downarrow$} & \multicolumn{3}{c}{s-RMSE $\downarrow$} & \multicolumn{3}{c}{AbsRel $\downarrow$}  & \multicolumn{3}{c}{SSIM $\uparrow$}
\\\cmidrule(lr){4-6}\cmidrule(lr){7-9}\cmidrule(lr){10-12}\cmidrule(lr){13-15}
Input & Train & Method & Mirror & Other & All & Mirror & Other & All & Mirror & Other & All & Mirror & Other & All   \\
\midrule
sensor-D &          * &                                             * & 2.676 $\pm$ 0.071 & 0.959 $\pm$ 0.041 & 1.325 $\pm$ 0.038 & 2.293 $\pm$ 0.077 & 1.051 $\pm$ 0.040 & 1.178 $\pm$ 0.039 & 0.986 $\pm$ 0.022 & 0.143 $\pm$ 0.006 & 0.324 $\pm$ 0.015 & 0.287 $\pm$ 0.007 & 0.787 $\pm$ 0.006 & 0.707 $\pm$ 0.006\\
mesh-D &          * &                                             * & 0.661 $\pm$ 0.051 &\best{0.000 } $\pm$  0.000&\best{0.155 } $\pm$  0.010& 0.630 $\pm$ 0.050 &\best{0.029 } $\pm$  0.004&\best{0.146 } $\pm$  0.009& 0.144 $\pm$ 0.009 &\best{0.000 } $\pm$  0.000&\best{0.021 } $\pm$  0.002& 0.841 $\pm$ 0.009 &\best{1.000 } $\pm$  0.000&\best{0.980 } $\pm$  0.001\\
sensor-D &          * &                                         \mnet & 1.249 $\pm$ 0.039 & 0.949 $\pm$ 0.023 & 1.017 $\pm$ 0.023 & 1.137 $\pm$ 0.038 & 0.954 $\pm$ 0.023 & 0.965 $\pm$ 0.023 & 0.362 $\pm$ 0.010 & 0.150 $\pm$ 0.004 & 0.204 $\pm$ 0.006 & 0.738 $\pm$ 0.006 & 0.785 $\pm$ 0.003 & 0.772 $\pm$ 0.003\\
mesh-D &          * &                                         \mnet &\best{0.477 } $\pm$  0.022& 0.156 $\pm$ 0.007 & 0.226 $\pm$ 0.008 &\best{0.439 } $\pm$  0.020& 0.162 $\pm$ 0.007 & 0.208 $\pm$ 0.007 &\best{0.128 } $\pm$  0.005& 0.016 $\pm$ 0.001 & 0.034 $\pm$ 0.001 &\best{0.893 } $\pm$  0.003& 0.979 $\pm$ 0.001 & 0.964 $\pm$ 0.001\\
\bottomrule\end{tabular}}
\resizebox{\linewidth}{!}{
\begin{tabular}{@{} lll ccc ccc ccc ccc ccc @{}}
\toprule
& & & \multicolumn{3}{c}{1.05 $\uparrow$} & \multicolumn{3}{c}{1.10 $\uparrow$} & \multicolumn{3}{c}{1.25 $\uparrow$} & \multicolumn{3}{c}{${1.25}^2 \uparrow$} & \multicolumn{3}{c}{ ${1.25}^3\uparrow$}
\\\cmidrule(lr){4-6}\cmidrule(lr){7-9}\cmidrule(lr){10-12}\cmidrule(lr){13-15}\cmidrule(lr){16-18}
Input & Train & Method & Mirror & Other & All & Mirror & Other & All & Mirror & Other & All & Mirror & Other & All & Mirror & Other & All  \\
\midrule
sensor-D &          * &                                             * & 0.121 $\pm$ 0.006 & 0.834 $\pm$ 0.006 & 0.716 $\pm$ 0.008 & 0.140 $\pm$ 0.006 & 0.848 $\pm$ 0.006 & 0.730 $\pm$ 0.008 & 0.190 $\pm$ 0.007 & 0.861 $\pm$ 0.006 & 0.748 $\pm$ 0.007 & 0.286 $\pm$ 0.009 & 0.873 $\pm$ 0.005 & 0.772 $\pm$ 0.007 & 0.365 $\pm$ 0.011 & 0.879 $\pm$ 0.005 & 0.792 $\pm$ 0.007\\
mesh-D &          * &                                             * &\best{0.718 } $\pm$  0.013&\best{1.000 } $\pm$  0.000&\best{0.952 } $\pm$  0.004&\best{0.788 } $\pm$  0.011&\best{1.000 } $\pm$  0.000&\best{0.966 } $\pm$  0.003&\best{0.856 } $\pm$  0.009&\best{1.000 } $\pm$  0.000&\best{0.981 } $\pm$  0.002& 0.895 $\pm$ 0.008 &\best{1.000 } $\pm$  0.000&\best{0.987 } $\pm$  0.002& 0.910 $\pm$ 0.008 &\best{1.000 } $\pm$  0.000& 0.990 $\pm$ 0.001\\
sensor-D &          * &                                         \mnet & 0.231 $\pm$ 0.007 & 0.798 $\pm$ 0.004 & 0.703 $\pm$ 0.005 & 0.353 $\pm$ 0.008 & 0.817 $\pm$ 0.004 & 0.736 $\pm$ 0.005 & 0.561 $\pm$ 0.009 & 0.842 $\pm$ 0.004 & 0.789 $\pm$ 0.004 & 0.739 $\pm$ 0.008 & 0.867 $\pm$ 0.003 & 0.839 $\pm$ 0.004 & 0.804 $\pm$ 0.007 & 0.881 $\pm$ 0.003 & 0.866 $\pm$ 0.003\\
\bottomrule\end{tabular}}

    \caption{Additional quantitative metrics for experiments on \mpmeshref test set images containing mirrors (Matterport3D~\cite{matterport3d} with ground truth depth refined using 3D mirror plane annotations).}
    \label{tab:m3d-refined-results-complete}
\end{table*}

\begin{table*}
\resizebox{\linewidth}{!}{
\begin{tabular}{@{} lll ccc ccc ccc ccc @{}}
\toprule
& & & \multicolumn{3}{c}{RMSE $\downarrow$} & \multicolumn{3}{c}{s-RMSE $\downarrow$} & \multicolumn{3}{c}{AbsRel $\downarrow$}  & \multicolumn{3}{c}{SSIM $\uparrow$}
\\\cmidrule(lr){4-6}\cmidrule(lr){7-9}\cmidrule(lr){10-12}\cmidrule(lr){13-15}
Input & Train & Method & Mirror & Other & All & Mirror & Other & All & Mirror & Other & All & Mirror & Other & All   \\
\midrule
sensor-D &          * &                                             * & 2.682 $\pm$ 0.073 & 0.960 $\pm$ 0.041 & 1.314 $\pm$ 0.038 & 2.335 $\pm$ 0.078 & 1.043 $\pm$ 0.040 & 1.179 $\pm$ 0.039 & 0.937 $\pm$ 0.021 & 0.144 $\pm$ 0.006 & 0.310 $\pm$ 0.015 & 0.289 $\pm$ 0.007 & 0.787 $\pm$ 0.006 & 0.710 $\pm$ 0.006\\
mesh-D &          * &                                             * &\best{0.000 } $\pm$  0.000&\best{0.000 } $\pm$  0.000&\best{0.000 } $\pm$  0.000&\best{0.000 } $\pm$  0.000&\best{0.000 } $\pm$  0.000&\best{0.000 } $\pm$  0.000&\best{0.000 } $\pm$  0.000&\best{0.000 } $\pm$  0.000&\best{0.000 } $\pm$  0.000&\best{1.000 } $\pm$  0.000&\best{1.000 } $\pm$  0.000&\best{1.000 } $\pm$  0.000\\
sensor-D &          * &                                         \mnet & 1.361 $\pm$ 0.040 & 0.950 $\pm$ 0.023 & 1.033 $\pm$ 0.023 & 1.253 $\pm$ 0.039 & 0.954 $\pm$ 0.023 & 0.984 $\pm$ 0.023 & 0.379 $\pm$ 0.009 & 0.151 $\pm$ 0.004 & 0.204 $\pm$ 0.006 & 0.685 $\pm$ 0.006 & 0.785 $\pm$ 0.003 & 0.767 $\pm$ 0.003\\
mesh-D &          * &                                         \mnet & 0.445 $\pm$ 0.018 & 0.156 $\pm$ 0.007 & 0.225 $\pm$ 0.008 & 0.413 $\pm$ 0.015 & 0.158 $\pm$ 0.007 & 0.210 $\pm$ 0.007 & 0.115 $\pm$ 0.005 & 0.016 $\pm$ 0.001 & 0.030 $\pm$ 0.001 & 0.877 $\pm$ 0.003 & 0.979 $\pm$ 0.001 & 0.965 $\pm$ 0.001\\
\bottomrule\end{tabular}}
\resizebox{\linewidth}{!}{
\begin{tabular}{@{} lll ccc ccc ccc ccc ccc @{}}
\toprule
& & & \multicolumn{3}{c}{1.05 $\uparrow$} & \multicolumn{3}{c}{1.10 $\uparrow$} & \multicolumn{3}{c}{1.25 $\uparrow$} & \multicolumn{3}{c}{${1.25}^2 \uparrow$} & \multicolumn{3}{c}{ ${1.25}^3\uparrow$}
\\\cmidrule(lr){4-6}\cmidrule(lr){7-9}\cmidrule(lr){10-12}\cmidrule(lr){13-15}\cmidrule(lr){16-18}
Input & Train & Method & Mirror & Other & All & Mirror & Other & All & Mirror & Other & All & Mirror & Other & All & Mirror & Other & All  \\
\midrule
sensor-D &          * &                                             * & 0.154 $\pm$ 0.007 & 0.834 $\pm$ 0.006 & 0.724 $\pm$ 0.008 & 0.171 $\pm$ 0.007 & 0.847 $\pm$ 0.006 & 0.737 $\pm$ 0.008 & 0.218 $\pm$ 0.008 & 0.861 $\pm$ 0.006 & 0.755 $\pm$ 0.007 & 0.310 $\pm$ 0.010 & 0.872 $\pm$ 0.005 & 0.778 $\pm$ 0.007 & 0.386 $\pm$ 0.011 & 0.879 $\pm$ 0.005 & 0.798 $\pm$ 0.007\\
mesh-D &          * &                                             * &\best{1.000 } $\pm$  0.000&\best{1.000 } $\pm$  0.000&\best{1.000 } $\pm$  0.000&\best{1.000 } $\pm$  0.000&\best{1.000 } $\pm$  0.000&\best{1.000 } $\pm$  0.000&\best{1.000 } $\pm$  0.000&\best{1.000 } $\pm$  0.000&\best{1.000 } $\pm$  0.000&\best{1.000 } $\pm$  0.000&\best{1.000 } $\pm$  0.000&\best{1.000 } $\pm$  0.000&\best{1.000 } $\pm$  0.000&\best{1.000 } $\pm$  0.000&\best{1.000 } $\pm$  0.000\\
sensor-D &          * &                                         \mnet & 0.222 $\pm$ 0.006 & 0.798 $\pm$ 0.004 & 0.705 $\pm$ 0.005 & 0.334 $\pm$ 0.008 & 0.817 $\pm$ 0.004 & 0.736 $\pm$ 0.005 & 0.529 $\pm$ 0.009 & 0.842 $\pm$ 0.004 & 0.787 $\pm$ 0.004 & 0.712 $\pm$ 0.008 & 0.867 $\pm$ 0.003 & 0.837 $\pm$ 0.004 & 0.786 $\pm$ 0.007 & 0.881 $\pm$ 0.003 & 0.864 $\pm$ 0.003\\
\bottomrule\end{tabular}}
    \caption{Additional quantitative metrics for experiments on \mpmesh test set images containing mirrors (Matterport3D~\cite{matterport3d} using original mesh-rendered depth as ground truth). Compare to \Cref{tab:m3d-refined-results-complete} and note the incorrect ranking of methods with this imperfect ground truth.}
    \label{tab:m3d-raw-results-complete}
\end{table*}

\clearpage
\clearpage

{\small
\bibliographystyle{plainnat}
\setlength{\bibsep}{0pt}
\bibliography{main}
}

\end{document}